\documentclass{article}

 \PassOptionsToPackage{numbers, compress}{natbib}

 \usepackage[eandd, final]{neurips_2026}

\usepackage[utf8]{inputenc}
\usepackage[T1]{fontenc}   
\usepackage{hyperref} 
\hypersetup{hidelinks}
\usepackage{url}      
\usepackage{amsfonts} 
\usepackage{nicefrac} 
\usepackage{microtype}
\usepackage[table]{xcolor}
\usepackage{graphicx}
\usepackage{makecell}
\usepackage{pifont}
\newcommand{\cmark}{\ding{51}}
\newcommand{\xmark}{\ding{55}}
\usepackage{ltablex}
\usepackage{placeins}
\usepackage{subcaption}
\usepackage{comment}
\usepackage{float}
\usepackage{capt-of}
\usepackage{multirow} 
\usepackage{booktabs}
\usepackage{longtable}
\usepackage{titletoc}
\usepackage{array}    
\usepackage{amsmath}
\usepackage{tikz}
\usetikzlibrary{arrows.meta,positioning}
\keepXColumns

\title{TabBioMed: A Large-Scale Benchmark for Biomedical Tabular Learning}

\newcommand{\aff}[1]{\textsuperscript{\mdseries #1}}
\newcommand{\equalcontrib}{\textsuperscript{*}}

\author{%
  \textbf{Pau Mateo-Bernadó}\aff{1}\equalcontrib \quad
  \textbf{Saivenkata Nagavyjayanthi Polapragada}\aff{2}\equalcontrib \\
  \textbf{David Bonet}\aff{2,3,4}\equalcontrib \quad
  \textbf{Pol Arbiol Rakuljic}\aff{5} \quad
  \textbf{Laia Mogas Pladevall}\aff{1} \\
  \textbf{Marçal Comajoan Cara}\aff{2,6} \quad
  \textbf{Jesus Gonzalez-Ferrer}\aff{3} \quad
  \textbf{Jordi Abante}\aff{4} \\
  \textbf{Daniel Mas Montserrat}\aff{2,3} \quad
  \textbf{David Haussler}\aff{3} \quad
  \textbf{Alexander G. Ioannidis}\aff{2,3}
  \\[0.5ex]
  \aff{1}Universitat Politècnica de Catalunya \hspace{0em}
\aff{2}Stanford University \hspace{0em}
\aff{3}UC Santa Cruz Genomics Institute \\
  \aff{4}Universitat de Barcelona \hspace{0.2em}
  \aff{5}Universitat Pompeu Fabra \hspace{0.2em}
  \aff{6}UC Berkeley \\[0.5ex]
  \texttt{contact@ioannidislab.ai}
}

\begin{document}

\maketitle
\begingroup
\renewcommand{\thefootnote}{\fnsymbol{footnote}}
\footnotetext[1]{These authors contributed equally to this work.}
\endgroup

\begin{abstract}

Recent advances in machine learning have accelerated biomedical prediction, but evaluation on structured biomedical data remains fragmented, especially for datasets with high dimensionality, large sample size, missingness, and class imbalance. We introduce TabBioMed, a large-scale benchmark of 94 curated public biomedical tabular datasets spanning electronic health records, drug discovery, genomics, transcriptomics, proteomics, metabolomics, single-cell omics, and systems biology. TabBioMed unifies these datasets through a standardized, dataset-aware preprocessing and evaluation framework, enabling reproducible comparison across classification, regression, and multi-target tasks. We benchmark 23 models across classical baselines, gradient-boosted trees, neural tabular architectures, and tabular foundation models. Foundation models deliver the strongest family-level predictive performance. However, their gains are regime-dependent: tree-based models and tuned multilayer perceptrons remain highly competitive, often dominating under inference-time or training-cost constraints. Overall, TabBioMed shows that biomedical tabular learning is strongly context-dependent, with optimal model choice depending on data regime, task type, and computational budget. We release TabBioMed with curated datasets and splits, preprocessing code, baselines, and results, providing a reproducible benchmark for practical model selection and future methods development in biomedical machine learning, available at \url{https://github.com/AI-sandbox/TabBioMed}.
\end{abstract}

\section{Introduction}
Tabular data are widely used in biomedical machine learning, representing structured information in electronic health records (EHR), genomics, proteomics, metabolomics, drug discovery, and single-cell studies. These datasets support predictive tasks critical to clinical decision-making and biological discovery \cite{wenteler2024ai, krones2025review}. 
Despite the success of modern machine learning methods, including gradient-boosted decision trees, deep neural networks, and emerging tabular foundation models (TFMs), systematic comparisons in biomedical prediction settings remain fragmented and inconsistent. 

Recent benchmarking efforts have improved the reproducibility and systematic evaluation of tabular models \cite{OpenML2025}. In particular, TabArena \cite{erickson2025tabarena} puts emphasis on standardized evaluation pipelines, careful dataset curation, and continuous benchmarking. However, existing benchmarks are designed largely for general-purpose tabular data and do not systematically capture the unique challenges of biomedical datasets, which often exhibit widely varying sample sizes and feature dimensionalities \citep{Bycroft2018, feldner2023dealing}, pervasive missingness \cite{wells2013strategies}, class imbalance, and modality-specific measurement processes \cite{abram2022comprehensive}. Consequently, the conclusions of the existing benchmarks may not transfer directly to many biomedical applications.
Prior studies have compared neural networks, gradient-boosted decision trees \cite{grinsztajn2022tree, mcelfresh2023neural}, and more recently tabular foundation models \cite{hollmann2025accurate}, but primarily on non-biomedical datasets. As a result, there remains no comprehensive benchmark spanning the full range of biomedical regimes.

\begin{figure}[t]
    \centering
    \includegraphics[width=\linewidth]{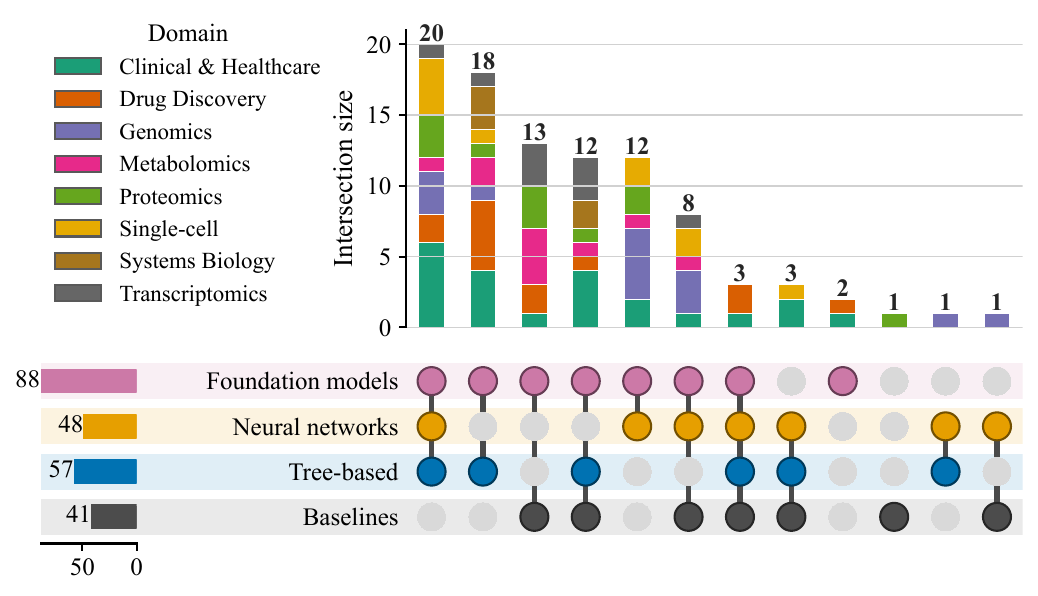}
    \caption{\textbf{TabBioMed overview across eight biomedical domains and four model families.} The UpSet plot summarizes which model families appear among the top five models for each dataset. Horizontal bars on the left show the total number of datasets in which each family is represented. Connected dots indicate the exact combination of families represented among the top five models for each dataset, while the stacked bars above show the number of datasets exhibiting that combination, stratified by biomedical domain.}
    \label{fig:results_summary_upset_plot}
\end{figure}

In this work, we address this gap through \textbf{TabBioMed},\footnote{\url{https://huggingface.co/spaces/TabBioMed}} a large-scale benchmark suite for biomedical tabular machine learning.
Our main contributions are: \textbf{(i) large-scale biomedical dataset curation}, comprising 94 biomedical tabular datasets across eight domains, including EHR, genomics, transcriptomics, drug discovery and molecular properties, proteomics, metabolomics, single-cell omics, and systems biology, \textbf{(ii) a unified preprocessing framework}, with a standardized but flexible pipeline that preserves domain-specific structure while enabling fair and reproducible comparisons, \textbf{(iii) comprehensive model evaluation}, benchmarking 23 models spanning linear methods, gradient-boosted decision trees, deep tabular architectures, and tabular foundation models, and \textbf{(iv) a robust benchmarking protocol}, providing standardized evaluation across datasets, models, and data regimes to support reliable large-scale benchmarking in biomedical machine learning.

We analyze how model families behave across domains (Figure~\ref{fig:results_summary_upset_plot}), how predictive performance relates to computational efficiency, and how dataset characteristics influence performance variation.
Tabular foundation models emerge as the strongest predictive family, although this advantage reflects family-level performance rather than uniform superiority of individual models. Missing evaluations occur more frequently for foundation models on larger, high-dimensional datasets due to computational constraints, but sensitivity analyses show that the main ranking trends remain under complete-case analysis and missing-score imputation. Under limited computational budgets, tree-based methods and well-tuned multilayer perceptrons (MLPs) remain highly competitive and often preferable. 
Overall, our analysis shows that performance is highly regime-dependent. 
Rather than identifying a universally optimal method, the benchmark highlights systematic variation across datasets, domains, and computational constraints, supporting more informed and dataset-aware model selection.

\section{Related Work}

The tabular machine learning community has developed numerous benchmark suites to evaluate predictive models on structured data. 
As summarized in Table \ref{tab:benchmark-comparison}, these benchmarks vary widely in their scale, dataset size regimes, and curation procedures. Large-scale collections such as PMLB \citep{olson2017pmlbalarge}, TALENT \citep{liu2025talent, ye2024acloserlook}, and TabRepo \citep{salinas2024tabrepo} prioritize breadth, offering hundreds of datasets across diverse domains to capture a wide array of learning scenarios. Conversely, efforts like OpenML-CC18 \citep{bischl2021openml}, TabZilla \citep{mcelfresh2023neural}, and TabArena \citep{erickson2025tabarena} employ more selective dataset curation strategies, often distilling hundreds of candidates down to high-quality, medium-sized subsets. Other suites address specific data regimes, such as TabReD \citep{rubachev2025tabred}, which targets industry-scale datasets.

Despite this progress, there is no single large-scale benchmark for tabular biomedical data that matches the scope of these general-purpose suites. Instead, biomedical datasets remain scattered as incidental subsets within broader multi-domain collections, which limits their utility for healthcare and biological settings due to a lack of domain-specific standardization. Although there are domain-specific benchmarks and comparative studies, such as MoleculeNet \citep{wu2018moleculenet} for molecular property prediction, MLOmics \citep{yang2025mlomics} for cancer multi-omics, tree-based and linear model comparison for biobank phenotype prediction \citep{melendez2023assessing}, and comparative studies of machine learning methods on metabolomics datasets \citep{mendez2019comparative, dlugas2025comparative}, they remain focused on particular biomedical domains or task families and do not provide a unified evaluation framework in biomedical settings.
Furthermore, exhaustive benchmarking on massive general-purpose suites can quickly become computationally prohibitive. For instance, the TabArena authors estimate a total of ${\sim}15$ wall-clock years for their experiments without parallelization \citep{erickson2025tabarena}. Through TabBioMed, we address these gaps with 94 curated datasets, balancing biomedical breadth with a tractable, standardized evaluation protocol.

\begin{table}[h]
  \begin{center}
  \caption{\textbf{Benchmark comparison.} We compare \textsc{TabBioMed} to other tabular benchmarks, not restricted to biomedicine. \emph{Manual curation}: whether the benchmark filters datasets based on criteria beyond simple automation. 
  \emph{HPO \#confs.\ /\ \#hours}: hyperparameter optimization budgets in \# of configurations and wall-clock hours, with exceptions noted below (``---'' if not specified, TabBioMed: 2 hours per CV fold or 6 hours per fixed-split search). 
  \emph{Missingness}: how missing values are handled. \emph{Biomedical scope}: whether the benchmark explicitly targets biomedical data (\emph{Incidental} = biomedical datasets present but not the focus).}
  \label{tab:benchmark-comparison}
  \small
  \setlength{\tabcolsep}{4pt}
  \renewcommand{\arraystretch}{1.15}
  \resizebox{\textwidth}{!}{%
  \begin{tabular}{l c r r r r r l l}
    \toprule
    \textbf{Benchmark} & \makecell[c]{\textbf{Manual}\\\textbf{curation}} & \textbf{Datasets} & \makecell[r]{\textbf{HPO}\\\textbf{\#confs.}} & \makecell[r]{\textbf{HPO}\\\textbf{\#hours}} & \makecell[r]{\textbf{Max}\\\textbf{features}} & \makecell[r]{\textbf{Max}\\\textbf{samples}} & \textbf{Missingness} & \makecell[l]{\textbf{Biomedical}\\\textbf{scope}} \\
    \midrule
    TabZilla \citep{mcelfresh2023neural}            & \xmark & 176$^*$  & 30        & 10/search         & 7{,}200       & 1{,}025{,}009 & Imputed       & Incidental \\
    PMLB \citep{olson2017pmlbalarge,pmlb_snapshot_20250225} & \xmark & 450      & ---       & ---               & 1{,}000 & $\sim1{,}025{,}010$ & Excluded      & Incidental \\
    TALENT \citep{liu2025talent, ye2024acloserlook} & \cmark & 300      & 100       & ---               & 308  & 153{,}004 & Imputed       & Incidental \\
    TabRepo \citep{salinas2024tabrepo}              & \xmark & 200      & 200       & 1/config.         & 10{,}936      & 400{,}000     & Preproc.      & Incidental \\
    AMLB \citep{gijsbers2024amlb}                   & \cmark & 104      & Not fixed & $\{1,4\}$/search  & 14{,}891  & 10{,}000{,}000  & Kept       & Incidental \\
    RealMLP\ \citep{holzmuller2025realmlp}          & \cmark & 90       & 50        & ---               & 10{,}000      & 500{,}000     & Drop / encode & Incidental \\
    OpenML-CC18 \citep{bischl2021openml}            & \cmark & 72       & ---       & ---               & 3{,}072    & 96{,}320     & Kept       & Incidental \\
    TabArena-v0.1 \citep{erickson2025tabarena}      & \cmark & 51       & 200       & 1/config.         & 1{,}776                                   & 150{,}000     & Kept          & Incidental \\
    Grinsztajn et al.\ \citep{grinsztajn2022tree}   & \cmark & 45       & $\sim$400 & ---               & 419                                       & 5{,}465{,}575\textsuperscript{a} & Dropped & Incidental \\
    TabReD \citep{rubachev2025tabred}               & \cmark & 8        & 100       & ---               & 986                                       & 423{,}795 & Imputed          & No \\
    MoleculeNet \citep{wu2018moleculenet}           & \cmark & 17       & 20        & 10/fit            & 1{,}024\textsuperscript{**}               & 437{,}929 & None           & \cmark  \\
    MLOmics \citep{yang2025mlomics}                 & \cmark & 20\textsuperscript{b}& ---               & ---       & 57{,}568\textsuperscript{c}   & 8{,}314       & Imputed       & \cmark     \\
    \midrule
    \rowcolor{green!8}
    \textbf{\textsc{TabBioMed} (Ours)} & \cmark & \textbf{94} & 100 & \makecell[r]{2/fold, 6/search} & 39{,}969 &  226{,}342 & Kept & \cmark \\
    \bottomrule
  \end{tabular}%
  }
  \\
  \end{center}
  {\footnotesize \cmark\ = yes, \xmark\ = no. \textsuperscript{*}TabZilla evaluated 176 datasets in total, from which a 36-dataset ``hard'' suite was curated. \textsuperscript{**}MoleculeNet's feature count refers to molecular fingerprint dimensions when first published. \textsuperscript{a}The largest transformed dataset contains 5{,}465{,}575 samples, while training sets were capped at 10{,}000 samples in the main benchmark and 50{,}000 in the large-scale analysis. \textsuperscript{b}MLOmics provides 20 task-specific datasets: 6 classification, 9 clustering, and 5 imputation datasets. Thus, not all 20 are supervised prediction benchmarks. \textsuperscript{c}Maximum feature count is the concatenated dimensionality across four omics modalities, rather than the dimensionality of a single modality.}
\end{table}

\section{TabBioMed Benchmark}

TabBioMed is a large-scale benchmark of heterogeneous biomedical structured prediction tasks. We next describe its dataset curation, preprocessing, and evaluation protocol.

\subsection{Datasets}
\label{sec:datasets}

\begin{figure}[h]
    \centering
    \includegraphics[width=1\linewidth]{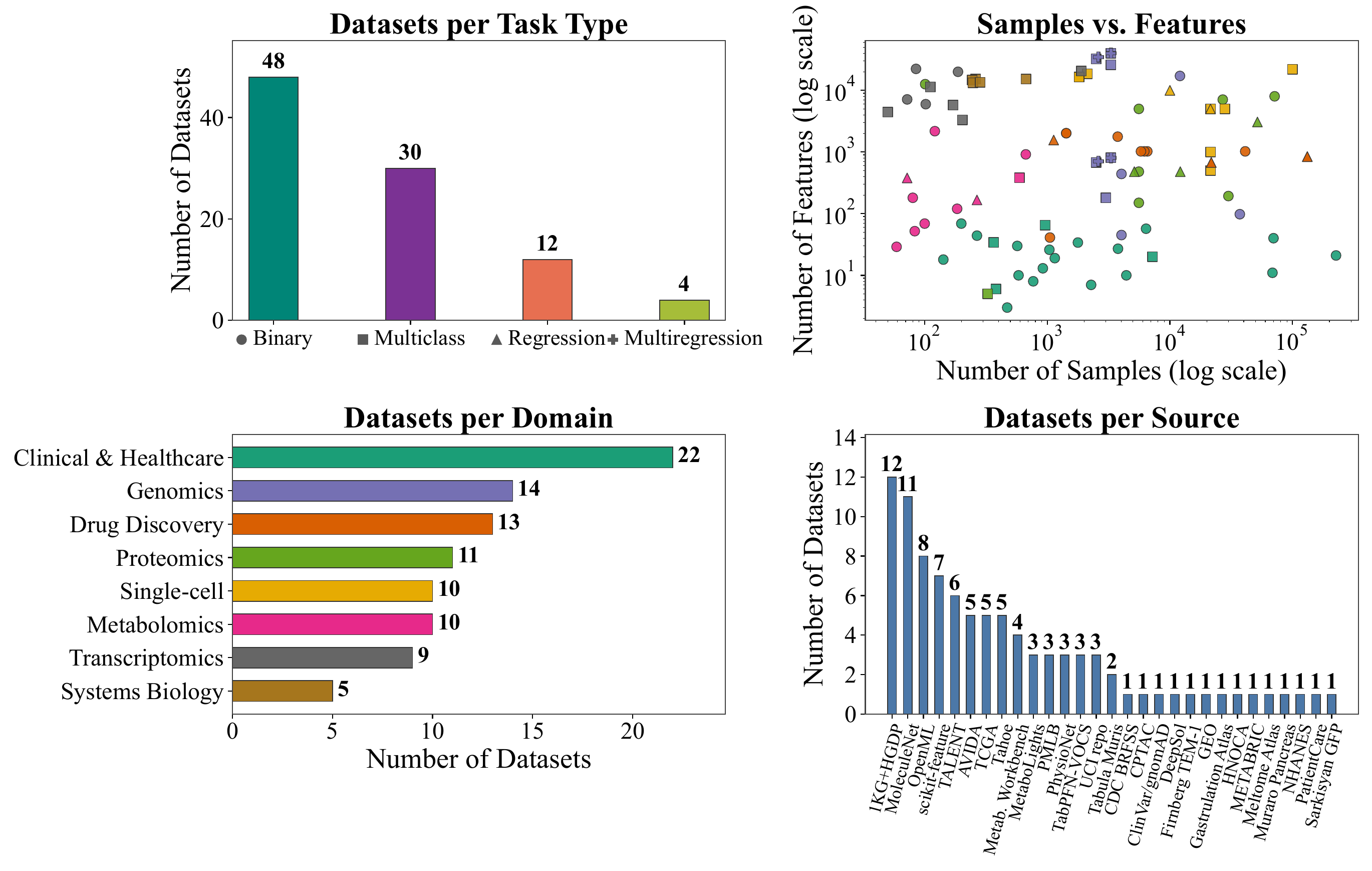}
    \caption{\textbf{TabBioMed Benchmark Datasets.} Distribution of datasets across task types, biomedical domains, and data sources, along with the relationship between dataset size and feature dimensionality.}
    \label{fig:dataset_summary_1}
\end{figure}

{\textbf{Dataset Curation.}}
We focus on biomedical classification and regression tasks based on clinical variables, omics measurements, and engineered molecular or sequence features. These datasets are represented as sample-by-feature tables. We exclude tasks requiring native sequence, image, graph, or temporal modeling. While some datasets exhibit dependencies across samples (e.g., repeated measurements per patient or batch effects), we do not assume that samples are independent. Instead, we use dataset-specific splitting protocols, including group-aware and scaffold-aware splits where appropriate. Figure~\ref{fig:dataset_summary_1} summarizes the resulting benchmark.

{\textbf{Dataset Unification and Preprocessing.}} Raw datasets are converted into a common representation consisting of a feature matrix $X$, target array $y$, and metadata describing task type, dataset statistics, and split information. A general validation stage checks data integrity, normalizes labels where needed, and verifies schema consistency. Model-specific imputation, encoding, and scaling are deferred to training. Dataset-specific details appear in Appendix~\ref{apx:dataset_curation_processing}. A consolidated curation flow appears in Figure~\ref{fig:curation_flow}.

{\textbf{Splitting Strategy.}} Evaluation splits are assigned during dataset preparation and stored in metadata. Datasets with $\leq 500$ samples use 5-fold cross-validation by default, with stratification for classification when possible. Larger datasets use predefined dataset-specific splits when available, otherwise 70/15/15 train/validation/test splits. Dataset-specific protocols take precedence over these defaults. They include grouped cross-validation and group-aware or scaffold-aware fixed splits, preserving the designated groups across partitions.

\subsection{Models and Hyperparameter Optimization}
\label{sec:models_hpo}

We benchmark 23 models across four families: classical baselines, tree-based models, neural tabular models, and tabular foundation models (Table~\ref{tab:tabbiomed_model_suite}). Classical baselines comprise LinearModel, LassoModel \citep{tibshirani1996lasso}, RidgeModel \citep{hoerl1970ridge}, ElasticNet \citep{zou2005regularization}, KNN \citep{cover1967nearest}, SVM \citep{cortes1995support}, and NeoLsSVM \citep{sorber2024neolssvm}. We also evaluate DecisionTree, RandomForest \citep{breiman2001random}, XGBoost \citep{chen2016xgboost}, CatBoost \citep{prokhorenkova2018catboost}, LightGBM \citep{ke2017lightgbm}, and xRFM \citep{beaglehole2026xrfm}. Neural tabular models include MLP, RealMLP \citep{holzmuller2025realmlp}, and TabM \citep{gorishniy2024tabm}. Tabular foundation models comprise TabPFNv2.5 and RealTabPFNv2.5 \citep{grinsztajn2025tabpfn}, TabPFN-Wide \citep{kolberg2025tabpfnwide}, TabICLv2 \citep{qu2026tabiclv2}, TabDPT \citep{ma2024tabdpt}, HyperFast \citep{bonet2024hyperfast}, and iLTM \citep{bonet2025iltm}.

\textbf{Implementation.} All models use a unified interface for preprocessing, training, hyperparameter optimization (HPO), and evaluation, with transformations fit only on training data and re-fit independently within each cross-validation fold. Default preprocessing uses numerical imputation, categorical encoding, and scaling, with model-specific overrides for methods with native missing-value, categorical, or internal preprocessing support (details in Appendix~\ref{sec:model_implementation}). 

\textbf{Hyperparameter Optimization.} Each model is evaluated with default settings, and HPO is performed for all models except LinearModel and NeoLsSVM. HPO details and search spaces are provided in Appendix~\ref{apx:hpo_search_space}. Optimization is constrained by wall-clock budgets of 6 hours per search for standard train/validation/test splits and 2 hours per outer fold for cross-validation. Main benchmark selection considers the default plus up to 100 attempted HPO configurations. These are equal wall-clock allowances, not equally exhaustive searches, and realized configuration counts are reported in Appendix~\ref{apx:hpo_budget_sensitivity}. Hyperparameters are selected based on validation performance using ROC-AUC for binary classification, log loss for multiclass classification, and MSE for regression tasks. For cross-validation, HPO is performed independently using inner validation within each outer fold's training data, and test metrics are averaged across the five outer folds.

\textbf{Computational cost.} The main benchmark required approximately
5{,}334 hours of cumulative logged runtime ($\sim$ 0.61 serial wall-clock years),
with 5\% spent on default evaluations and 95\% on tuning and
selected-model refitting.

\subsection{Evaluation}
\label{sec:evaluation_metrics}

We evaluate all methods on the benchmark splits introduced in Subsection~\ref{sec:datasets}, covering both fixed-split and cross-validation settings. For cross-validation, we follow an outer-fold protocol and report mean $\pm$ standard deviation across folds. For fixed splits, we report performance on a single held-out test set. For each dataset-model pair, we evaluate a default configuration followed, where applicable, by model-specific hyperparameter optimization. Hyperparameters are selected solely based on validation performance, with the test set reserved for final evaluation. The selected configuration is then refit according to model-specific protocols, including training on combined training and validation data where applicable (see Appendix~\ref{apx:final_refit_and_eval}). Final performance is evaluated on the held-out test set using the same task-specific primary metrics as HPO (Subsection~\ref{sec:models_hpo}).

We summarize predictive performance using ranks within each dataset and mean and median scores after per-dataset min-max normalization (Table~\ref{tab:rank_summary}). Normalized scores are also used in the pairwise Wilcoxon signed-rank tests. Elo ratings are computed from within-dataset pairwise comparisons of primary test scores, respecting metric directionality. Details are provided in Appendix~\ref{apx:benchmark_evaluation}. Pareto analyses compare Elo ratings with median training costs and inference times per 1,000 rows, including HPO in the training cost of Tuned configurations (Section~\ref{sec:results_analysis}).

\section{Results and Analysis}
\label{sec:results_analysis}

We compare models using normalized scores, dataset-level ranks, runtimes, and Elo ratings, and examine performance across biomedical domains and dataset size regimes.

\textbf{Rank-based analysis.} For each dataset, models are ranked using the primary metric from Section~\ref{sec:models_hpo} respecting metric directionality. We identify models appearing in the top-5 across datasets (Figure~\ref{fig:results_summary_upset_plot}). Foundation models are most frequently represented, appearing in the top-5 for 88 datasets (vs. 57 for tree-based methods, 48 for neural methods, and 41 for classical baselines). Tree-based and classical methods also appear frequently among the top-performing models, and top-performing sets often include a mix of foundation, tree-based, and baseline approaches, suggesting complementary strengths across biomedical tasks. Pairwise win rates across all datasets are reported in Appendix~\ref{appendix_global_analysis}. 
We restrict the main analysis to 21 models applicable to all 94 datasets, excluding classification-only models (TabPFN-Wide \citep{kolberg2025tabpfnwide} and HyperFast \citep{bonet2024hyperfast}).
As shown in Table \ref{tab:rank_summary}, foundation models occupy several of the leading positions in the aggregate rankings across the benchmark. TabDPT \citep{ma2024tabdpt} achieves the highest mean normalized score (0.865), while TabICLv2 \citep{qu2026tabiclv2} achieves the best mean rank (8.011). RealTabPFNv2.5 \citep{grinsztajn2025tabpfn} and TabPFNv2.5 \citep{grinsztajn2025tabpfn} also demonstrate consistently strong performance across diverse biomedical datasets. We compare these rankings with TabArena in Appendix~\ref{app:tabarena_comparison}.

\begin{table}[h]
\centering
\caption{\textbf{Aggregate model comparison} for 21 models across all 94 benchmark datasets.
Models are ranked per dataset using task-specific primary metrics and ordered by mean rank.
The table reports minimum, mean, and maximum rank, mean and median normalized scores,
median inference and final-fit training time per 1k samples (in seconds), and the number of datasets with available test results.
Score selection and failure handling follow Appendix~\ref{apx:cd_analysis}.
Timing medians use successful evaluations.
Higher normalized scores and lower mean ranks indicate better performance.}
\label{tab:rank_summary}
\small
\resizebox{\textwidth}{!}{
\begin{tabular}{llcccrrrrr}
\toprule
\multirow{2}{*}[-0.5ex]{Model} & \multirow{2}{*}[-0.5ex]{Model Family} &
\multicolumn{3}{c}{Rank} &
\multicolumn{2}{c}{Normalized Score} &
\multirow{2}{*}[-0.5ex]{\makecell{Inference \\ time/1k (s)}} &
\multirow{2}{*}[-0.5ex]{\makecell{Training \\ time/1k (s)}} &
\multirow{2}{*}[-0.5ex]{\makecell{Datasets \\ ($n/N$)}} \\
\cmidrule(lr){3-5}
\cmidrule(lr){6-7}
& & Min & Mean & Max & \multicolumn{1}{c}{Mean} & \multicolumn{1}{c}{Median} & & & \\
\midrule
TabICLv2 & Foundation models & 1.0 & 8.011 & 21.0 & 0.836 & 0.929 & 24.169 & 1.385 & 93/94 \\
RealTabPFNv2.5 & Foundation models & 1.0 & 8.048 & 20.0 & 0.797 & 0.926 & 76.282 & 4.277 & 84/94 \\
TabM & Neural networks & 1.0 & 8.170 & 21.0 & 0.861 & 0.914 & 0.417 & 23.758 & 94/94 \\
TabPFNv2.5 & Foundation models & 1.0 & 8.245 & 20.5 & 0.786 & 0.930 & 70.671 & 3.512 & 83/94 \\
TabDPT & Foundation models & 1.0 & 8.372 & 20.0 & 0.865 & 0.931 & 31.769 & 0.213 & 94/94 \\
RealMLP & Neural networks & 1.0 & 8.484 & 21.0 & 0.841 & 0.906 & 1.251 & 27.968 & 94/94 \\
iLTM & Foundation models & 1.0 & 8.649 & 20.0 & 0.848 & 0.909 & 21.224 & 104.105 & 94/94 \\
CatBoost & Tree-based & 1.0 & 9.059 & 21.0 & 0.823 & 0.918 & 0.046 & 3.854 & 93/94 \\
xRFM & Tree-based & 1.0 & 9.622 & 20.0 & 0.830 & 0.893 & 0.097 & 0.482 & 93/94 \\
LightGBM & Tree-based & 1.0 & 10.027 & 21.0 & 0.761 & 0.872 & 0.020 & 6.157 & 94/94 \\
XGBoost & Tree-based & 1.0 & 10.548 & 21.0 & 0.796 & 0.898 & 0.080 & 2.905 & 94/94 \\
MLP & Neural networks & 1.0 & 10.628 & 21.0 & 0.803 & 0.854 & 0.403 & 11.150 & 94/94 \\
SVM & Baselines & 1.0 & 10.809 & 21.0 & 0.744 & 0.862 & 0.829 & 1.865 & 94/94 \\
ElasticNet & Baselines & 1.0 & 10.936 & 21.0 & 0.765 & 0.884 & 0.006 & 3.532 & 94/94 \\
RidgeModel & Baselines & 1.0 & 11.213 & 20.0 & 0.760 & 0.880 & 0.003 & 0.042 & 94/94 \\
LassoModel & Baselines & 1.0 & 11.362 & 20.0 & 0.758 & 0.878 & 0.005 & 1.055 & 94/94 \\
RandomForest & Tree-based & 1.0 & 11.622 & 20.0 & 0.778 & 0.845 & 0.127 & 1.881 & 94/94 \\
LinearModel & Baselines & 1.0 & 15.207 & 21.0 & 0.538 & 0.658 & 0.006 & 1.250 & 94/94 \\
KNN & Baselines & 2.0 & 15.910 & 21.0 & 0.609 & 0.613 & 1.425 & 0.002 & 94/94 \\
NeoLsSVM & Baselines & 2.0 & 17.239 & 21.0 & 0.416 & 0.463 & 0.076 & 7.855 & 94/94 \\
DecisionTree & Tree-based & 1.0 & 18.840 & 21.0 & 0.279 & 0.076 & 0.002 & 0.069 & 94/94 \\

\bottomrule
\end{tabular}
}
\end{table}

Tree-based ensemble methods remain competitive. LightGBM \citep{ke2017lightgbm},
XGBoost \citep{chen2016xgboost}, and CatBoost \citep{prokhorenkova2018catboost} achieve mean normalized scores of 0.761, 0.796, and 0.823, respectively,
with median inference times of 0.020--0.080 seconds per 1k samples, which highlights an important trade-off between predictive performance and deployment efficiency.

Neural-network-based methods also perform strongly. Advanced architectures such as
RealMLP \citep{holzmuller2025realmlp} and TabM \citep{gorishniy2024tabm} achieve mean normalized scores of 0.841 and 0.861,
respectively, compared with 0.803 for the vanilla MLP baseline. TabM also achieves the third-best mean rank (8.170).

Classical baselines such as ElasticNet \citep{zou2005regularization}, RidgeModel, and SVM \citep{cortes1995support} remain moderately competitive,
with mean normalized scores of 0.765, 0.760, and 0.744, respectively. Their low inference costs, especially for the linear models,
indicate that simpler models may still be efficient for resource-constrained biomedical applications.

{\textbf{Statistical Significance.}} Consistent with the work of TabZilla \citep{mcelfresh2023neural}, we conducted statistical tests to assess whether observed differences in model performance are statistically significant. We first applied the Friedman test \citep{friedman1937use} to assess global differences in model rankings. The results indicate statistically significant differences between models at $p < 0.05$. We then performed pairwise Wilcoxon signed-rank tests \citep{wilcoxon1945individual} with Holm-Bonferroni correction to control for multiple comparisons \citep{holm1979simple}. For this test, we used normalized scores because the comparison spans all four task types. Score selection and failure handling are detailed in Appendix~\ref{apx:cd_analysis}.

\begin{figure}[t]
    \centering
    \includegraphics[width=\linewidth,trim=0 0 -54bp 0,clip]{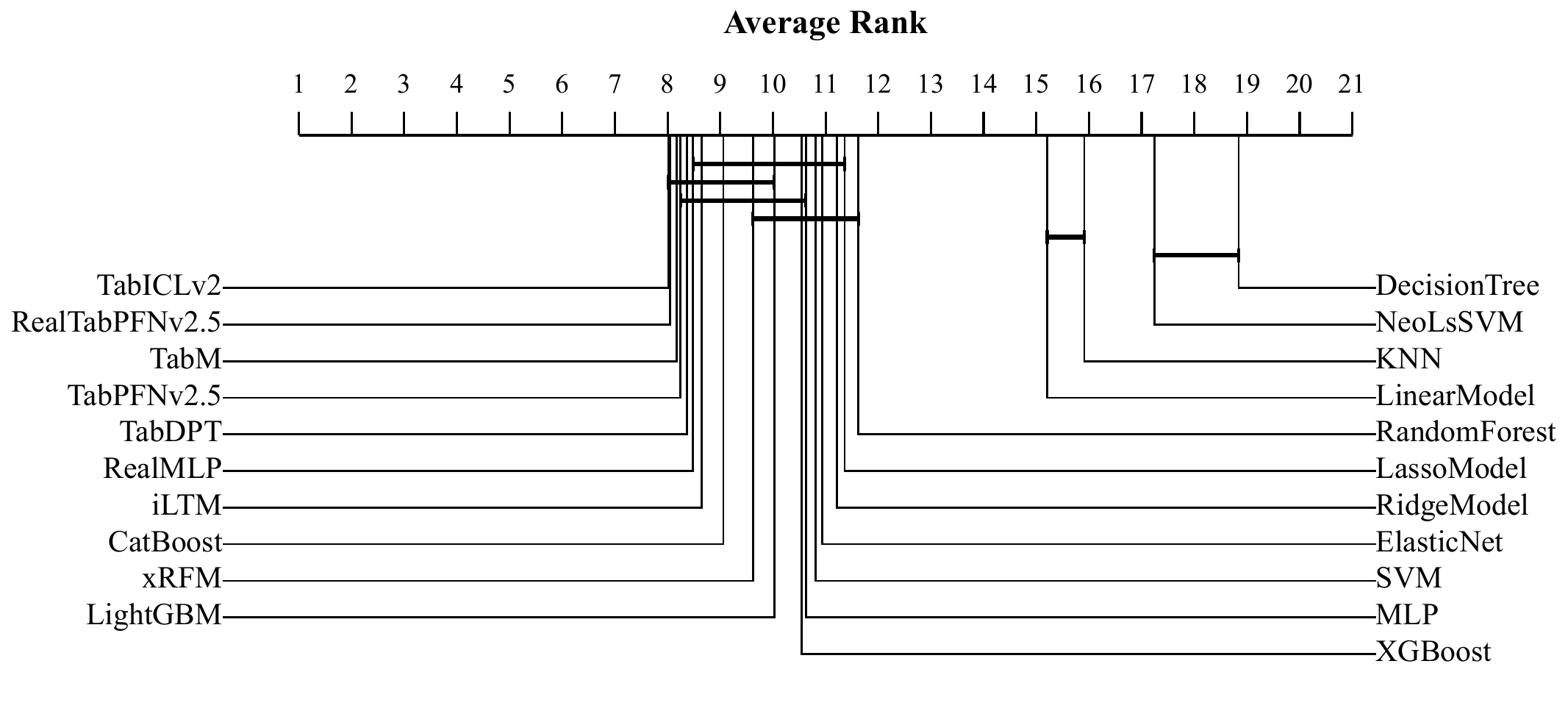}
    \caption{\textbf{Critical difference} (CD) diagram based on pairwise Wilcoxon signed-rank tests with Holm-Bonferroni correction across all 94 datasets. Models connected by horizontal bars do not show statistically significant pairwise differences at $\alpha=0.05$. Average ranks are computed using normalized scores across heterogeneous biomedical tasks.}
    \label{fig:wilcoxon_test_global}
\end{figure}

Although the Friedman test identified statistically significant global differences, the subsequent Wilcoxon-Holm analysis revealed substantial overlap among the leading methods as shown in Figure \ref{fig:wilcoxon_test_global}. The ten models with the best mean ranks are TabICLv2, RealTabPFNv2.5, TabM, TabPFNv2.5, TabDPT, RealMLP, iLTM, CatBoost, xRFM, and LightGBM, with no statistically significant pairwise differences within this group. This suggests that differences in average ranks among top-performing methods do not necessarily correspond to statistically significant performance differences. At the same time, weaker-performing methods such as DecisionTree, NeoLsSVM, KNN and LinearModel performed significantly worse than the leading methods, indicating lower aggregate performance across heterogeneous biomedical datasets.

{\textbf{Elo Rating.}} We adopt a pairwise ranking framework using Elo-style ratings \citep{elo1967proposed}. Default and Tuned configurations are rated jointly using a Bradley-Terry model based on pairwise comparisons, with ties counted as half a win for each configuration. We estimate 95\% confidence intervals using 200 bootstrap resamples over datasets and anchor the Default Random Forest rating to 1000, similar to TabArena \citep{erickson2025tabarena}. This analysis includes the same 21 models as Table~\ref{tab:rank_summary}.

\begin{figure}[!h]
    \centering
    \includegraphics[width=\linewidth]{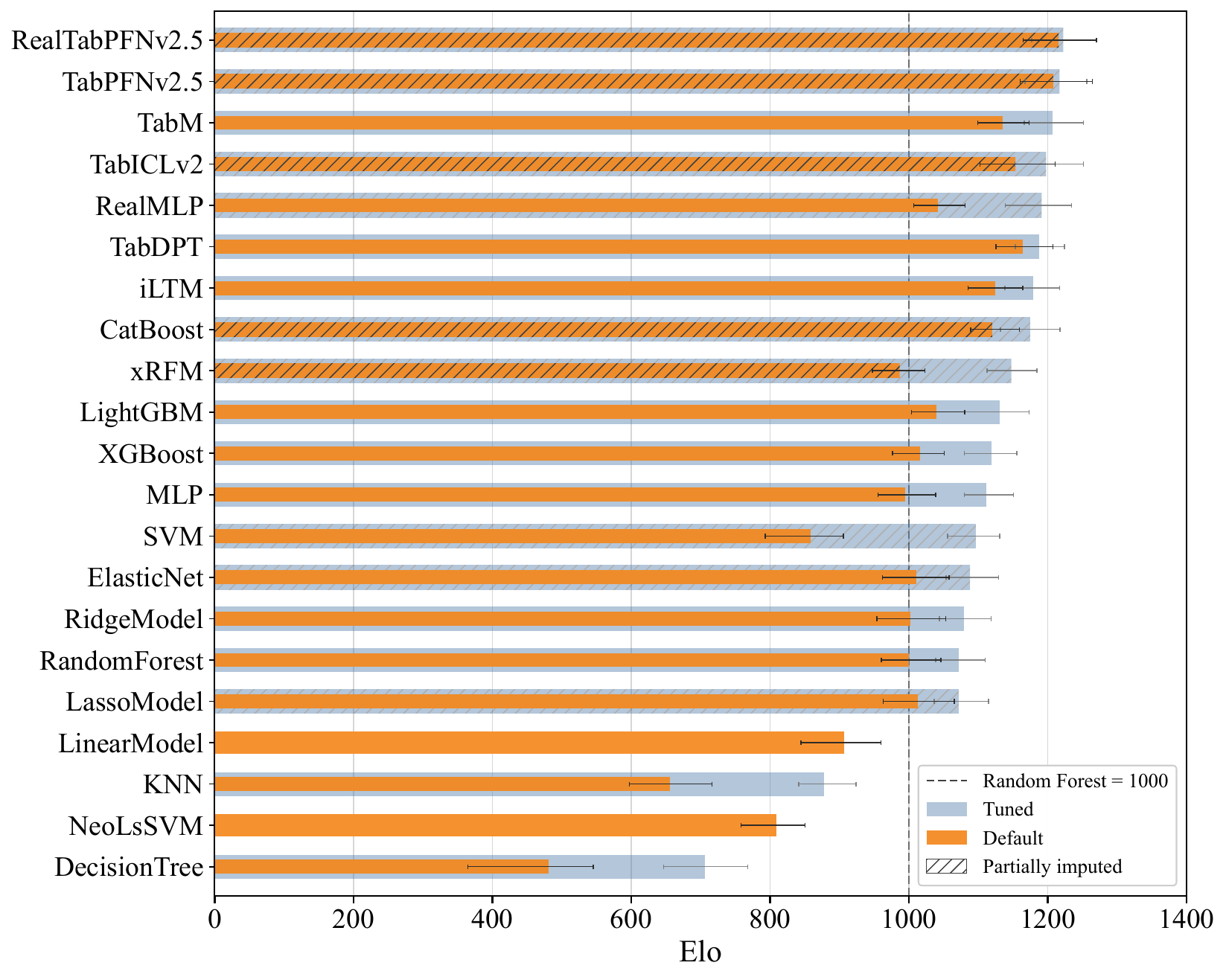}
    \caption{\textbf{TabBioMed Leaderboard}. Elo ratings compare Default and Tuned configurations of 21 models across all 94 datasets. Missing scores are imputed using Default Random Forest performance on that dataset. Diagonal hatching marks partially imputed Default or Tuned bars. Default Random Forest rating is calibrated to 1000 Elo. Error bars are 95\% dataset-bootstrap confidence intervals.}
    \label{fig:elo_rankings_global}
\end{figure}

Some Default and Tuned evaluations did not complete due to timeouts, out-of-memory errors, or crashes (Appendix~\ref{apx:failure_reporting}). We therefore report a Random-Forest-imputed analysis across all 94 datasets in Figure~\ref{fig:elo_rankings_global} and complete-case Elo rankings on 79 datasets in Appendix~\ref{appendix_global_analysis} (Figure~\ref{fig:elo_c_complete}). In the imputed analysis, each missing configuration--dataset score inherits the Default Random Forest score for that dataset. Further details are provided in Appendix~\ref{apx:elo_analysis}.

Elo analysis in Figure \ref{fig:elo_rankings_global} revealed a leading set of foundation models and advanced neural architectures. Among Tuned configurations, RealTabPFNv2.5 and TabPFNv2.5 have the highest Elo point estimates, followed by TabM, TabICLv2, RealMLP, TabDPT, iLTM, CatBoost, and xRFM. Their bootstrap intervals overlap, so the ordering should be interpreted cautiously. The complete-case analysis retains the same nine leading models at the top, with RealTabPFNv2.5 and TabPFNv2.5 remaining first and second, respectively (see Figure~\ref{fig:elo_c_complete}).
In contrast, DecisionTree and KNN remained below the Default Random Forest baseline even after tuning, while the Default-only LinearModel and NeoLsSVM also scored below this baseline.

\begin{figure}[!h]
    \centering
    \includegraphics[width=\linewidth]{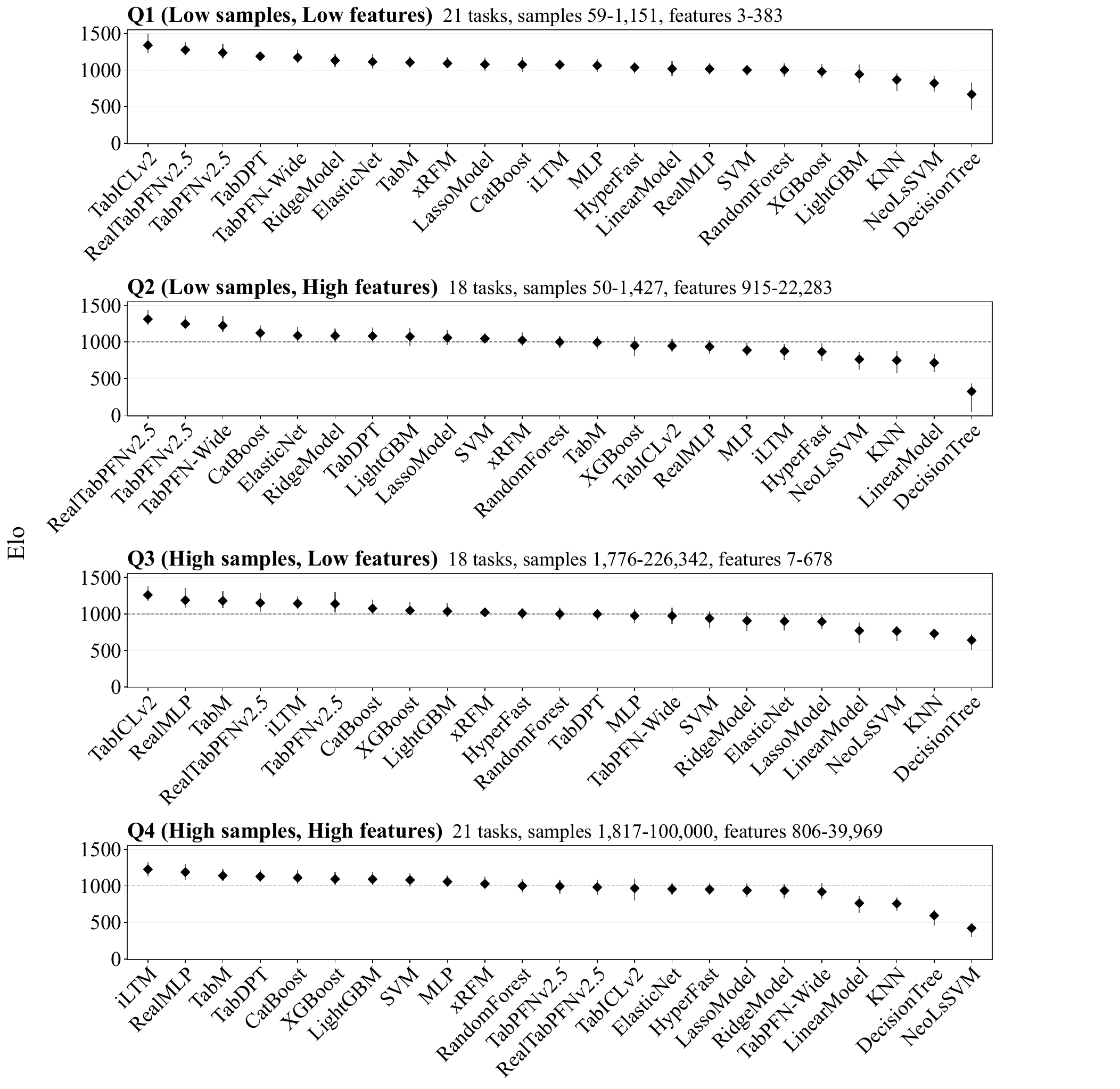}
    \caption{\textbf{Elo across dataset size and dimensionality regimes for 23 models on 78 classification tasks.} Median splits within the classification pool define Q1--Q4, containing 21, 18, 18, and 21 tasks, respectively. Tuned scores fall back to own Default, then RF Default. Diamonds show Elo point estimates and bars show 95\% confidence intervals from 200 dataset bootstrap resamples.}
    \label{fig:elo_regimes_classification_23}
\end{figure}

We audit pretraining source overlaps on 15 benchmark tasks and assess ranking sensitivity in Appendix~\ref{apx:pretraining_overlap}.
To assess the sensitivity of model rankings to GPU memory, we also retry evaluations that failed with GPU out-of-memory errors on the benchmark's 24\,GB A5500 GPUs using 80\,GB A100 GPUs (Appendix~\ref{apx:a100_sensitivity}). 
For binary classification, ROC-AUC and PR-AUC rankings largely agree (median Spearman $\rho=0.926$), whereas the proper scoring rules log loss and Brier score favor TabICLv2 over RealTabPFNv2.5, the ROC-AUC leader (see Appendix~\ref{apx:binary_metric_robustness}).

{\textbf{Impact of Dataset Size and Dimensionality.}} We analyze dataset size and dimensionality regimes by splitting datasets at the medians of $\log_{10}$ sample size and feature count, yielding four regimes (Q1--Q4). Figure~\ref{fig:elo_regimes_classification_23} focuses on the 78 binary and multiclass classification tasks across all 23 models. This classification-only view additionally includes TabPFN-Wide, designed for high-dimensional classification \citep{kolberg2025tabpfnwide}, and HyperFast \citep{bonet2024hyperfast}. The corresponding analysis of all 94 tasks with the 21 models supporting classification, regression, and multi-target regression, excluding TabPFN-Wide and HyperFast, is reported in Appendix~\ref{apx:elo_regime_all_tasks}.

Foundation models achieve the highest Elo point estimate in all four classification regimes, with different models leading across regimes. TabICLv2 ranks first in the low-sample, low-feature (Q1) and high-sample, low-feature (Q3) regimes. In Q3, it is followed by the MLP-based RealMLP and TabM, then RealTabPFNv2.5 and iLTM (a foundation model whose meta-trained hypernetwork generates an MLP-based prediction network). RealTabPFNv2.5 leads the high-dimensional, low-sample-size (HDLSS) regime (Q2), followed by TabPFNv2.5 and TabPFN-Wide, which uses a TabPFNv2 backbone. In the high-sample, high-feature regime (Q4), iLTM ranks first, followed by RealMLP and TabM.

We additionally test whether principal component analysis (PCA) before a TFM improves performance on high-dimensional inputs, comparing TabICLv2, TabPFNv2.5, and RealTabPFNv2.5 with their corresponding models without external PCA on all 36 datasets with more than 2,000 features (Appendix~\ref{apx:pca_tfm}). TabICLv2 benefits most, with 26 paired improvements and one successful evaluation where its non-PCA baseline failed, together covering 27 of 36 datasets. TabPFNv2.5 improves on 8 of 30 and obtains one new successful evaluation, and RealTabPFNv2.5 improves on 9 of 31 paired datasets. We hypothesize that the greater benefit for TabICLv2 might partly reflect its lack of native PCA/SVD preprocessing, whereas the TabPFN variants already combine feature subsampling with optional SVD-derived features.

\begin{figure}[ht]
    \centering
    \includegraphics[width=\linewidth]{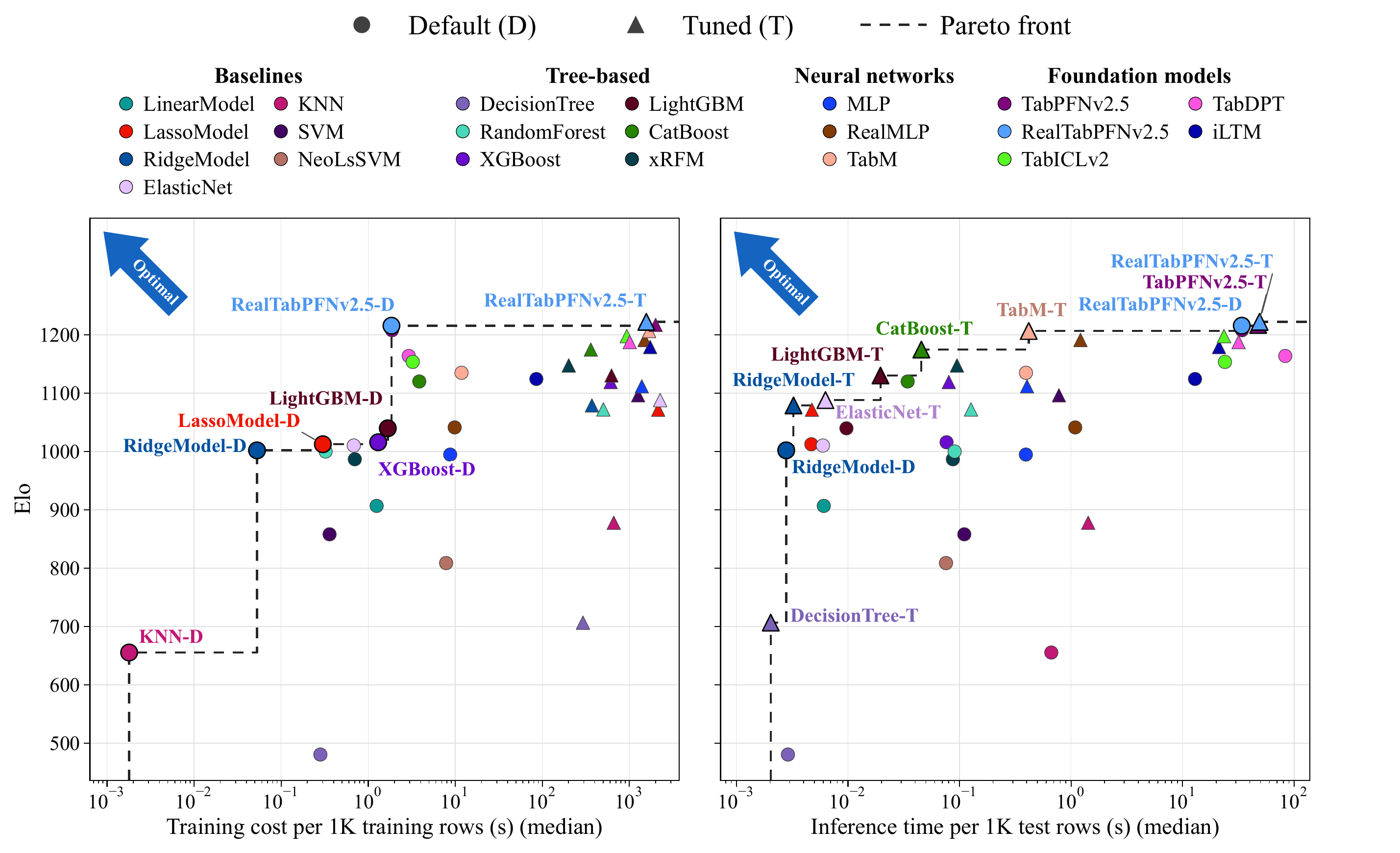}
    \caption{\textbf{Pareto frontier of predictive performance and computational cost.} Elo versus median training cost (left, including HPO for Tuned) and test inference time (right), per 1K training and test rows, respectively, across the 94 biomedical datasets. Circles are default (D) and triangles tuned (T) configurations. Points on the frontier (dashed line) identify Pareto-optimal model configurations in terms of predictive performance and computational cost.}
    \label{fig:pareto_scatter}
\end{figure}

We further analyze dataset metafeatures using PyMFE \cite{pymfe}, measuring family performance by the best observed normalized test score within each family in the 21-model comparison. Across the benchmark, classical baselines are more competitive at higher feature-to-sample ratios (Spearman $\rho = +0.435$, $p = 0.0021$), while the foundation--baseline score gap narrows ($\rho = -0.404$, $p = 0.0092$).

Performance of neural models is positively associated with sample size ($\rho = +0.373$, $p = 0.036$) and, among the 78 classification tasks, class count ($\rho = +0.498$, $p = 0.00063$), and negatively associated with mean absolute feature correlation ($\rho = -0.378$, $p = 0.029$). Reported correlation $p$-values are Holm-adjusted across all 180 metafeature--performance comparisons. Figures~\ref{fig:metafeature_correlations} and~\ref{fig:discovery_archetypes} summarize these associations and compare families within dataset clusters (Appendix~\ref{apx:metafeature_regime_analysis}).

{\textbf{Pareto Frontiers.}} We construct Pareto frontiers to analyze trade-offs between Elo ratings and computational cost across models. Pareto optimality is defined over two objectives: minimizing median inference time or training cost per 1,000 rows and maximizing Elo rating. We also report domain-specific Pareto-optimal models in Appendix~\ref{sec:domain_wise_analysis_appendix}. As shown in Figure~\ref{fig:pareto_scatter}, including HPO shifts the training-cost frontier toward Default configurations. Default LightGBM and RealTabPFNv2.5 combine competitive Elo ratings with low fitting cost, while Tuned RealTabPFNv2.5 reaches the highest Elo rating at greater training cost. For inference, Tuned LightGBM, CatBoost, and TabM offer favorable performance-latency trade-offs. LightGBM reaches near-frontier performance with inference times on the order of $10^{-2}$ seconds per 1k samples, several orders of magnitude faster than high-performing foundation models such as TabDPT and TabPFNv2.5.

\section{Conclusion}
\label{sec:conclusion_limitations}

In this work, we introduced \textbf{TabBioMed}, a large-scale benchmark for biomedical tabular machine learning comprising 94 curated datasets across diverse clinical and biological domains. 
By standardizing the evaluation pipeline and benchmarking 23 models spanning four major algorithmic families, we demonstrated that no single model class universally dominates biomedical structured prediction. 
Foundation models occupy several of the leading positions in aggregate rankings, alongside advanced neural architectures such as TabM and RealMLP. TabDPT achieves the highest mean normalized score, TabICLv2 the best mean rank, and RealTabPFNv2.5 the highest Tuned Elo rating. 
Dataset properties and preprocessing affect these comparisons. Classical baselines are more competitive on datasets with higher feature-to-sample ratios.
When accounting for computational efficiency, gradient-boosted decision trees and classical linear models remain on the Pareto frontier. Including HPO shifts the training-cost frontier toward Default configurations, while Tuned LightGBM, CatBoost, and TabM offer favorable inference-time trade-offs. 

Ultimately, TabBioMed emphasizes that optimal model selection in biomedicine is highly context- and regime-dependent. We release our curated datasets and splits, preprocessing code, baseline implementations, and benchmark results.

\textbf{Limitations and Broader Impact.} The benchmark is intended for comparative evaluation and methodological research rather than for direct clinical deployment or decision-making. Although TabBioMed spans 94 datasets across diverse biomedical domains, publicly available datasets may not fully represent real-world deployment settings, including clinical prediction over time \citep{zhou2023evaluating}, institution-specific pipelines \citep{otles2021mind}, rare diseases \citep{banerjee2023machine}, and underrepresented patient populations \citep{bonet2024underrepresented}. Our framework supports model-specific preprocessing, but may not fully capture specialized workflows commonly used across biomedical domains. 
Comparisons are conditional on the supplied tabular representations and do not establish the relative performance of alternative featurizations or modality-specific models.
In addition, not all model--dataset evaluations completed successfully because of timeouts, memory exhaustion, or crashes. Strong benchmark performance should not be interpreted as evidence of clinical readiness or deployment safety, as biomedical systems require additional considerations beyond predictive accuracy, including calibration, robustness, fairness, interpretability, prospective validation, and regulatory oversight.
TabBioMed may positively impact biomedical machine learning research by improving transparency, reproducibility, and standardized comparison of modern machine learning approaches. At the same time, biomedical datasets may contain demographic, institutional, geographic, and socioeconomic biases inherited from the original data collection processes, potentially leading to unequal model performance across populations or healthcare systems. Benchmark overfitting, overreliance on leaderboard rankings, and the computational costs associated with large-scale benchmarking should also be considered.

\begin{ack}

We thank Priyanka Manoj for helpful single-cell data discussions.
We thank the patients, study participants, and donors who contributed to the datasets used in TabBioMed, and the researchers, clinical teams, and consortia who created and shared the source datasets. We also thank the maintainers of the data repositories, benchmark collections, models, and research software used in this work, and the reviewers for their constructive feedback.
The Breast Cancer Wisconsin dataset was obtained from the University of Wisconsin Hospitals, Madison, from Dr. William H. Wolberg.
Some analyses in this work are based on data generated by The Cancer Genome Atlas (TCGA) Research Network (\url{https://www.cancer.gov/tcga}). Data used in this publication were generated by the Clinical Proteomic Tumor Analysis Consortium (CPTAC). M.C.C. was supported by a fellowship from ``la Caixa'' Foundation (ID 100010434, code B006360)
during part of the development of this project. This work was also supported by the National Institute of Mental Health (U24MH132628 to D.H.). The authors received no other external funding in direct support of this work.

\section*{Competing Interests}

D.B., M.C.C., D.M.M., and A.G.I. are a subset of the authors of iLTM. 
D.B., D.M.M., and A.G.I. are a subset of the authors of HyperFast. 
A.G.I. and D.M.M. have shares in Galatea Bio, Inc. The authors declare no other competing interests.
\end{ack}

\bibliographystyle{unsrtnat-clean}
\bibliography{references}

\clearpage
\appendix

\section*{Appendix contents}
\startcontents[appendix]
\printcontents[appendix]{}{1}{\setcounter{tocdepth}{2}}
\clearpage

\section{Dataset Curation and Preprocessing}
\label{apx:dataset_curation_processing}

\subsection{Curation Protocol at a Glance}
\label{apx:curation_overview}

\begin{figure}[!htbp]
    \centering
    \begin{tikzpicture}[
        curationbox/.style={draw=black!45, fill=black!2, rounded corners=2pt,
            line width=0.55pt, text width=0.28\linewidth, minimum height=24mm,
            inner sep=5pt, align=left, font=\footnotesize\rmfamily},
        curationarrow/.style={-{Latex[length=2mm]}, line width=0.65pt, draw=black!65},
        curationnote/.style={font=\scriptsize\rmfamily, text=black!70, fill=white, inner sep=2pt}
    ]
    \node[curationbox] (sources) {
        \textbf{1. Identify public sources}\\[3pt]
        Clinical, molecular and omics studies and repositories. Record the source dataset or release.
    };
    \node[curationbox, right=5mm of sources] (eligible) {
        \textbf{2. Check task eligibility}\\[3pt]
        Explicit classification or regression target and data that can be represented as fixed-length feature vectors.
    };
    \node[curationbox, right=5mm of eligible] (selection) {
        \textbf{3. Select complementary tasks}\\[3pt]
        Cover domains and data regimes. Limit duplicate tasks and overlapping endpoints.
    };
    \node[curationbox, below=12mm of selection] (unify) {
        \textbf{4. Curate and unify}\\[3pt]
        Define labels, remove leakage-prone variables and duplicates, preserve source-specific missingness and feature types.
    };
    \node[curationbox, below=12mm of unify] (features) {
        \textbf{5. Construct tabular features}\\[3pt]
        Where needed, construct tabular features, e.g., molecular fingerprints, Coulomb matrices, amino-acid encodings, k-mer features and sequence embeddings.
    };
    \node[curationbox, left=5mm of features] (splits) {
        \textbf{6. Define biomedical splits}\\[3pt]
        Define and store CV or fixed splits tailored to each dataset, e.g., molecular scaffold, sequence-cluster, patient- or batch-based splits.
    };
    \node[curationbox, left=5mm of splits] (release) {
        \textbf{7. Validate and document}\\[3pt]
        Check features, labels and partitions. Release source references, code, metadata and splits.
    };
    \node[draw=black!35, rounded corners=2pt, fill=white,
        text width=64mm, align=center, inner sep=4pt,
        font=\scriptsize\rmfamily, anchor=north east] (exclude) at (eligible.east |- unify.north) {
        Exclude tasks requiring native sequence, image, graph or temporal modeling, or lacking a defined supervised target.
    };
    \draw[curationarrow] (sources) -- (eligible);
    \draw[curationarrow] (eligible) -- (selection);
    \draw[curationarrow] (eligible.south) -- node[curationnote, right] {exclude} (eligible.south |- exclude.north);
    \draw[curationarrow] (selection) -- (unify);
    \draw[curationarrow] (unify) -- (features);
    \draw[curationarrow] (features) -- (splits);
    \draw[curationarrow] (splits) -- (release);
    \end{tikzpicture}
    \caption{\textbf{Dataset selection and curation workflow.} Dataset-specific protocols take precedence over default splits. Model-specific preprocessing is fitted on training data only.}
    \label{fig:curation_flow}
\end{figure}
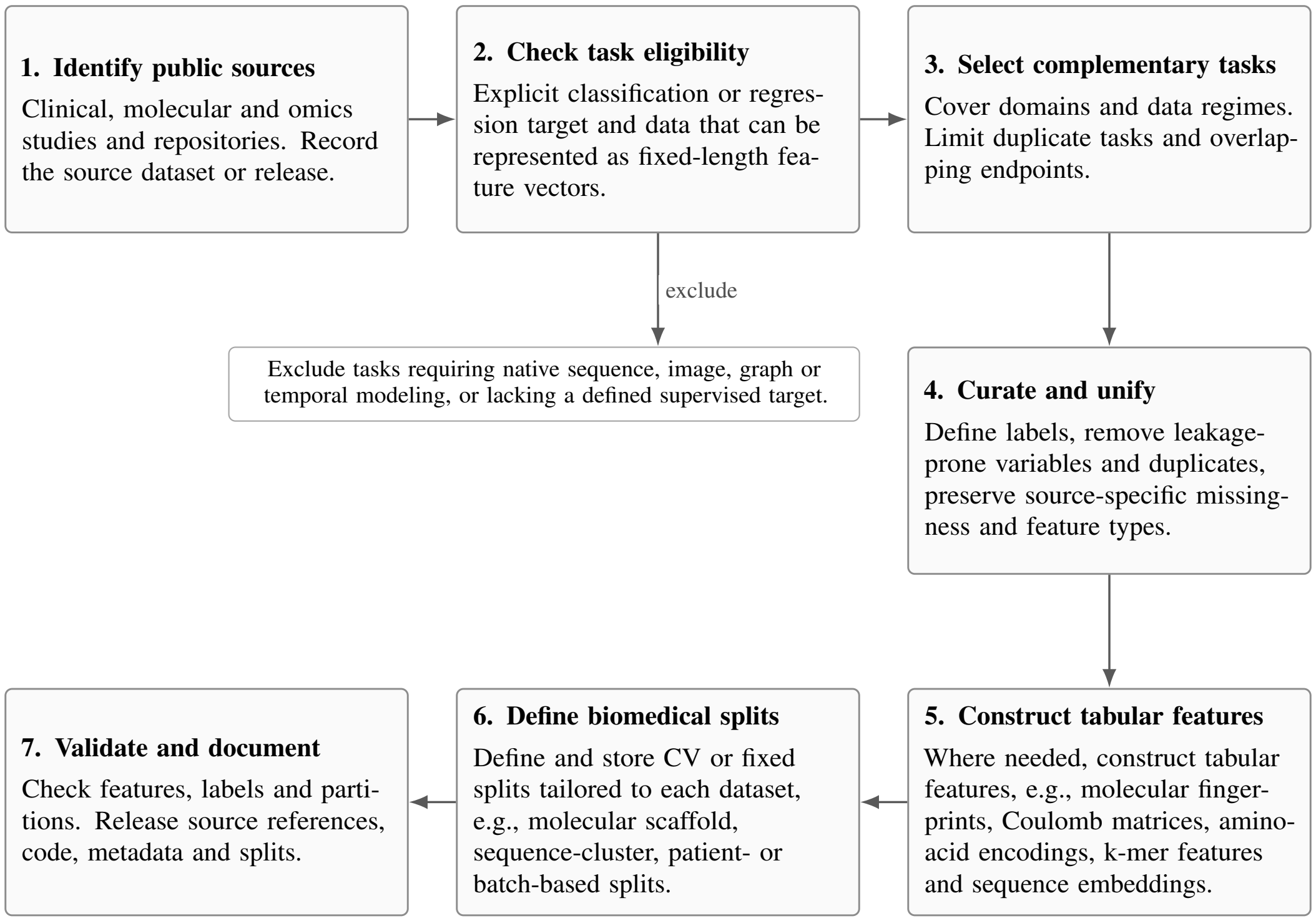

Figure~\ref{fig:curation_flow} summarizes dataset selection and preparation under the eligibility criteria in Section~\ref{sec:datasets}.

Dataset characteristics and sources are described in Appendix~\ref{apx:datset_inventory}, and domain-specific curation and preprocessing in Appendix~\ref{apx:dataset_unification}.
\FloatBarrier

\subsection{Dataset Inventory}
\label{apx:datset_inventory}

TabBioMed comprises 94 curated supervised learning datasets spanning diverse biomedical structured prediction tasks across molecular, clinical, and omics domains. The benchmark is designed to capture diversity while maintaining a unified tabular prediction setting, where each sample is represented as a fixed-dimensional feature vector derived from a shared schema.

Datasets were collected from several public repositories, published studies, and benchmark resources, including TALENT \citep{liu2025talent}, OpenML \citep{OpenML2025}, UCI Machine Learning Repository \citep{uci_ml_repository}, PhysioNet \cite{goldberger2000physiobank}, MoleculeNet \citep{wu2018moleculenet}, scikit-feature~\cite{li2018feature}, PMLB \citep{olson2017pmlbalarge}, MLOmics \citep{yang2025mlomics}, Clinical Proteomic Tumor Analysis Consortium (CPTAC) \citep{cptac}, Genome Aggregation Database (gnomAD) \citep{chen2024gnomad}, METABRIC \citep{curtis2012metabric}, Metabolomics Workbench \citep{sud2016metabolomics}, MetaboLights \cite{2024metabolights}, Hugging Face~\citep{lhoest-etal-2021-datasets} repositories, and multiple genomics and single-cell resources. To reduce redundancy and improve benchmark diversity, representative subsets were selected when repositories contained highly overlapping tasks or correlated endpoints. For example, representative tasks were chosen from Tox21 \citep{wu2018moleculenet, tox21challenge} and SIDER \cite{wu2018moleculenet, kuhn2016sider} to maximize coverage across complementary prediction objectives while avoiding excessive duplication. Duplicate or near-duplicate tasks across repositories were similarly excluded where appropriate.

Table~\ref{tab:dataset_domains} summarizes domain coverage and source composition, while Tables~\ref{tab:clinical-healthcare-datasets} through~\ref{tab:transcriptomics-datasets} provide the complete dataset inventory, including source references, benchmark identifiers, dataset characteristics, feature statistics, missingness, and task types. Task abbreviations are defined as follows: Bin = binary classification, MultiC = multiclass classification, Reg = regression, and MultiReg = multi-target regression. Classes / Targets indicates the number of classes for multiclass tasks and the number of targets for multi-target regression tasks.

\begin{table}[htbp]
    \centering
    \caption{Benchmark dataset distribution across biomedical domains and task types.}
    \resizebox{\textwidth}{!}{
    \begin{tabular}{lcccccc}
    \toprule
    Domain & \makecell{No. of \\ Datasets} & Binary &
    MultiClass & Regression &
    MultiRegression & Coverage  \\
    \midrule
    Clinical \& Healthcare & 22 & 18 & 4 & - & - & 23.4\% \\
    \makecell[l]{Drug Discovery and\\Molecular Properties} & 13 & 8 & - & 5 & - & 13.8\% \\
    Genomics & 14 & 4 & 6 & - & 4 & 14.9\% \\
    Metabolomics & 10 & 7 & 1 & 2 & - & 10.6\% \\
    Proteomics & 11 & 7 & 1 & 3 & - & 11.7\% \\
    Single-cell & 10 & - & 8 & 2 & - & 10.6\% \\
    Systems Biology & 5 & - & 5 & - & - & 5.3\% \\
    Transcriptomics & 9 & 4 & 5 & - & - & 9.6\% \\
    \midrule
    Total & 94 & 48 & 30 & 12 & 4 & 100\% \\
    \bottomrule
    \end{tabular}
    }
    \label{tab:dataset_domains}
\end{table}

\FloatBarrier
\begin{table}[!htbp]
\centering
\caption{Clinical and Healthcare dataset inventory. Each row reports the dataset source, number of samples and features, categorical feature fraction, missingness, class count, and prediction task.}
\label{tab:clinical-healthcare-datasets}
\resizebox{\textwidth}{!}{
\begin{tabular}{llrrrrrl}
\toprule
Dataset & Ref. &
\makecell{No. of \\ Samples} &
\makecell{No. of \\ Features} &
\makecell{Categorical \\ Features (\%)} &
\makecell{Missingness \\ (\%)} &
\makecell{Classes / \\ Targets} &
Task \\
\midrule
CDC\_Diabetes\_Health\_Indicators & CDC BRFSS \citep{liu2025talent}& 226,342
& 21
& 66.67
& -&

-& Bin
\\
Cardiovascular-Disease-dataset & \citep{liu2025talent}& 68,606
& 11
& 54.55
& -&

-& Bin
\\
Diabetes\_UCI & \citep{clore2014diabetes}& 69,973
& 40
& 80.00
& -&

-& Bin
\\
Diabetic\_Retinopathy\_Debrecen & \citep{liu2025talent}& 1,151
& 19
& 15.79
& -&

-& Bin
\\
dermatology & UCI-ID-33 \citep{olson2017pmlbalarge,uci_ml_repository}& 366
& 34 & 97.06 & 0.06 & 6 & MultiC \\
HeartDisease\_UCI & \citep{janosi1989heart}& 920
& 13
& 61.54
& 16.15
&

-& Bin
\\
IACCD & \citep{PhysioNet-mimic2-iaccd-1.0,data2016secondary,goldberger2000physiobank}& 1,776
& 34
& 38.24
& 2.06
&

-& Bin
\\
MUSIC & \citep{PhysioNet-music-sudden-cardiac-death-1.0.1} \citep{martin2024music} \citep{goldberger2000physiobank}& 961
& 65
& 41.54
& 2.05
& 4
& MultiC
\\
National\_Health\_and\_Nutrition\_Health\_Survey & NHANES \citep{liu2025talent}& 2,278
& 7
& 42.86& 0.01&

-& Bin
\\
Parkinson\_Speech & \citep{6451090} \citep{kursun2014parkinson}& 1,040
& 26
& -& -&

-& Bin
\\
PatientCare & \citep{sadikin2020ehr}& 4,412
& 10
& 10.00
& -&

-& Bin
\\
VitalDB & \citep{PhysioNet-vitaldb-1.0.0} \citep{lee2022vitaldb} \citep{goldberger2000physiobank}& 6,388
& 57
& 36.84
& 2.90
&

-& Bin
\\
audiology & OpenML-ID-999& 199
& 69
& 100.00
& 2.04
&

-& Bin
\\
blood-transfusion-service & OpenML-ID-1464 \citep{blood_transfusion_service_center_176}& 471
& 3
& -& -&

-& Bin
\\
ilpd\_patient\_data & OpenML-ID-1480 \citep{ramana2012critical}& 583
& 10
& 10.00
& -&

-& Bin
\\
lymphography & OpenML-ID-10 \citep{zwitter1988university}& 142
& 18
& 83.33
& -& -
& Bin
\\
maternal\_health\_risk & \citep{liu2025talent}& 381
& 6
& -& -& 3
& MultiC
\\
pima\_diabetes & OpenML-ID-37& 768
& 8
& -& 10.61&

-& Bin
\\
sick & OpenML-ID-38 \citep{thyroid_disease_102}& 3,772
& 27
& 77.78
& 2.25
&

-& Bin
\\
thyroid & \citep{liu2025talent}& 7,200
& 20
& 70.00& -& 3
& MultiC
\\
spectf & UCI-ID-96 \citep{olson2017pmlbalarge,uci_ml_repository}& 267
& 44 & - & - & - & Bin \\
wisconsin-breast-cancer & OpenML-ID-1510 \citep{street1993nuclear}& 569 & 30 & -& - & -& Bin \\
\bottomrule
\end{tabular}
}
\end{table}
\begin{table}[!htbp]
\centering
\caption{Drug Discovery and Molecular Properties dataset inventory. Each row reports the dataset source, number of samples and features, categorical feature fraction, missingness, class or target count, and prediction task.}
\label{tab:drug-discovery-datasets}
\resizebox{\textwidth}{!}{
\begin{tabular}{llrrrrrl}
\toprule
Dataset & Ref. &
\makecell{No. of \\ Samples} &
\makecell{No. of \\ Features} &
\makecell{Categorical \\ Features (\%)} &
\makecell{Missingness \\ (\%)} &
\makecell{Classes / \\ Targets} &
Task \\
\midrule
ESOL & \citep{wu2018moleculenet} & 1,128
& 1,563
& -
& -
& -& Reg
\\
HIV & \citep{wu2018moleculenet} & 41,120
& 1,024
& -& -& -& Bin
\\
QM8\_E1-CC2 & \citep{wu2018moleculenet}& 21,747
& 676
& -& -& -& Reg
\\
QM8\_f1-CC2 &\citep{wu2018moleculenet} & 21,747
& 676
& -& -& -& Reg
\\
QM9\_g298 &\citep{wu2018moleculenet} & 131,970
& 841
& -& -& -& Reg
\\
QM9\_gap &\citep{wu2018moleculenet} & 131,970
& 841
& -& -& -& Reg
\\
QSAR\_biodegradation & \citep{liu2025talent}& 1,051
& 41
& -& -& -& Bin
\\
SIDER\_gastro & \citep{wu2018moleculenet}& 1,427
& 2,018
& -& -& -& Bin
\\
SIDER\_nervous &\citep{wu2018moleculenet} & 1,427
& 2,018
& -& -& -& Bin
\\
Tox21\_NRAhR & \citep{wu2018moleculenet}& 6,542
& 1,024
& -& -& -& Bin
\\
Tox21\_NRER &\citep{wu2018moleculenet} & 6,186
& 1,024
& -& -& -& Bin
\\
Tox21\_SRMMP &\citep{wu2018moleculenet} & 5,804
& 1,024
& -& -& -& Bin
\\
bioresponse & OpenML-ID-4134 \citep{bioresponse} & 3,751
& 1,773
& -& -& -& Bin
\\
\bottomrule
\end{tabular}
}
\end{table}
\begin{table}[!htbp]
\centering
\caption{Genomics dataset inventory. Each row reports the dataset source, number of samples and features, categorical feature fraction, missingness, class or target count, and prediction task.}
\label{tab:genomics-datasets}
\resizebox{\textwidth}{!}{
\begin{tabular}{lcrrrrrl}
\toprule
Dataset & Ref. &
\makecell{No. of \\ Samples} &
\makecell{No. of \\ Features} &
\makecell{Categorical \\ Features (\%)} &
\makecell{Missingness \\ (\%)} &
\makecell{Classes / \\ Targets} &
Task \\
\midrule
continental\_1k & \citep{koenig2024harmonized} & 3,309
& 806
& -& 9.08
& 7
& MultiC
\\
continental\_50k & \citep{koenig2024harmonized}& 3,309
& 39,969
& -& 9.52
& 7
& MultiC
\\
continental\_50k\_missing &\citep{koenig2024harmonized} & 3,309
& 25,921
& -& 63.98
& 7
& MultiC
\\
coords\_1k &\citep{koenig2024harmonized} & 3,309
& 806
& -& 9.08
&
2& MultiReg
\\
coords\_50k &\citep{koenig2024harmonized} & 3,309
& 39,969
& -& 9.52
&
2& MultiReg
\\
dna &\citep{liu2025talent} & 3,000
& 180
& 100
& -& 3
& MultiC
\\
pca\_1k & \citep{koenig2024harmonized}& 2,609
& 707
& -& 8.70
&
3& MultiReg
\\
pca\_50k & \citep{koenig2024harmonized}& 2,609
& 34,635
& -& 9.72
&
3& MultiReg
\\
sexcheck\_chrX\_50 & \citep{koenig2024harmonized}& 4,026
& 45
& -& 6.63
&
-& Bin
\\
sexcheck\_chrX\_500 &\citep{koenig2024harmonized} & 4,026
& 441
& -& 7.29
&
-& Bin
\\
sexcheck\_chrX\_50k &\citep{koenig2024harmonized} & 12,078
& 17,118
& -& 95.44
&
-& Bin
\\
subpop\_1000\_consec &\citep{koenig2024harmonized} & 2,501
& 678
& -& 12.48
& 26
& MultiC
\\
subpop\_50k & \citep{koenig2024harmonized}& 2,501 & 32,339 & -& 8.66 & 26& MultiC\\
vep\_pathogenic\_coding & \citep{instadeep2024codingdataset,kao2024advancing}& 37,166 & 98 & 67.35 & - & -& Bin \\
\bottomrule
\end{tabular}
}
\end{table}
\begin{table}[!htbp]
\centering
\caption{Metabolomics dataset inventory. Each row reports the dataset source, number of samples and features, categorical feature fraction, missingness, class or target count, and prediction task.}
\label{tab:metabolomics-datasets}
\resizebox{\textwidth}{!}{
\begin{tabular}{lcrrrrrl}
\toprule
Dataset & Ref. &
\makecell{No. of \\ Samples} &
\makecell{No. of \\ Features} &
\makecell{Categorical \\ Features (\%)} &
\makecell{Missingness \\ (\%)} &
\makecell{Classes / \\ Targets} &
Task \\
\midrule
Gum1\_s & \citep{granitto2025use} & 267
& 167
& -& -& -& Reg
\\
MTBLS136\textsuperscript{b} & \citep{MTBLS136_stevens2018serum} & 668
& 915
& -& 16.91
& -& Bin
\\
MTBLS161\textsuperscript{b} & \citep{MTBLS161_armstrong2015metabolic} & 59
& 29
& -& -& -& Bin
\\
MTBLS404\textsuperscript{b} & \citep{MTBLS404_thevenot2015analysis} & 184
& 120
& -& -& -& Bin
\\
Noc\_s & \citep{granitto2025use} & 72
& 380
& -& -& -& Reg
\\
PT-Mush21 & \citep{granitto2025use} & 593
& 383
& -& -& 6& MultiC
\\
ST000369\textsuperscript{a} & \citep{ST000369_fahrmann2015investigation} & 80
& 181
& -& 0.06
& -& Bin
\\
ST000496\textsuperscript{a} & \citep{ST000496_sakanaka2017distinct} & 100
& 69
& -& -& -& Bin
\\
ST001000\textsuperscript{a} & \citep{ST001000_franzosa2019gut} & 121
& 2,177
& -& 35.41
& -& Bin
\\
ST001047\textsuperscript{a} & \citep{ST001047_chan2016nmr} & 83
& 52
& -& 2.13
& -& Bin \\
\bottomrule
\end{tabular}
}
\footnotesize{
\textsuperscript{a} Indicates data sourced from Metabolomics Workbench (\url{https://www.metabolomicsworkbench.org}). The corresponding repository study ID is given by the dataset name shown in the table.
\textsuperscript{b} Indicates data sourced from MetaboLights (\url{https://www.ebi.ac.uk/metabolights/}). The corresponding repository study ID is given by the dataset name shown in the table.
}
\end{table}
\begin{table}[!htbp]
\centering
\caption{Proteomics dataset inventory. Each row reports the dataset source, number of samples and features, categorical feature fraction, missingness, class or target count, and prediction task.}
\label{tab:proteomics-datasets}
\resizebox{\textwidth}{!}{
\begin{tabular}{lcrrrrrl}
\toprule
Dataset & Ref. &
\makecell{No. of \\ Samples} &
\makecell{No. of \\ Features} &
\makecell{Categorical \\ Features (\%)} &
\makecell{Missingness \\ (\%)} &
\makecell{Classes / \\ Targets} &
Task \\
\midrule
avida\_hil6\_onehot & \cite{tsuruta2023avida,cognano_2025} & 29,995
& 193
& 100& -& -& Bin
\\
avida\_htnfa\_3mer\textsuperscript{a} & \citep{cognano_htnfa} & 5,580
& 5,000
& -& -& -& Bin
\\
avida\_htnfa\_esm2\textsuperscript{a} & \citep{cognano_htnfa} & 5,580
& 480
& -& -& -& Bin
\\
avida\_htnfa\_onehot\textsuperscript{a} & \citep{cognano_htnfa} & 5,580
& 150
& 100& -& -& Bin
\\
avida\_sarscov2\_3mer & \citep{tsuruta2024sars} & 26,979
& 7,072
& -& -& -& Bin
\\
cptac\_histology & \citep{cptac,dou2020proteogenomic,cptac2023archive} & 101
& 12,592
& -& 19.19& -& Bin
\\
ecoli & UCI-ID-39 \citep{olson2017pmlbalarge,uci_ml_repository} & 327
& 5 & - & - & 5 & MultiC \\
proteinea\_solubility & \citep{khurana2018deepsol,proteinea_2023} & 71,421
& 8,000
& -& -& -& Bin
\\
true\_betalactamase\_unique & \citep{firnberg2014comprehensive} & 5,128
& 480
& -& -& -& Reg
\\
true\_fluorescence & \citep{sarkisyan2016local} & 51,738
& 3,068
& -& -& -& Reg
\\
true\_melting\_point & \citep{jarzab2020meltome} & 12,144& 480& -& -& -& Reg\\
\bottomrule
\end{tabular}
}
\footnotesize{
\textsuperscript{a} hTNFa dataset and details are available at
\url{https://cognanous.com/datasets/avida-htnfa}
and \url{https://huggingface.co/datasets/COGNANO/AVIDa-hTNFa}.}
\end{table}
\begin{table}[!htbp]
\centering
\caption{Single-cell dataset inventory. Each row reports the dataset source, number of samples and features, categorical feature fraction, missingness, class or target count, and prediction task.}
\label{tab:single-cell-datasets}
\resizebox{\textwidth}{!}{
\begin{tabular}{lcrrrrrl}
\toprule
Dataset & Ref. &
\makecell{No. of \\ Samples} &
\makecell{No. of \\ Features} &
\makecell{Categorical \\ Features (\%)} &
\makecell{Missingness \\ (\%)} &
\makecell{Classes / \\ Targets} &
Task \\
\midrule
hnoca\_cells30 &\citep{he2024integrated} & 100,000
& 22,000
& -& -& 30& MultiC
\\
sc\_gastrulation\_stage & \citep{pijuansala2019gastrulation} & 10,000
& 10,000
& -& -& -& Reg
\\
sc\_muris\_age & \citep{tabulamuris2020ageing} & 1,817
& 16,436
& -& -& 3& MultiC
\\
sc\_muraro\_celltype & \citep{muraro2016pancreas} & 2,119
& 18,508
& -& -& 8& MultiC
\\
sc\_muris\_celltype & \citep{tabulamuris2020ageing} & 1,817
& 16,472
& -& -& 9& MultiC
\\
tahoe\_batch\_integration\_14plates &\citep{zhang2026tahoe} & 28,000
& 5,000
& -& -& 14& MultiC
\\
tahoe\_cell\_name\_1000 & \citep{zhang2026tahoe}& 21,426
& 1,000
& -& -& 50& MultiC
\\
tahoe\_cell\_name\_500 & \citep{zhang2026tahoe}& 21,426
& 500
& -& -& 50& MultiC
\\
tahoe\_cell\_name\_5000 & \citep{zhang2026tahoe}& 21,426
& 5,000
& -& -& 50& MultiC
\\
tahoe\_g2m\_score & \citep{zhang2026tahoe} & 21,426 & 5,000& -& -& -& Reg\\
\bottomrule
\end{tabular}
}
\end{table}
\begin{table}[!htbp]
\centering
\caption{Systems Biology dataset inventory. Each row reports the dataset source, number of samples and features, categorical feature fraction, missingness, class or target count, and prediction task.}
\label{tab:systems-biology-datasets}
\resizebox{\textwidth}{!}{
\begin{tabular}{llrrrrrl}
\toprule
Dataset & Ref. &
\makecell{No. of \\ Samples} &
\makecell{No. of \\ Features} &
\makecell{Categorical \\ Features (\%)} &
\makecell{Missingness \\ (\%)} &
\makecell{Classes / \\ Targets} &
Task \\
\midrule
MultiOmics\_GS-BRCA & TCGA \citep{yang2025mlomics}& 671
& 15,200
& -& -& 5
& MultiC
\\
MultiOmics\_GS-COAD & TCGA \citep{yang2025mlomics}& 260
& 15,200
& -& -& 4
& MultiC
\\
MultiOmics\_GS-GBM & TCGA \citep{yang2025mlomics}& 244
& 14,640
& -& 0.05
& 5
& MultiC
\\
MultiOmics\_GS-LGG & TCGA \citep{yang2025mlomics}& 247
& 13,242
& -& 0.03
& 3
& MultiC
\\
MultiOmics\_GS-OV & TCGA \citep{yang2025mlomics}& 284& 13,501& -& 0.07& 4& MultiC\\
\bottomrule
\end{tabular}
}
\end{table}
\begin{table}[!htbp]
\centering
\caption{Transcriptomics dataset inventory. Each row reports the dataset source, number of samples and features, categorical feature fraction, missingness, class or target count, and prediction task.}
\label{tab:transcriptomics-datasets}
\resizebox{\textwidth}{!}{
\begin{tabular}{llrrrrrl}
\toprule
Dataset & Ref. &
\makecell{No. of \\ Samples} &
\makecell{No. of \\ Features} &
\makecell{Categorical \\ Features (\%)} &
\makecell{Missingness \\ (\%)} &
\makecell{Classes / \\ Targets} &
Task \\
\midrule
ALLAML & \citep{li2018feature}& 72
& 7,129
& -& -& -& Bin
\\
CLL\_SUB\_111 & \citep{li2018feature}& 111
& 11,340
& -& -& 3& MultiC
\\
GLIOMA & \citep{li2018feature}& 50
& 4,434
& -& -& 4& MultiC
\\
GLI\_85 & \citep{li2018feature}& 85
& 22,283
& -& -& -& Bin
\\
Prostate\_GE & \citep{li2018feature}& 102
& 5,966
& -& -& -& Bin
\\
SMK\_CAN\_187 & \citep{li2018feature}& 187
& 19,993
& -& -& -& Bin
\\
TOX\_171 & GEO-GSE1725 \citep{li2018feature}& 171
& 5,748
& -& -& 4& MultiC
\\
lung & \citep{li2018feature}& 203& 3,312& -& -& 5& MultiC\\
metabric\_grade & \citep{curtis2012metabric}& 1,892& 20,603& -& $<0.001$& 3& MultiC\\
\bottomrule
\end{tabular}
}
\end{table}

\FloatBarrier

\subsection{Dataset-Specific Unification}
\label{apx:dataset_unification}

Although each source required domain-specific handling, a shared unification protocol was followed wherever applicable. Across datasets, curation generally involved:

\begin{itemize}
    \item Standardizing ingestion formats and metadata.
    \item Defining explicit predictive tasks.
    \item Removing leakage-prone or redundant variables.
    \item Preserving missing values as NaN unless domain-specific imputation was justified.
    \item Explicitly tracking feature types (numerical, categorical, sequence-derived).
    \item Creating fixed-split or cross-validation partitions, using 70/15/15 train/validation/test proportions by default for fixed-split tasks and preserving dataset-specific splitting requirements.
\end{itemize}

Below we describe domain-specific curation procedures.

\subsubsection{Genomics}

Genomics datasets were constructed primarily from the gnomAD v3.1.2 callset combining 1000 Genomes Project (1KG) and Human Genome Diversity Project (HGDP) data \citep{chen2024gnomad,koenig2024harmonized, 10002015global,bergstrom2020insights}, supplemented by the TALENT DNA dataset \citep{liu2025talent,ye2024acloserlook}, and the coding-variant pathogenicity task \texttt{vep\_pathogenic\_coding} \citep{instadeep2024codingdataset}. The 1KG+HGDP data provide whole-genome genotype information together with harmonized metadata describing sample sex, population labels, super-population labels, and geographic origin.

Raw genotype data were initially processed using PLINK2 \citep{chang2015second}. This included conversion from Variant Call Format (VCF) to PLINK 2's genotype format (PGEN), handling of pseudo-autosomal regions (PARs) using the hg38 definitions, sex-label correction, and reproducible subsampling of single nucleotide polymorphisms (SNPs). Depending on the benchmark task, SNP subsets were selected either randomly or from consecutive genomic positions, allowing us to construct datasets with different feature dimensionalities while preserving reproducibility across runs.

Subsequent processing used snputils \citep{bonet2026snputils}. Genotypes were converted to dosages by summing allele counts, yielding numerical features encoded as 0, 1, 2, or NaN, arranged in a samples-by-features matrix. Samples were then aligned with the corresponding task labels, and samples without valid labels for a given task were removed.

Labels were constructed in a task-specific manner. For sex prediction, binary sex labels were obtained from the available 1KG and HGDP sample metadata, and only chromosome X variants were used as input features \cite{molina2026sex}. For ancestry prediction, labels were derived from the harmonized 1KG+HGDP metadata. Continental ancestry was formulated as a seven-way classification task over the super-population categories AFR, AMR, EAS, EUR, OCE, SAS, and WAS. Fine-grained ancestry prediction used subpopulation labels, such as GBR, CHB, or YRI, as multiclass targets. Geographic prediction was defined as a multivariate regression task in which each sample was assigned latitude and longitude coordinates from the harmonized metadata. Finally, PCA prediction was defined as a multivariate regression task using the first three ancestry principal components as targets. These coordinates were computed from a genome-wide set of informative SNPs across autosomal chromosomes 1--22.

From this unified genotype representation, we generated the following genomics benchmark datasets:

\begin{itemize}
    \item \textbf{Sexcheck datasets:} \texttt{sexcheck\_chrX\_50}, \texttt{sexcheck\_chrX\_500}, and \texttt{sexcheck\_chrX\_50k}. These datasets predict sample sex from 50, 500, or 50,000 initially sampled chromosome X SNPs, retaining 45, 441, and 17,118 features after preprocessing, respectively. For \texttt{sexcheck\_chrX\_50k}, we create three versions per individual by independently masking approximately 95\% of genotype entries. All versions of an individual remain in the same partition.
    
    \item \textbf{Continental ancestry datasets:} \texttt{continental\_1k}, \texttt{continental\_50k}, and \texttt{continental\_50k\_missing}. These datasets use chromosome 22 SNPs to predict continental superpopulation labels. The \texttt{continental\_50k\_missing} variant randomly masks approximately 60\% of genotype entries to evaluate robustness to incomplete genotype data. It retains one row per individual and the same train/validation/test assignments as \texttt{continental\_50k}.
    
    \item \textbf{Subpopulation ancestry datasets:} \texttt{subpop\_1000\_consec} and \texttt{subpop\_50k}. These datasets predict fine-grained population labels and differ in both feature dimensionality and SNP selection strategy, including a consecutive-SNP variant.
    
    \item \textbf{Geographic coordinate regression datasets:} \texttt{coords\_1k} and \texttt{coords\_50k}. These datasets use genotype features to predict each sample's geographic latitude and longitude.
    
    \item \textbf{PCA regression datasets:} \texttt{pca\_1k} and \texttt{pca\_50k}. These datasets use genotype features to predict the first three ancestry principal components.
    
    \item \textbf{TALENT DNA dataset:} \texttt{dna} task is three-class splice-site classification, included as an additional genomics task from the TALENT benchmark suite. Duplicate sequences were collapsed and a sequence with conflicting labels was excluded before splitting.
    \item \textbf{Variant pathogenicity dataset:} \texttt{vep\_pathogenic\_coding} classifies coding single-nucleotide variants (SNVs) as ClinVar pathogenic or common gnomAD variants using only the surrounding GRCh38 reference sequence and the alternative allele. Allele-frequency, consequence-annotation, and genomic-position fields are excluded because they define or reveal the label. Splits hold out whole chromosomes, using Chr1, Chr4, and Chr21 for testing and Chr5, Chr6, and Chr11 for validation.

\end{itemize}

This design intentionally covers both classical genotype classification and structured representation learning tasks. In particular, the genomics subset spans binary classification, multiclass ancestry prediction, and multivariate regression, while also varying the number of SNP features, the genomic selection strategy, and the degree of missingness.

\subsubsection{Clinical and Healthcare}

Clinical datasets were gathered from heterogeneous repositories and unified through task-specific but consistent preprocessing. 
Common curation steps across EHR datasets included:

\begin{itemize}
    \item Removal of leakage-prone variables (e.g. survival times, alternative outcomes).
    \item Harmonization of categorical clinical variables.
    \item Clinical grouping of diagnoses, such as International Classification of Diseases (ICD) categories.
    \item Explicit identification of categorical variables for downstream encoding.
    \item Subject-aware splitting when repeated patients were present.
\end{itemize}

Several datasets required substantial dataset-specific curation. For example:

\textbf{VitalDB} predicts in-hospital mortality using features available by the end of the operative/anesthesia case, including engineered indicators for devices and procedures, grouped diagnosis variables, missingness indicators, and duration-based features, with group-stratified splitting by patient. Hospital length of stay and ICU stay duration were excluded, and two impossible anesthesia durations were set to missing.

\textbf{MUSIC} maps cause-of-death outcomes into four classes, excluding 11 patients lost to follow-up and 20 heart-transplant patients with unresolved outcomes. Predictors comprise clinical covariates, excluding follow-up variables and baseline ECG/Holter measurements.

\textbf{IACCD} (Indwelling Arterial Catheter Clinical Dataset) is the separately released PhysioNet \texttt{mimic2-iaccd} dataset, available without credentialed access under the Open Data Commons Attribution License v1.0.\footnote{\url{https://physionet.org/content/mimic2-iaccd/1.0/}} We removed redundant mortality indicators, administrative variables, and the whole-stay blood-gas test count (\texttt{abg\_count}).

\textbf{Diabetes UCI} predicts 30-day readmission at discharge. We retained the complete earliest encounter per patient, grouped the primary ICD-9 diagnosis and age bands, dropped secondary and tertiary diagnoses, and excluded hospice/death-related discharges following Strack et al. \citep{strack2014impact}. Admission type, discharge disposition, and admission source were treated as categorical features. Unmeasured HbA1c and glucose results form an explicit ``Not Measured'' category.

\textbf{Heart Disease UCI} pools four source cohorts (Cleveland (USA), VA Long Beach (USA), Hungary, Switzerland) and binarizes disease severity. Zero cholesterol and resting blood pressure are treated as missing, and missing categorical values remain NaN for training-fitted imputation.

\textbf{Thyroid} restores the released UCI feature values before TALENT's per-file standardization and marks binary flags as categorical. \textbf{NHANES} uses the CDC/NCHS National Health and Nutrition Examination Survey 2013--2014 age-prediction subset (UCI-ID-887). Sex, physical activity, and diabetes responses are treated as categorical, with refused responses marked missing.

Clinical datasets from TALENT \citep{liu2025talent} and OpenML \citep{OpenML2025} retained their source feature representations, with task-specific curation. In \texttt{audiology}, the binary target distinguishes the source diagnosis \texttt{cochlear\_age} from all other diagnoses. Exact duplicate feature patterns were collapsed in \texttt{audiology}, \texttt{blood-transfusion-service}, and \texttt{maternal\_health\_risk}, and patterns with conflicting labels were excluded from the latter two. The same deduplication reduces \texttt{CDC\_Diabetes\_Health\_Indicators}, derived from the CDC Behavioral Risk Factor Surveillance System (BRFSS), from 253,680 to 226,342 records. In \texttt{pima\_diabetes}, zero glucose, blood pressure, skinfold thickness, insulin, and BMI values are treated as missing. The redundant monetary feature, equal to 250 times donation frequency, was removed from \texttt{blood-transfusion-service}. In \texttt{lymphography}, the binary task distinguishes metastases from malignant lymph, retaining 142 samples and excluding the normal and fibrosis cases. In \texttt{Cardiovascular-Disease-dataset}, duplicate feature patterns were collapsed, conflicting-label patterns were excluded, and outlier records violating fixed blood-pressure and anthropometric plausibility bounds were removed.

Two clinical tasks were obtained from PMLB \citep{olson2017pmlbalarge} and verified against their original UCI releases. \texttt{dermatology} predicts six erythemato-squamous diagnoses from clinical and histopathological findings after biopsy. The original numeric ages are used, preserving eight unknown ages as missing. \texttt{spectf} predicts normal versus abnormal cardiac single-photon emission computed tomography (SPECT) from 44 image summaries. We removed 82 duplicate copies, recovering the corrected 267-patient UCI cohort. We pooled its source training and test files for five-fold outer cross-validation with three-fold inner validation, as for \texttt{dermatology}.

\subsubsection{Drug Discovery and Molecular Properties}

Drug discovery and molecular-property datasets were primarily sourced from MoleculeNet \citep{wu2018moleculenet}, supplemented with additional tasks from OpenML \citep{OpenML2025} and TALENT \citep{liu2025talent,ye2024acloserlook}.

From MoleculeNet, we select QM8, QM9, SIDER, Tox21, HIV, and ESOL. These datasets are chosen to ensure coverage across the main molecular property categories defined in MoleculeNet, namely quantum mechanics, physical chemistry, biophysics, and physiology. QM8 and QM9 are selected over earlier benchmarks such as QM7 to cover larger molecular collections and a broader range of quantum-chemical properties \citep{wu2018moleculenet}. QM9 in particular covers more complex molecules and is widely adopted in molecular machine learning \citep{ramakrishnan2014quantum,jang_quantum_2025}. ESOL is selected as the representative physical chemistry regression task, as it is one of the most widely used molecular regression benchmarks and is consistently included across major evaluation frameworks such as MoleculeNet and OGB~\citep{hu_open_2021}. HIV is included as the representative biophysics single-task classification dataset, PCBA and MUV are discarded due to their excessive number of tasks and samples, and HIV is preferred over PDBbind and BACE for its larger sample size, given that small datasets are already represented by SIDER. Tox21 is chosen to include a toxicity classification dataset over ToxCast, as it is more widely adopted and was central to the landmark Tox21 Data Challenge \citep{tox21challenge} in machine learning for drug discovery. SIDER is selected as the representative adverse drug reaction dataset, containing approved drugs and their reported adverse drug reactions.

Except for ESOL and HIV, which contain a single prediction target, all remaining datasets consist of multiple labels. For the quantum chemistry targets, we include only E1CC2 and f1CC2, as CC2 corresponds to the reference level of theory used in the original dataset, while alternative computational methods are primarily included for comparison purposes. For electronic structure and thermodynamic targets, we select the energy gap between the highest occupied molecular orbital (HOMO) and lowest unoccupied molecular orbital (LUMO), and Gibbs free energy at 298 K. These targets capture distinct aspects of molecular behavior, including electronic excitation, reactivity, and thermodynamic stability, and were selected based on both physical interpretability and low redundancy, as assessed through correlation and clustering analysis. For Tox21, we select three tasks spanning both regulatory panels and a range of difficulty levels: NR-AhR, NR-ER and SR-MMP. For SIDER, we select two adverse-effect endpoints: gastrointestinal disorders and nervous system disorders.

For the QM8 and QM9 datasets from MoleculeNet \citep{wu2018moleculenet}, we generate random 70/15/15 train/validation/test splits. For the remaining datasets (SIDER, Tox21, HIV, and ESOL), we employ scaffold splitting \citep{wu2018moleculenet}, following a growing consensus in recent literature that scaffold splitting provides a more realistic and challenging evaluation setting by better testing model generalization across chemical structures \citep{cremer_equivariant_2023, huang_learning_2023, kim_merged_2021}. All datasets use a consistent 70/15/15 train/validation/test split. Regression targets for QM8, QM9, and ESOL are z-scored using training-split statistics.

Featurization is adapted to the nature of each task. For quantum chemistry datasets (QM8 and QM9), we use the Coulomb matrix representation \citep{montavon2012learning}, computed using the DeepChem package \citep{Ramsundar-et-al-2019}. Coulomb matrices explicitly encode nuclear charges and interatomic distances, whereas purely topological fingerprints such as extended-connectivity fingerprints (ECFP) omit explicit three-dimensional geometry \citep{wu2018moleculenet}. For the remaining datasets, we use binary Morgan fingerprints \citep{rogers_extended-connectivity_2010}, a widely adopted baseline in molecular machine learning due to their strong empirical performance across diverse property prediction tasks. Radius is set to 2, and initial fingerprint dimensionality is set to 1024 for datasets with more than 2,000 samples and 2048 otherwise.

The two tasks from TALENT and OpenML, \texttt{QSAR\_biodegradation} and \texttt{bioresponse}, provide precomputed molecular descriptors without molecular structures in the releases we use, so both use random 70/15/15 splits. \texttt{QSAR\_biodegradation} predicts ready biodegradability from 41 descriptors. Three duplicate descriptor rows with matching labels were collapsed before stratified splitting, leaving 1,051 molecules. \texttt{bioresponse} predicts a binary biological response for 3,751 molecules from molecular descriptors.

\subsubsection{Metabolomics}

Metabolomics datasets were assembled from clinical metabolomics studies in Metabolomics Workbench and MetaboLights \citep{mendez2019comparative,dlugas2025comparative}, together with volatile organic compound mass-spectrometry datasets from TabPFN-VOCS \citep{granitto2025use}. The clinical subset follows prior metabolomics benchmark selections \citep{mendez2019comparative,dlugas2025comparative} and includes the studies  \texttt{ST000369}, \texttt{ST000496}, \texttt{ST001000}, \texttt{ST001047}, \texttt{MTBLS136}, \texttt{MTBLS161}, and \texttt{MTBLS404}, formulated as study-specific binary classification tasks. For each study, the relevant analysis matrix was selected, sample identifiers were aligned with phenotype annotations, samples without valid labels were removed, and labels were encoded as 0 and 1. For \texttt{ST000496}, the paired pre- and post-treatment samples from 50 patients are grouped by patient in both outer and inner cross-validation, including the final-fit validation holdout, and patient identifiers are excluded from the features.

The TabPFN-VOCS subset comprises proton-transfer-reaction time-of-flight mass spectrometry (PTR-ToF-MS) measurements of chewing-gum formulations (\texttt{Gum1\_s}), roasted hazelnut paste (\texttt{Noc\_s}), and fungal cultures (\texttt{PT-Mush21}). The first two are regression tasks and the third is a classification task. Source IDs for \texttt{Gum1\_s} and \texttt{Noc\_s}, and biological-strain IDs for \texttt{PT-Mush21}, define five grouped outer folds that test each sample once. These datasets cover agrifood and fungal-culture settings, predicting component-1 concentration in chewing-gum formulations, hazelnut roasting time, and fungal species, respectively. Limited strain support requires two inner folds for \texttt{PT-Mush21}, while the regression tasks use three.

For the clinical metabolomics datasets, measurement features were transformed using \texttt{log1p} to reduce right-skewness, heteroscedasticity, and the large dynamic range typical of mass-spectrometry data. 
Because appropriate normalization depends on the measurement process, dataset characteristics, and downstream analysis \citep{sun2024pretreating,abram2022comprehensive}, we did not impose an additional uniform sample-wise normalization step across the clinical metabolomics datasets.

\subsubsection{Systems Biology}

The systems biology benchmark uses five cancer-cohort datasets from The Cancer Genome Atlas (TCGA), curated by MLOmics \citep{yang2025mlomics}, where multiple omics modalities are provided as separate feature matrices. For each dataset, modalities were first loaded independently and aligned using shared sample identifiers to ensure correct sample correspondence across data types. The aligned modalities were then concatenated into a single unified multi-omics feature table. For GS-LGG, methylation beta values were reconstructed from TCGA via UCSC Xena using patient barcodes and verified CpG identifiers, retaining the 4,705 source probes assayed on the HumanMethylation450 platform.\footnote{\url{https://tcga.xenahubs.net/download/TCGA.LGG.sampleMap/HumanMethylation450.json}} Features were coerced into a numeric format. All resulting features are treated as continuous numerical variables, and no additional feature engineering is applied beyond modality-level fusion.

\subsubsection{Single-Cell Omics}

Single-cell datasets were collected from Tahoe-100M \citep{zhang2026tahoe}, the mouse gastrulation atlas \citep{pijuansala2019gastrulation}, scUnified \citep{xu2025scunified}, and the Human Neural Organoid Cell Atlas (HNOCA) \citep{he2024integrated}. Appropriate normalization depends on the downstream analysis task \citep{heumos2023best}. In TabBioMed, we retained source-specific expression representations without imposing a common normalization or batch-correction pipeline. Gene filtering and selection used training cells only, and dataset-specific procedures are described below.

From Tahoe-100M, we generated cell-line classification datasets and a cell-cycle regression task. The included classification datasets, \texttt{tahoe\_cell\_name\_500}, \texttt{tahoe\_cell\_name\_1000}, and \texttt{tahoe\_cell\_name\_5000}, are three feature-width variants predicting cell-line identity using the top 500, 1,000, or 5,000 genes selected by Pearson residuals on training cells. Train, validation, and test splits were constructed using held-out plates, with random subsampling to obtain consistent split sizes. The \texttt{tahoe\_g2m\_score} dataset is a regression task predicting the G2/M cell-cycle score using the same feature-selection strategy. The \texttt{tahoe\_batch\_integration\_14plates} dataset predicts the plate of origin of each cell: 2,000 cells were randomly sampled from each of the 14 plates and randomly split 70/15/15 into training, validation, and test sets, and 5,000 genes were selected by Pearson residuals on the training cells.

From HNOCA, we included \texttt{hnoca\_cells30}, which predicts the 30 most frequent harmonized level-4 cell types. Publication--cell-line connected components were assigned to disjoint training, validation, and test sets, followed by class-stratified subsampling to 70,000/15,000/15,000 cells. We selected 22,000 genes by Pearson-residual variance using only the training cells' \texttt{counts\_lengthnorm} layer, and features retain the source log-transformed counts per million (CPM), $\log(1+\mathrm{CPM})$.

The \texttt{sc\_gastrulation\_stage} task predicts recorded embryonic collection day from the mouse gastrulation atlas \citep{pijuansala2019gastrulation}, using embryo-pool-disjoint splits and training-only gene selection.

From scUnified, \texttt{sc\_muris\_age} classifies recorded mouse age (3, 18, or 24 months), and \texttt{sc\_muris\_celltype} predicts nine released kidney cell-type annotations \citep{tabulamuris2020ageing}. The \texttt{sc\_muraro\_celltype} task predicts eight released human pancreas cell-type annotations, excluding three epsilon cells \citep{muraro2016pancreas}. These tasks use stratified cell-level splits, retaining kidney raw counts and released pancreas expression values after removing training-constant genes, and identifiers and other annotations are excluded from predictors.

\subsubsection{Proteomics}

Proteomics benchmarks combine quantitative mass-spectrometry datasets with protein sequence-based prediction tasks. The mass-spectrometry subset is derived from the CPTAC UCEC endometrial cancer cohort \citep{dou2020proteogenomic}, using the UMich protein-abundance and aliquot-to-patient mapping tables and MSSM \texttt{clinical\_Pan-cancer.May2022} annotations from Zenodo archive v41 \citep{cptac2023archive}. Tables were accessed using the \texttt{cptac} package version 1.5.14. Patient matching and histology selection yield 101 distinct tumor patients (80 endometrioid and 21 serous), with one proteomics profile per patient, for the binary \texttt{cptac\_histology} task. The general preprocessing pipeline removes constant or entirely missing columns, retaining 12,592 features with missing measurements preserved as NaN.

The PMLB/UCI \texttt{ecoli} task predicts five protein subcellular locations from 327 proteins. Protein identifiers are excluded from features.

From the collection \texttt{InstaDeepAI/true-cds-protein-tasks}  hosted on Hugging Face, we include three protein sequence regression tasks also represented in the PEER benchmark \citep{xu2022peer}: \texttt{true\_melting\_point}, \texttt{true\_betalactamase\_unique}, and \texttt{true\_fluorescence}. These tasks evaluate complementary aspects of protein sequence understanding, including thermostability, mutational fitness, and fluorescence. We used the train/validation/test splits provided by the benchmark collection. The source coding DNA sequences were first translated to amino-acid sequences, excluding sequences with internal stop codons. Melting point and beta-lactamase prediction were represented using mean-pooled ESM-2 embeddings \citep{lin2022language} from \texttt{esm2\_t12\_35M\_UR50D}, computed with the \texttt{protenc} Python package, whereas fluorescence prediction used TF-IDF features computed from amino-acid 3-mers.

From Proteinea \cite{proteinea_2023}, we include \texttt{proteinea\_solubility}, a binary protein-solubility task using the DeepSol dataset \citep{khurana2018deepsol}. We preserve the released train/validation/test partitions (62,478/6,942/2,001 sequences). Protein sequences were represented using normalized amino-acid 3-mer frequencies, yielding a fixed-length feature representation over all possible 3-mers from the standard amino-acid alphabet.

From AVIDa, we include binary binding prediction tasks for nanobodies, the variable domains of heavy-chain-only antibodies (VHHs), against hIL6 \citep{tsuruta2023avida,cognano_2025}, hTNFa \citep{cognano_htnfa}, and SARS-CoV-2 \citep{tsuruta2024sars}. The included datasets are \texttt{avida\_hil6\_onehot}, \texttt{avida\_htnfa\_onehot}, \texttt{avida\_htnfa\_3mer}, \texttt{avida\_htnfa\_esm2}, and \texttt{avida\_sarscov2\_3mer}. These variants intentionally cover multiple sequence encoding strategies: padded amino-acid token encodings, TF-IDF amino-acid 3-mer features, and mean-pooled ESM-2 embeddings. For paired sequence inputs, such as nanobody-antigen pairs, representations were computed separately for each sequence and concatenated.  Data-dependent encodings, such as TF-IDF vocabularies and padding lengths, are fitted on training rows only. For hIL6 and hTNFa, splits are VHH-family-disjoint: nanobody sequences are clustered with CD-HIT and whole families are assigned to the training, validation, or test set. For SARS-CoV-2, we keep the provided split by alpaca, draw validation from the training alpaca as whole VHH families, and remove training and validation pairs that exactly match a test nanobody-antigen pair.

\subsubsection{Transcriptomics}

Seven transcriptomics datasets were sourced from scikit-feature \citep{li2018feature}, alongside \texttt{TOX\_171} from GEO (GDS968/GSE1725), accessed through the same collection. These eight datasets were selected to represent both binary and multi-class classification tasks across diverse cancer and gene expression profiling studies. These raw inputs were converted into a unified structure where gene expression features ($X$) and corresponding labels ($y$) were extracted, with label encoding applied when necessary. For \texttt{TOX\_171}, the outer and inner cross-validation folds are grouped by donor.

The METABRIC task \citep{curtis2012metabric} predicts recorded histological grades 1--3 from primary breast-tumor expression in 1,892 distinct patients (169, 771, and 952 per grade). We exclude 88 expression-profiled patients without a recorded grade and retain 20,603 source expression measurements at their supplied log2 intensities. Evaluation uses patient-disjoint, grade-stratified data splits.

\begin{figure}[h]
    \centering
    \includegraphics[width=\linewidth]{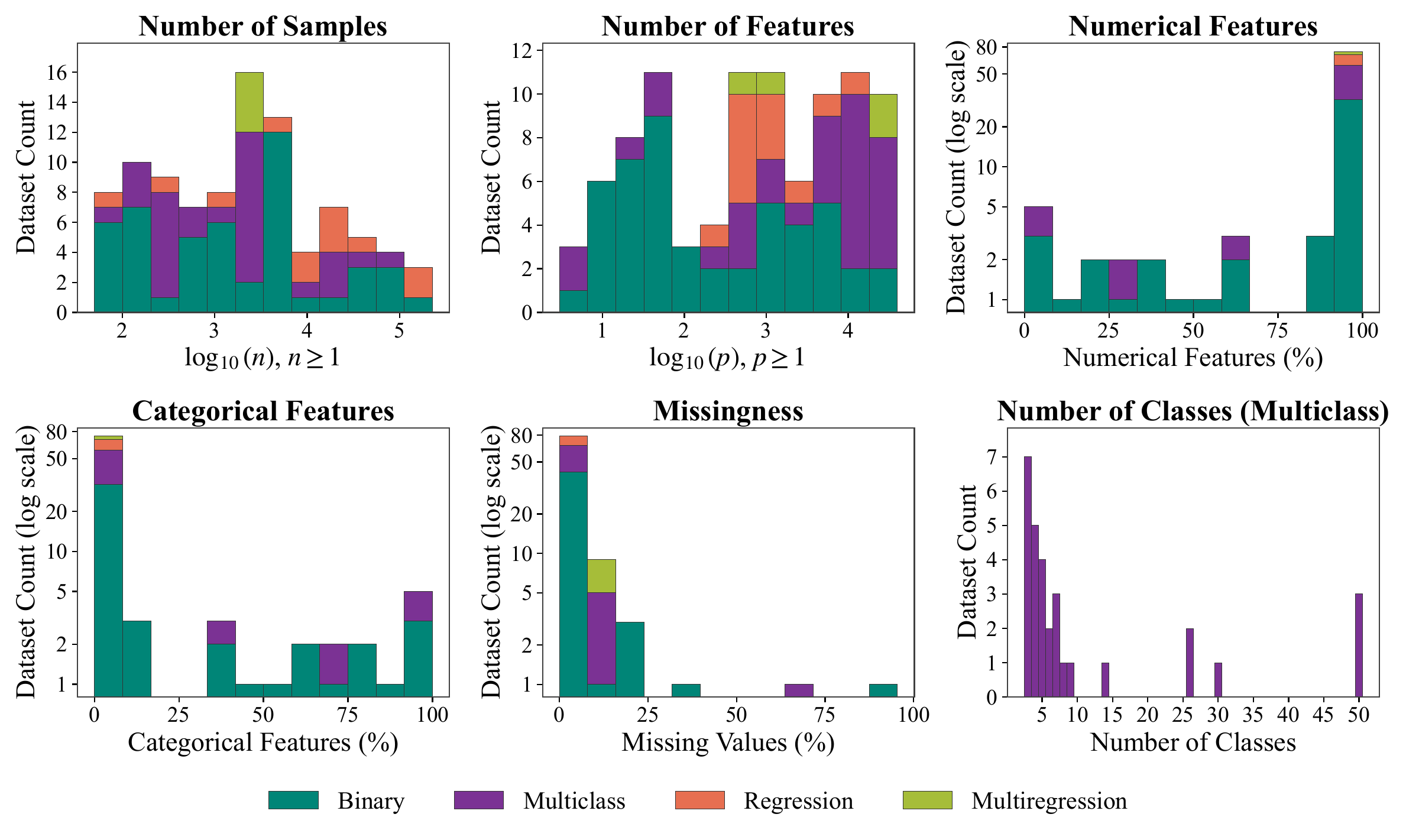}
    \caption{\textbf{Dataset Characteristics of the TabBioMed Benchmark.}
Distributions of key dataset properties, including number of samples, number of features, feature-type composition, missingness, and number of classes (for multiclass classification datasets only).}
    \label{fig:dataset_summary_2}
\end{figure}

The final benchmark contains curated datasets spanning eight biomedical domains and four task types as shown in Figure~\ref{fig:dataset_summary_1}. 
We provide additional visualizations of key dataset properties in TabBioMed, including sample size, feature dimensionality, feature-type composition, missingness, and class distribution in Figure~\ref{fig:dataset_summary_2}.

\section{Model Implementation}
\label{sec:model_implementation}

The modeling pipeline follows a unified experimental framework designed to ensure consistency among 23 models across four families ranging from classical machine learning methods to modern tabular foundation models and deep learning architectures (see Table \ref{tab:tabbiomed_model_suite}). The entire process is organized into three main stages: dataset preprocessing, training with HPO, and final refitting with evaluation. While a shared pipeline governs all models, each model class selectively overrides components of preprocessing, training behavior, and refitting logic depending on its inductive biases and internal learning mechanisms. This design allows fair comparison while respecting model-specific requirements such as native categorical handling, internal normalization strategies, or validation-dependent optimization.

\begin{table}[h]
    \centering
        \caption{\textbf{TabBioMed Model Suite.} Models evaluated in the TabBioMed benchmark, grouped by model family. An asterisk indicates models that are not evaluated on regression tasks because they are unsupported in that setting.}
    \begin{tabular}{llll}\toprule
         Baselines&  Tree-based&  Neural Networks& Foundation Models\\\midrule
         LinearModel&  DecisionTree&  MLP & TabPFNv2.5 \citep{grinsztajn2025tabpfn}\\
         LassoModel&  RandomForest \citep{breiman2001random}&  RealMLP \citep{holzmuller2025realmlp}& TabPFN-Wide* \citep{kolberg2025tabpfnwide}\\
         RidgeModel&  XGBoost \citep{chen2016xgboost}&  TabM \citep{gorishniy2024tabm}& RealTabPFNv2.5 \citep{grinsztajn2025tabpfn}\\
         ElasticNet&  CatBoost \citep{prokhorenkova2018catboost}&  & TabICLv2 \citep{qu2026tabiclv2}\\
         KNN&  LightGBM \citep{ke2017lightgbm}&  & TabDPT \citep{ma2024tabdpt}\\
         SVM&  xRFM \citep{beaglehole2026xrfm}&  & HyperFast* \citep{bonet2024hyperfast}\\ 
 NeoLsSVM& & &iLTM \citep{bonet2025iltm}\\ \bottomrule
    \end{tabular}
    \label{tab:tabbiomed_model_suite}
\end{table}

\subsection{Preprocessing}
\label{apx:model_impl_preprocessing}

A unified preprocessing pipeline is applied to all datasets prior to training, where the goal is to standardize data handling before model-specific overrides are applied.
At a global level, preprocessing consists of three steps:
Missing values in numerical features are handled only when required by the model. By default, numerical imputation uses K-Nearest Neighbors (KNN) imputation with five neighbors, fitted on the training split and applied to all splits. The default encoding for nonbinary categorical features is one-hot encoding, with model-specific alternatives described below. Scaling is fitted at training time and applied only to non-categorical, non-binary features using either StandardScaler or QuantileTransformer depending on model configuration.

{\textbf{Model-specific Preprocessing.}}
\begin{itemize}
    \item All \textbf{Baseline} models listed in Table \ref{tab:tabbiomed_model_suite}, along with \textbf{MLP}, rely on the full preprocessing pipeline. 
    \item \textbf{DecisionTree} uses categorical encoding without feature scaling or missing-value imputation. \textbf{Random Forest} uses categorical encoding and native numerical missing-value handling without feature scaling.
    \item \textbf{RealMLP} uses mean imputation with missingness indicators, while categorical encoding, robust scaling, and smooth clipping are handled internally by the model.
    \item \textbf{xRFM} uses mean imputation with missingness indicators followed by one-hot encoding of categorical features and standardization of numerical features.
    \item Gradient-boosted decision tree (GBDT) models (\textbf{XGBoost}, \textbf{LightGBM}, and \textbf{CatBoost}) do not use standard preprocessing. Missing values are handled natively without imputation. For categorical variables, XGBoost requires integer encoding, LightGBM uses internally supported categorical indices, and CatBoost directly consumes raw string categorical features. No feature scaling is applied.
    \item Tabular foundation models including \textbf{TabPFNv2.5}, \textbf{TabPFN-Wide}, \textbf{RealTabPFNv2.5}, \textbf{TabICLv2}, \textbf{TabDPT}, \textbf{HyperFast} and \textbf{iLTM} generally rely on internal preprocessing, with exceptions such as label encoding of categorical features for TabDPT. Missing values are handled internally, categorical features follow model-specific encoding procedures, and external feature scaling is disabled in favor of internal normalization mechanisms. 
    \item The hybrid neural model \textbf{TabM} partially modifies the standard pipeline. Missing values are imputed using mean imputation, and categorical variables are encoded using label encoding instead of one-hot encoding. Following the official TabM recipe, non-binary numerical features are transformed with a \texttt{QuantileTransformer} fitted on the training split, and the benchmark's generic scaler is disabled to avoid a redundant second transformation.
\end{itemize}

\subsection{Training and Hyperparameter Optimization Details}
\label{apx:model_impl_train_and_hpo}

Training is orchestrated through a unified experiment runner that supports both default baselines and hyperparameter optimization. The pipeline distinguishes between three execution modes: default baseline evaluation, HPO trials, and final test evaluation of the validation-selected configuration.

For final test evaluation of the default baseline, models are initialized using their predefined hyperparameters and trained either on the training split alone or on a merged training-validation set. The decision depends on whether a model uses validation data during fitting. Models without internal validation mechanisms are trained on the combined train-validation split to maximize data usage. In contrast, models that rely on early stopping or internal validation strictly train on the training set while reserving the validation split for monitoring.

During hyperparameter optimization, each trial executes a full training cycle but restricts evaluation to the validation set. Importantly, test data are never used during HPO. The training configuration during HPO explicitly disables train-validation merging to ensure strict separation between training and evaluation. Configurations are compared using the same validation data. For validation scoring, the default configuration is fitted on the training partition without train-validation merging and evaluated on the same validation data as the sampled configurations, with ties resolved in favor of the default.

Experiments are run using a Slurm-based pipeline. Each run is allocated up to 10 CPU cores, 128 GB RAM, and one NVIDIA RTX A5500 GPU when applicable (primarily for neural and foundation models, while tree-based models use CPU unless GPU execution is supported). Each Slurm job has a 24-hour wall-clock limit. Model selection, including validation evaluation of the default configuration, has a budget of six hours per fixed-split search or two hours per outer cross-validation fold. Standalone Default baseline evaluations and final refits (outside the HPO search) may exceed this selection budget, subject to the remaining Slurm job time.

Trial histories and progress reports are saved incrementally for reproducibility and fault tolerance, and completed evaluations retain test predictions. We record training and inference times as wall-clock measurements for both default and tuned models, along with evaluation metadata. Failed or resource-exceeded runs are retained as missing evaluations. In downstream analysis, we report complete-case results and analyses where missing entries are imputed using the default Random Forest result on the same dataset.

\textbf{Tree-based} models exhibit more diverse behavior. \textbf{Random Forest} operates without validation dependency and follows standard training procedures. We ran \textbf{CatBoost}, \textbf{LightGBM}, and \textbf{XGBoost} with up to 10{,}000 boosting iterations together with early stopping on the validation set, following the implementations used in AutoGluon and TabArena \citep{erickson2025tabarena}. \textbf{xRFM} is a tree-structured kernel method, rather than a conventional tree ensemble. It combines tree-based partitioning with kernel-based leaf predictors and uses validation-based early stopping. For family-level analyses, we group xRFM with tree-based models to retain four broad categories without introducing a separate category for a single method. These validation-dependent models therefore bypass train-validation merging entirely across all stages.

\textbf{MLP-based architectures}, including \textbf{MLP}, \textbf{RealMLP}, \textbf{TabM}, and also \textbf{iLTM} (for its generated target network, not for the meta-learned hypernetwork), follow validation-dependent optimization regimes. These models typically employ epoch-based training loops with validation monitoring for early stopping or performance tracking. For instance, \textbf{MLP} uses a custom PyTorch training loop with validation-based convergence control, while \textbf{TabM} uses internal early stopping. For these models, both hyperparameter optimization and final refit phases keep training and validation partitions separate. 

\textbf{TabM} removes numerical features with fewer than two unique values in the training partition before applying its quantile transformation and fitting the network. The same feature selection is applied to validation and test data, avoiding degenerate constant-feature embeddings.

\textbf{In-context learning models} introduce different training dynamics. The TabPFN-family models evaluated here (\textbf{TabPFNv2.5}, \textbf{TabPFN-Wide} and \textbf{RealTabPFNv2.5}) do not undergo gradient-based training during benchmark evaluation. Instead, they perform inference using pretrained models conditioned on the input context. Consequently, their ``training'' phase is effectively a preprocessing and inference setup step, although Optuna still tunes inference configuration parameters. \textbf{TabICLv2} and \textbf{TabDPT} similarly use in-context learning, with training reduced to configuring inference behavior rather than parameter optimization. For these models, validation data guide configuration selection without updating the pretrained weights. Training and validation data are then merged into the context pool for final test evaluation.

\subsection{Hyperparameter Optimization}
\label{apx:hpo_search_space}

In this section, we describe the hyperparameter optimization details used for tuning the models evaluated in the TabBioMed benchmark. The majority of the search spaces are adopted or adapted from \textbf{TabArena} \cite{erickson2025tabarena}. This includes the search spaces for \textbf{Random Forest}, \textbf{XGBoost}, \textbf{LightGBM}, \textbf{CatBoost}, \textbf{KNN}, \textbf{RealMLP}, \textbf{xRFM}, \textbf{TabM}, \textbf{TabDPT}, \textbf{TabICLv2}, \textbf{TabPFNv2.5} and \textbf{RealTabPFNv2.5}, including search spaces added to the official TabArena GitHub repository after its original publication. LinearModel and NeoLsSVM are evaluated using their default settings without external HPO. LinearModel is unpenalized, while Lasso, Ridge, and ElasticNet cover regularized linear models. NeoLsSVM automatically selects its regularization internally.

We use the following model-specific search spaces and adaptations:

\begin{itemize} 
\item \textbf{Baseline Models:} For \textbf{Lasso}, \textbf{Ridge} and \textbf{ElasticNet}, we defined search spaces covering regularization strengths and scaling options. For \textbf{SVM} and \textbf{DecisionTree}, the search spaces were adapted from \textbf{TabZilla} \citep{mcelfresh2023neural}, with additional DecisionTree parameters for the splitting strategy and minimum split and leaf sizes.

\item \textbf{Other Foundation Models:} For the remaining foundation models, we use the following search spaces and configurations:
\begin{itemize}
    \item For \textbf{HyperFast} we adopted the recommended search space, and for \textbf{iLTM} we evaluate up to 25 recommended configurations from the official GitHub repository.
    \item \textbf{TabPFN-Wide}: We aimed to closely match TabArena's TabPFNv2.5 search space while adapting it to the specific version of the \texttt{tabpfn} package required by TabPFN-Wide's implementation.
\end{itemize}
\item \textbf{Search-space adaptations:} For \textbf{XGBoost}, we lowered the minimum \texttt{min\_child\_weight} from $10^{-3}$ to $10^{-5}$ and additionally tuned \texttt{scale\_pos\_weight} log-uniformly over $[0.1, 10]$. For \textbf{TabM}, we implemented TabArena's existing mixtures for \texttt{weight\_decay} and \texttt{dropout} using conditional parameters, preserving their sampling distributions.
\end{itemize}

For models with sampled search spaces, we perform HPO using Optuna \citep{optuna_2019} with random search.
In Tables~\ref{space_lasso} through~\ref{space_tabicl}, Choice denotes uniform categorical sampling unless probabilities are specified. UniformInt and LogUniformInt denote integer-valued distributions with inclusive bounds.

\FloatBarrier
\begin{table}[!htbp]
\centering
\caption{Hyperparameter search space for LassoModel.}
\label{space_lasso}
\begin{tabular}{ll}
\toprule
Hyperparameter & \multicolumn{1}{l}{Space} \\
\midrule
\texttt{alpha} & \multicolumn{1}{l}{LogUniform([1e-06, 1.0])} \\
\texttt{fit\_intercept} & \multicolumn{1}{l}{Choice([True, False])} \\
\texttt{class\_weight} & \multicolumn{1}{l}{Choice(["balanced", None])} \\
\texttt{scaler} & \multicolumn{1}{l}{Choice(["standard", "quantile"])} \\
\bottomrule
\end{tabular}
\end{table}

\begin{table}[!htbp]
\centering
\caption{Hyperparameter search space for RidgeModel.}
\label{space_ridge}
\begin{tabular}{ll}
\toprule
Hyperparameter & \multicolumn{1}{l}{Space} \\
\midrule
\texttt{alpha} & \multicolumn{1}{l}{LogUniform([1e-06, 1.0])} \\
\texttt{fit\_intercept} & \multicolumn{1}{l}{Choice([True, False])} \\
\texttt{class\_weight} & \multicolumn{1}{l}{Choice(["balanced", None])} \\
\texttt{scaler} & \multicolumn{1}{l}{Choice(["standard", "quantile"])} \\
\bottomrule
\end{tabular}
\end{table}

\begin{table}[!htbp]
\centering
\caption{Hyperparameter search space for ElasticNet.}
\label{space_elasticnet}
\begin{tabular}{ll}
\toprule
Hyperparameter & \multicolumn{1}{l}{Space} \\
\midrule
\texttt{alpha} & \multicolumn{1}{l}{LogUniform([1e-08, 1.0])} \\
\texttt{l1\_ratio} & \multicolumn{1}{l}{Uniform([0.0, 1.0])} \\
\texttt{scaler} & \multicolumn{1}{l}{Choice(["standard", "quantile"])} \\
\bottomrule
\end{tabular}
\end{table}

\begin{table}[!htbp]
\centering
\caption{Hyperparameter search space for SVM.}
\label{space_svm}
\begin{tabular}{ll}
\toprule
Hyperparameter & \multicolumn{1}{l}{Space} \\
\midrule
\texttt{C} & \multicolumn{1}{l}{LogUniform([1e-08, 10000000000.0])} \\
\texttt{scaler} & \multicolumn{1}{l}{Choice(["standard", "quantile"])} \\
\bottomrule
\end{tabular}
\end{table}

\begin{table}[!htbp]
\centering
\caption{Hyperparameter search space for KNN. Candidate neighbor counts greater than the training partition size are excluded. For cross-validation, feasibility is checked against the smallest inner training fold.}
\label{space_knn}
\resizebox{\textwidth}{!}{%
\begin{tabular}{ll}
\toprule
Hyperparameter & \multicolumn{1}{l}{Space} \\
\midrule
\texttt{n\_neighbors} & \multicolumn{1}{l}{Choice([1, 2, 3, 4, 5, 6, 7, 8, 9, 10, 11, 13, 15, 20, 30, 40, 50, 100, 200, 300, 400, 500])} \\
\texttt{weights} & \multicolumn{1}{l}{Choice(["distance", "uniform"])} \\
\texttt{p} & \multicolumn{1}{l}{Choice([1, 1.5, 2])} \\
\texttt{scaler} & \multicolumn{1}{l}{Choice(["standard", "quantile"])} \\
\bottomrule
\end{tabular}
}
\end{table}

\begin{table}[!htbp]
\centering
\caption{Hyperparameter search space for DecisionTree.}
\label{space_decisiontree}
\begin{tabular}{ll}
\toprule
Hyperparameter & \multicolumn{1}{l}{Space} \\
\midrule
\texttt{max\_depth} & \multicolumn{1}{l}{UniformInt([2, 12])} \\
\texttt{splitter} & \multicolumn{1}{l}{Choice(["best", "random"])} \\
\texttt{min\_samples\_split} & \multicolumn{1}{l}{UniformInt([2, 10])} \\
\texttt{min\_samples\_leaf} & \multicolumn{1}{l}{UniformInt([1, 10])} \\
\bottomrule
\end{tabular}
\end{table}

\begin{table}[!htbp]
\centering
\caption{Hyperparameter search space for Random Forest.}
\label{space_rf}
\resizebox{\textwidth}{!}{%
\begin{tabular}{ll}
\toprule
Hyperparameter & Space \\
\midrule
\texttt{max\_depth} & Choice([None, 4, 6, 8, 12, 16]) \\
\texttt{n\_estimators} & UniformInt([100, 300]) \\
\texttt{max\_features} & Choice(["sqrt", "log2", 0.2, 0.33, 0.5, 0.67, 0.8, 1.0]) \\
\texttt{max\_samples} & Choice([None, 0.5, 0.6, 0.7, 0.8, 0.9, 1.0]) \\
\texttt{min\_samples\_split} & UniformInt([2, 10]) \\
\texttt{min\_samples\_leaf} & UniformInt([1, 20]) \\
\texttt{min\_impurity\_decrease} & Uniform([0.0, 1e-4]) \\
\texttt{class\_weight} & Choice([None, "balanced", "balanced\_subsample"]) \\
\bottomrule
\end{tabular}
}
\end{table}

\begin{table}[!htbp]
\centering
\caption{Hyperparameter search space for CatBoost. For GPU multi-target regression, \texttt{grow\_policy} is restricted to \texttt{SymmetricTree}.}
\label{space_catboost}
\begin{tabular}{ll}
\toprule
Hyperparameter & Space \\
\midrule
\texttt{iterations} & 10{,}000 (maximum; early stopping) \\
\texttt{learning\_rate} & LogUniform([0.01, 0.08]) \\
\texttt{bootstrap\_type} & Choice(["Bernoulli"]) \\
\texttt{subsample} & Uniform([0.7, 1.0]) \\
\texttt{grow\_policy} & Choice(["SymmetricTree", "Depthwise"]) \\
\texttt{depth} & UniformInt([3, 7]) \\
\texttt{l2\_leaf\_reg} & LogUniform([1.0, 30.0]) \\
\texttt{leaf\_estimation\_iterations} & LogUniformInt([1, 5]) \\
\texttt{min\_data\_in\_leaf} & LogUniformInt([1, 64]) \\
\texttt{random\_strength} & LogUniform([0.5, 20.0]) \\
\texttt{one\_hot\_max\_size} & LogUniformInt([8, 100]) \\
\texttt{model\_size\_reg} & LogUniform([0.1, 2.0]) \\
\texttt{max\_ctr\_complexity} & UniformInt([2, 5]) \\
\texttt{boosting\_type} & Choice(["Plain"]) \\
\texttt{max\_bin} & Choice([254]) \\
\texttt{auto\_class\_weights} & Choice([None, "Balanced", "SqrtBalanced"]) \\
\bottomrule
\end{tabular}
\end{table}

\begin{table}[!htbp]
\centering
\caption{Hyperparameter search space for XGBoost.}
\label{space_xgboost}
\begin{tabular}{ll}
\toprule
Hyperparameter & \multicolumn{1}{l}{Space} \\
\midrule
\texttt{n\_estimators} & \multicolumn{1}{l}{10{,}000 (maximum; early stopping)} \\
\texttt{max\_depth} & \multicolumn{1}{l}{LogUniformInt([4, 10])} \\
\texttt{learning\_rate} & \multicolumn{1}{l}{LogUniform([0.005, 0.1])} \\
\texttt{min\_child\_weight} & \multicolumn{1}{l}{LogUniform([1e-05, 5.0])} \\
\texttt{subsample} & \multicolumn{1}{l}{Uniform([0.6, 1.0])} \\
\texttt{colsample\_bylevel} & \multicolumn{1}{l}{Uniform([0.6, 1.0])} \\
\texttt{colsample\_bynode} & \multicolumn{1}{l}{Uniform([0.6, 1.0])} \\
\texttt{alpha} & \multicolumn{1}{l}{Uniform([0.0001, 5.0])} \\
\texttt{lambda} & \multicolumn{1}{l}{Uniform([0.0001, 5.0])} \\
\texttt{grow\_policy} & \multicolumn{1}{l}{Choice(["depthwise", "lossguide"])} \\
\texttt{max\_cat\_to\_onehot} & \multicolumn{1}{l}{LogUniformInt([8, 100])} \\
\texttt{max\_leaves} & \multicolumn{1}{l}{LogUniformInt([8, 1024])} \\
\texttt{enable\_categorical} & \multicolumn{1}{l}{Choice([True])} \\
\texttt{scale\_pos\_weight} & \multicolumn{1}{l}{LogUniform([0.1, 10])} \\
\bottomrule
\end{tabular}
\end{table}

\begin{table}[!htbp]
\centering
\caption{Hyperparameter search space for LightGBM.}
\label{space_lightgbm}
\begin{tabular}{ll}
\toprule
Hyperparameter & \multicolumn{1}{l}{Space} \\
\midrule
\texttt{n\_estimators} & \multicolumn{1}{l}{10{,}000 (maximum; early stopping)} \\
\texttt{num\_leaves} & \multicolumn{1}{l}{LogUniformInt([2, 200])} \\
\texttt{learning\_rate} & \multicolumn{1}{l}{LogUniform([0.005, 0.1])} \\
\texttt{feature\_fraction} & \multicolumn{1}{l}{Uniform([0.4, 1.0])} \\
\texttt{bagging\_fraction} & \multicolumn{1}{l}{Uniform([0.7, 1.0])} \\
\texttt{bagging\_freq} & \multicolumn{1}{l}{Choice([1])} \\
\texttt{min\_data\_in\_leaf} & \multicolumn{1}{l}{LogUniformInt([1, 64])} \\
\texttt{extra\_trees} & \multicolumn{1}{l}{Choice([False, True])} \\
\texttt{min\_data\_per\_group} & \multicolumn{1}{l}{LogUniformInt([1, 100])} \\
\texttt{cat\_l2} & \multicolumn{1}{l}{LogUniform([0.005, 2.0])} \\
\texttt{cat\_smooth} & \multicolumn{1}{l}{LogUniform([0.001, 100])} \\
\texttt{max\_cat\_to\_onehot} & \multicolumn{1}{l}{LogUniformInt([8, 100])} \\
\texttt{lambda\_l1} & \multicolumn{1}{l}{Uniform([0.0001, 1.0])} \\
\texttt{lambda\_l2} & \multicolumn{1}{l}{Uniform([0.0001, 2.0])} \\
\texttt{class\_weight} & \multicolumn{1}{l}{Choice([None, "balanced"])} \\
\bottomrule
\end{tabular}
\end{table}

\begin{table}[!htbp]
\centering
\caption{Hyperparameter search space for xRFM.}
\label{space_xrfm}
\begin{tabular}{ll}
\toprule
Hyperparameter & \multicolumn{1}{l}{Space} \\
\midrule
\texttt{bandwidth} & \multicolumn{1}{l}{LogUniform([0.5, 200.0])} \\
\texttt{kernel} & \multicolumn{1}{l}{Choice(["lpq\_kermac", "l2"], probs=[0.8, 0.2])} \\
\texttt{exponent} & \multicolumn{1}{l}{Uniform([0.7, 1.4])} \\
\texttt{p\_interp} & \multicolumn{1}{l}{Uniform([0.0, 0.8])} \\
\texttt{reg} & \multicolumn{1}{l}{LogUniform([1e-6, 1e1])} \\
\texttt{diag} & \multicolumn{1}{l}{Choice([False, True])} \\
\texttt{bandwidth\_mode} & \multicolumn{1}{l}{Choice(["constant"])} \\
\texttt{standardize\_cats} & \multicolumn{1}{l}{Choice([False])} \\
\texttt{early\_stop\_multiplier} & \multicolumn{1}{l}{Choice([1.1])} \\
\bottomrule
\end{tabular}
\end{table}

\begin{table}[!htbp]
\centering
\caption{Hyperparameter search space for MLP.}
\label{space_mlp}
\begin{tabular}{ll}
\toprule
Hyperparameter & \multicolumn{1}{l}{Space} \\
\midrule
\texttt{hidden\_dim} & \multicolumn{1}{l}{UniformInt([8, 256])} \\
\texttt{activation} & \multicolumn{1}{l}{Choice(["relu", "elu"])} \\
\texttt{n\_layers} & \multicolumn{1}{l}{UniformInt([1, 5])} \\
\texttt{learning\_rate} & \multicolumn{1}{l}{LogUniform([0.0005, 0.001])} \\
\texttt{weight\_decay} & \multicolumn{1}{l}{LogUniform([1e-12, 0.1])} \\
\texttt{dropout\_rate} & \multicolumn{1}{l}{Uniform([0.0, 0.4])} \\
\texttt{use\_batch\_norm} & \multicolumn{1}{l}{Choice([True, False])} \\
\texttt{scaler} & \multicolumn{1}{l}{Choice(["standard", "quantile"])} \\
\bottomrule
\end{tabular}
\end{table}

\begin{table}[!htbp]
\centering
\caption{Hyperparameter search space for TabM.}
\label{space_tabm}
\begin{tabular}{lll}
\toprule
Hyperparameter & \multicolumn{2}{l}{Space} \\
\midrule
\texttt{arch\_type} & \multicolumn{2}{l}{Choice(["tabm-mini"])} \\
\texttt{tabm\_k} & \multicolumn{2}{l}{Choice([32])} \\
\texttt{share\_training\_batches} & \multicolumn{2}{l}{Choice([False])} \\
\texttt{batch\_size} & \multicolumn{2}{l}{Choice(["auto"])} \\
\texttt{patience} & \multicolumn{2}{l}{Choice([16])} \\
\texttt{lr} & \multicolumn{2}{l}{LogUniform([1e-4, 3e-3])} \\
\texttt{weight\_decay\_choice} & \multicolumn{2}{l}{Choice([0.0, 1.0])} \\
\texttt{gradient\_clipping\_norm} & \multicolumn{2}{l}{Choice([1.0])} \\
\texttt{amp} & \multicolumn{2}{l}{Choice([False])} \\
\texttt{compile\_model} & \multicolumn{2}{l}{Choice([False])} \\
\texttt{n\_blocks} & \multicolumn{2}{l}{Choice([2, 3, 4, 5])} \\
\texttt{d\_block} & \multicolumn{2}{l}{
Choice(128:16:1024) \textit{(step size = 16)}
} \\
\texttt{dropout\_choice} & \multicolumn{2}{l}{Choice([0.0, 1.0])} \\
\texttt{num\_emb\_type} & \multicolumn{2}{l}{Choice(["pwl"])} \\
\texttt{d\_embedding} & \multicolumn{2}{l}{Choice([8, 12, 16, 20, 24, 28, 32])} \\
\texttt{num\_emb\_n\_bins} & \multicolumn{2}{l}{UniformInt([2, 128])} \\
\texttt{eval\_batch\_size} & \multicolumn{2}{l}{Choice([64])} \\
\texttt{bool\_to\_cat} & \multicolumn{2}{l}{Choice([True])} \\
\texttt{average\_logits} & \multicolumn{2}{l}{Choice([False])} \\
\midrule
 & 0.0 & 1.0 \\
\cmidrule{2-3}
\texttt{weight\_decay} & 0.0 & LogUniform([1e-4, 1e-1]) \\
\midrule
 & 0.0 & 1.0 \\
\cmidrule{2-3}
\texttt{dropout} & 0.0 & Uniform([0.0, 0.5]) \\
\bottomrule
\end{tabular}
\end{table}

\begin{table}[!htbp]
\centering
\caption{Hyperparameter search space for RealMLP.}
\label{space_realmlp}
\begin{tabular}{lll}
\toprule
Hyperparameter & \multicolumn{2}{l}{Space} \\
\midrule
\texttt{config\_type} & \multicolumn{2}{l}{Choice(["default", "large"])} \\
\texttt{n\_hidden\_layers} & \multicolumn{2}{l}{UniformInt([2, 4])} \\
\texttt{hidden\_sizes} & \multicolumn{2}{l}{rectangular} \\
\texttt{hidden\_width} & \multicolumn{2}{l}{Choice([256, 384, 512])} \\
\texttt{p\_drop} & \multicolumn{2}{l}{Uniform([0.0, 0.5])} \\
\texttt{act} & \multicolumn{2}{l}{mish} \\
\texttt{plr\_sigma} & \multicolumn{2}{l}{LogUniform([0.01, 50.0])} \\
\texttt{plr\_lr\_factor} & \multicolumn{2}{l}{LogUniform([0.05, 0.3])} \\
\texttt{scale\_lr\_factor} & \multicolumn{2}{l}{LogUniform([2.0, 10.0])} \\
\texttt{first\_layer\_lr\_factor} & \multicolumn{2}{l}{LogUniform([0.3, 1.5])} \\
\texttt{sq\_mom} & \multicolumn{2}{l}{1 $-$ LogUniform([0.005, 0.05])} \\
\texttt{lr} & \multicolumn{2}{l}{LogUniform([0.02, 0.3])} \\
\texttt{wd} & \multicolumn{2}{l}{LogUniform([0.001, 0.05])} \\
\texttt{ls\_eps} & \multicolumn{2}{l}{LogUniform([0.005, 0.1])} \\
\texttt{ls\_eps\_sched} & \multicolumn{2}{l}{coslog4} \\
\texttt{use\_ls} & \multicolumn{2}{l}{Choice([False, True])} \\
\texttt{p\_drop\_sched} & \multicolumn{2}{l}{Choice(["flat\_cos"])} \\
\texttt{max\_one\_hot\_cat\_size} & \multicolumn{2}{l}{$\lfloor$ LogUniform
([4.0, 33.0]) $\rfloor$} \\
\texttt{embedding\_size} & \multicolumn{2}{l}{Choice([4, 8, 16])} \\
\texttt{n\_ens} & \multicolumn{2}{l}{8} \\
\texttt{early\_stopping\_additive\_patience} & \multicolumn{2}{l}{40} \\
\texttt{early\_stopping\_multiplicative\_patience} & \multicolumn{2}{l}{3} \\
\midrule
\texttt{config\_type} & Default & Large \\
\cmidrule{2-3}
\texttt{plr\_hidden\_1} & 16 & Choice([8, 16, 32, 64]) \\
\texttt{plr\_hidden\_2} & 4 & Choice([8, 16, 32, 64]) \\
\texttt{n\_epochs} & 256 & Choice([256, 512]) \\
\texttt{use\_early\_stopping} & False & True \\
\bottomrule
\end{tabular}
\end{table}

\begin{table}[!htbp]
\centering
\caption{Hyperparameter search space for HyperFast. The \texttt{nn\_bias\_scale} range applies only when \texttt{nn\_bias=True}. Fine-tuning settings apply only when \texttt{optimization="ensemble\_optimize"}.}
\label{space_hyperfast}
\begin{tabular}{ll}
\toprule
Hyperparameter & Space \\
\midrule
\texttt{n\_ensemble} & Choice([4, 8, 16, 32]) \\
\texttt{batch\_size} & Choice([1024, 2048]) \\
\texttt{nn\_bias} & Choice([True, False]) \\
\texttt{nn\_bias\_scale} & LogUniform([0.01, 1.0]) \\
\texttt{stratify\_sampling} & Choice([True, False]) \\
\texttt{optimization} & \makecell[l]{Choice(["ensemble\_optimize", None],\\ probs=[0.9, 0.1])} \\
\midrule
\multicolumn{2}{l}{\textit{Fine-tuning settings when optimization is enabled}} \\
\midrule
\texttt{optimize\_steps} & Choice([1, 4, 8, 16, 32, 64, 128, None]) \\
\texttt{finetuning\_lr} & LogUniform([1e-5, 3e-3]) \\
\texttt{finetuning\_dropout} & Uniform([0.0, 0.5]) \\
\texttt{finetuning\_weight\_decay} & LogUniform([1e-6, 0.1]) \\
\texttt{finetuning\_parameter\_scope} & \makecell[l]{Choice(["all", "network", "head"],\\ probs=[0.8, 0.1, 0.1])} \\
\texttt{finetuning\_batch\_size} & Choice([128, 256, 1024]) \\
\texttt{label\_smoothing} & Choice([0.0, 0.05, 0.1]) \\
\bottomrule
\end{tabular}
\end{table}

\begin{table}[!htbp]
\centering
\caption{Hyperparameter search space for TabPFNv2.5.}
\label{space_tabpfnv2.5}
\resizebox{\textwidth}{!}{%
\begin{tabular}{ll}
\toprule
Hyperparameter & Space \\
\midrule
\multicolumn{2}{l}{\textit{Main Parameters}} \\
\midrule
\makecell[l]{\texttt{zip\_model\_path}\\(classification)} &
\makecell[l]{
Choice(["tabpfn-v2.5-classifier-v2.5\_default-2.ckpt",\\
"tabpfn-v2.5-classifier-v2.5\_large-features-L.ckpt",\\
"tabpfn-v2.5-classifier-v2.5\_large-features-XL.ckpt",\\
"tabpfn-v2.5-classifier-v2.5\_large-samples.ckpt",\\
"tabpfn-v2.5-classifier-v2.5\_variant.ckpt"])
} \\
\makecell[l]{\texttt{zip\_model\_path}\\(regression)} &
\makecell[l]{
Choice(["tabpfn-v2.5-regressor-v2.5\_default.ckpt",\\
"tabpfn-v2.5-regressor-v2.5\_low-skew.ckpt",\\
"tabpfn-v2.5-regressor-v2.5\_quantiles.ckpt",\\
"tabpfn-v2.5-regressor-v2.5\_small-samples.ckpt",\\
"tabpfn-v2.5-regressor-v2.5\_variant.ckpt"])
} \\
\texttt{softmax\_temperature} &
Choice([0.25, 0.5, 0.6, 0.7, 0.8, 0.9, 1.0, 1.25, 1.5]) \\
\texttt{balance\_probabilities} &
Choice([True, False]) \\
\midrule
\multicolumn{2}{l}{\textit{Inference Config}} \\
\midrule
\texttt{OUTLIER\_REMOVAL\_STD} &
Choice([3, 6, 12]) \\
\texttt{POLYNOMIAL\_FEATURES} &
Choice(["no", 25]) \\
\texttt{REGRESSION\_Y\_PREPROCESS\_TRANSFORMS} &
\makecell[l]{
Choice([[None], [None, "safepower"], ["safepower"],\\
{}["kdi\_alpha\_0.3"], ["kdi\_alpha\_1.0"],\\
{}["kdi\_alpha\_3.0"], ["quantile\_uni"]])
} \\
\midrule
\multicolumn{2}{l}{\textit{Preprocessing}} \\
\midrule
\texttt{scaling} &
\makecell[l]{
Choice([["none"], ["quantile\_uni\_coarse"],\\
{}["quantile\_norm\_coarse"], ["kdi\_uni"],\\
{}["kdi\_alpha\_0.3"], ["kdi\_alpha\_3.0"],\\
{}["safepower", "quantile\_uni"],\\
{}["none", "quantile\_uni\_coarse"],\\
{}["squashing\_scaler\_default", "quantile\_uni\_coarse"],\\
{}["squashing\_scaler\_default"]])
} \\
\texttt{categoricals} &
Choice(["numeric", "onehot", "none"]) \\
\texttt{append\_original} &
Choice([True, False]) \\
\texttt{global} &
Choice([None, "svd", "svd\_quarter\_components"]) \\
\bottomrule
\end{tabular}%
}
\end{table}

\begin{table}[!htbp]
\centering
\caption{Hyperparameter search space for TabPFN-Wide. Setting \texttt{scaling=None} selects the upstream TabPFNv2 preprocessing preset.}
\label{space_tabpfn-wide}
\resizebox{\textwidth}{!}{%
\begin{tabular}{ll}
\toprule
Hyperparameter & \multicolumn{1}{l}{Space} \\
\midrule
\multicolumn{2}{l}{\textit{Main Parameters}} \\
\midrule
\texttt{checkpoint} & \multicolumn{1}{l}{\makecell[l]{Choice(["1.5k", "1.5k-nocat", "5k", "5k-nocat", "8k", "8k-nocat"])}} \\
\texttt{n\_estimators} & \multicolumn{1}{l}{Choice([1, 4, 8, 16])} \\
\texttt{softmax\_temperature} & \multicolumn{1}{l}{Choice([0.25, 0.5, 0.6, 0.7, 0.8, 0.9, 1.0, 1.25, 1.5])} \\
\texttt{balance\_probabilities} & \multicolumn{1}{l}{Choice([True, False])} \\
\texttt{average\_before\_softmax} & \multicolumn{1}{l}{Choice([True, False])} \\
\midrule
\multicolumn{2}{l}{\textit{Inference Config}} \\
\midrule
\texttt{OUTLIER\_REMOVAL\_STD} & \multicolumn{1}{l}{Choice([3, 6, 12])} \\
\midrule
\multicolumn{2}{l}{\textit{Preprocessing}} \\
\midrule
\texttt{scaling} & \multicolumn{1}{l}{
\makecell[l]{Choice([ None, ["none"], ["quantile\_uni\_coarse"], ["quantile\_norm\_coarse"],
\\ \hphantom{Choice([} ["kdi\_uni"],
["kdi\_alpha\_0.3"], ["kdi\_alpha\_3.0"],
\\ \hphantom{Choice([} ["safepower", "quantile\_uni"],
["none", "quantile\_uni\_coarse"],
\\ \hphantom{Choice([} ["squashing\_scaler\_default", "quantile\_uni\_coarse"],
\\ \hphantom{Choice([} ["squashing\_scaler\_default"] ])
}}\\
\texttt{categoricals} & \multicolumn{1}{l}{Choice(["numeric", "onehot", "none"])} \\
\texttt{append\_original} & \multicolumn{1}{l}{Choice(["auto", False])} \\
\texttt{global} & \multicolumn{1}{l}{Choice([None, "svd", "svd\_quarter\_components"])} \\
\bottomrule
\end{tabular}
}
\end{table}

\begin{table}[!htbp]
\centering
\caption{Hyperparameter search space for RealTabPFNv2.5.}
\label{space_realtabpfn}
\resizebox{\textwidth}{!}{%
\begin{tabular}{ll}
\toprule
Hyperparameter & Space \\
\midrule
\multicolumn{2}{l}{\textit{Main Parameters}} \\
\midrule
\makecell[l]{\texttt{zip\_model\_path}\\(classification)} &
\makecell[l]{
Choice(["tabpfn-v2.5-classifier-v2.5\_default.ckpt",\\
"tabpfn-v2.5-classifier-v2.5\_default-2.ckpt",\\
"tabpfn-v2.5-classifier-v2.5\_large-features-L.ckpt",\\
"tabpfn-v2.5-classifier-v2.5\_large-features-XL.ckpt",\\
"tabpfn-v2.5-classifier-v2.5\_large-samples.ckpt",\\
"tabpfn-v2.5-classifier-v2.5\_real-large-features.ckpt",\\
"tabpfn-v2.5-classifier-v2.5\_real-large-samples-and-features.ckpt",\\
"tabpfn-v2.5-classifier-v2.5\_real.ckpt",\\
"tabpfn-v2.5-classifier-v2.5\_variant.ckpt"])
} \\
\makecell[l]{\texttt{zip\_model\_path}\\(regression)} &
\makecell[l]{
Choice(["tabpfn-v2.5-regressor-v2.5\_default.ckpt",\\
"tabpfn-v2.5-regressor-v2.5\_low-skew.ckpt",\\
"tabpfn-v2.5-regressor-v2.5\_quantiles.ckpt",\\
"tabpfn-v2.5-regressor-v2.5\_real-variant.ckpt",\\
"tabpfn-v2.5-regressor-v2.5\_real.ckpt",\\
"tabpfn-v2.5-regressor-v2.5\_small-samples.ckpt",\\
"tabpfn-v2.5-regressor-v2.5\_variant.ckpt"],\\
probs=[$1/9$, $1/9$, $1/9$, $1/9$, $1/9$, $1/9$, $1/3$])
} \\
\texttt{softmax\_temperature} &
Choice([0.25, 0.5, 0.6, 0.7, 0.8, 0.9, 1.0, 1.25, 1.5]) \\
\texttt{balance\_probabilities} &
Choice([True, False]) \\
\midrule
\multicolumn{2}{l}{\textit{Inference Config}} \\
\midrule
\texttt{OUTLIER\_REMOVAL\_STD} &
Choice([3, 6, 12]) \\
\texttt{POLYNOMIAL\_FEATURES} &
Choice(["no", 25]) \\
\texttt{REGRESSION\_Y\_PREPROCESS\_TRANSFORMS} &
\makecell[l]{
Choice([[None], [None, "safepower"], ["safepower"],\\
{}["kdi\_alpha\_0.3"], ["kdi\_alpha\_1.0"],\\
{}["kdi\_alpha\_3.0"], ["quantile\_uni"]])
} \\
\midrule
\multicolumn{2}{l}{\textit{Preprocessing}} \\
\midrule
\texttt{scaling} &
\makecell[l]{
Choice([["none"], ["quantile\_uni\_coarse"],\\
{}["quantile\_norm\_coarse"], ["kdi\_uni"],\\
{}["kdi\_alpha\_0.3"], ["kdi\_alpha\_3.0"],\\
{}["safepower", "quantile\_uni"],\\
{}["none", "quantile\_uni\_coarse"],\\
{}["squashing\_scaler\_default", "quantile\_uni\_coarse"],\\
{}["squashing\_scaler\_default"]])
} \\
\texttt{categoricals} &
Choice(["numeric", "onehot", "none"]) \\
\texttt{append\_original} &
Choice([True, False]) \\
\texttt{global} &
Choice([None, "svd", "svd\_quarter\_components"]) \\
\bottomrule
\end{tabular}%
}
\end{table}

\begin{table}[!htbp]
\centering
\caption{Hyperparameter search space for TabDPT. Parameters with * apply only to classification.}
\label{space_tabdpt}
\begin{tabular}{ll}
\toprule
Hyperparameter & \multicolumn{1}{l}{Space} \\
\midrule
\texttt{temperature}* & \multicolumn{1}{l}{Choice([0.1, 0.2, 0.3, 0.4, 0.5, 0.6, 0.7, 0.8, 0.9, 1.0, 1.25, 1.5])} \\
\texttt{context\_size} & \multicolumn{1}{l}{Choice([256, 768, 2048])} \\
\texttt{permute\_classes}* & \multicolumn{1}{l}{Choice([True, False])} \\
\texttt{normalizer} & \multicolumn{1}{l}{
\makecell[l]{Choice(["standard", None, "minmax", "robust",\\
\hphantom{Choice([}"power", "quantile-uniform", "quantile-normal", "log1p"])}
} \\
\texttt{missing\_indicators} & \multicolumn{1}{l}{Choice([True, False])} \\
\texttt{clip\_sigma} & \multicolumn{1}{l}{Choice([2, 4, 6, 8])} \\
\texttt{feature\_reduction} & \multicolumn{1}{l}{Choice(["pca", "subsample"])} \\
\texttt{faiss\_metric} & \multicolumn{1}{l}{Choice(["l2", "ip"])} \\
\bottomrule
\end{tabular}
\end{table}

\begin{table}[H]
\centering
\caption{Hyperparameter search space for TabICLv2.
Parameters with * apply only to classification.}
\label{space_tabicl}
\resizebox{\linewidth}{!}{%
\begin{tabular}{ll}
\toprule
Hyperparameter & Space \\
\midrule
\texttt{norm\_methods} & Choice(["none", "power", "robust", "quantile\_rtdl", ["none", "power"]]) \\
\texttt{outlier\_threshold} & Uniform([4.0, 12.0]) \\
\texttt{average\_logits}* & Choice([True, False]) \\
\texttt{softmax\_temperature}* & Uniform([0.7, 1.0]) \\
\bottomrule
\end{tabular}%
}
\end{table}

\FloatBarrier

\subsection{Checkpoints and Pretrained Weights}
\label{apx:checkpoints_pretrained_weights}

Several models in the benchmark rely on pretrained weights or checkpoint-based initialization rather than training from scratch. The pretrained weights loaded from these checkpoints remain frozen during both hyperparameter optimization and final evaluation, while the checkpoint choice may vary across HPO configurations.

The TabPFN family of models uses pretrained transformer-based tabular priors. For TabPFNv2.5, the default evaluation uses \verb|tabpfn-v2.5-classifier-v2.5_default-2.ckpt| for classification and \verb|tabpfn-v2.5-regressor-v2.5_default.ckpt| for regression. HPO selects among five synthetic-only checkpoints for each task type, as listed in Table~\ref{space_tabpfnv2.5}.

RealTabPFNv2.5 follows the same architectural pipeline, with default checkpoints that underwent additional training on real-world data. Its default evaluation uses \verb|tabpfn-v2.5-classifier-v2.5_default.ckpt| for classification and \verb|tabpfn-v2.5-regressor-v2.5_real.ckpt| for regression. HPO selects among nine classification and seven regression checkpoints, including both synthetic-only and real-data-trained candidates, as listed in Table~\ref{space_realtabpfn}.

TabPFN-Wide extends TabPFNv2 with a custom prior for high-dimensional data. It operates on pretrained \verb|.pt| weight files, with available options including \verb|tabpfn-wide-v2-1.5k.pt|, \verb|tabpfn-wide-v2-5k.pt|, and \verb|tabpfn-wide-v2-8k.pt|, together with their corresponding \verb|-nocat.pt| variants. The default configuration uses the standard \verb|tabpfn-wide-v2-5k.pt| checkpoint with one estimator. HPO selects among all six checkpoints, as listed in Table~\ref{space_tabpfn-wide}. TabPFN-Wide supports classification only. We run TabPFN-Wide in a separate environment to accommodate its dependencies on TabPFNv2. 

HyperFast similarly relies on a fixed pretrained checkpoint, \verb|hyperfast.ckpt|, which is loaded prior to inference. The model does not support regression tasks and is restricted to classification.

TabICLv2 uses \texttt{tabicl==2.0.2} with the library-default checkpoints \verb|tabicl-classifier-v2-20260212.ckpt| for classification and \verb|tabicl-regressor-v2-20260212.ckpt| for regression.

HyperFast and iLTM use pretrained hypernetworks to generate task-specific prediction networks. The hypernetwork checkpoints remain frozen, while gradient-based fine-tuning, when enabled, updates the generated prediction networks. iLTM checkpoints are referenced indirectly through configuration keys rather than explicit filenames. The 25 recommended configurations use four checkpoint identifiers: \verb|r128bn|, \verb|rtr|, \verb|cbrconcat|, and \verb|xgbrconcat|, with \verb|xgbrconcat| used as the default. These identifiers correspond to internally defined model configurations within the iLTM library, which resolves the actual pretrained weights and associated preprocessing pipelines through its caching mechanism.

\subsection{Final Refit and Evaluation}
\label{apx:final_refit_and_eval}
 
In general, the refitting procedure re-executes training using the best hyperparameters identified during HPO. However, the handling of training and validation data differs significantly across model families. For models that do not use validation during training, the final refit is performed on a merged training and validation dataset. This includes linear models, tree-based models such as Random Forest, kernel methods such as SVM and NeoLsSVM, and foundation models such as TabPFN, TabICLv2, and TabDPT. This merging maximizes training data availability since these models do not use validation data during fitting.

Conversely, models that depend on validation during training explicitly avoid merging training and validation data in the final stage. This includes boosting models (XGBoost, LightGBM, CatBoost), all neural architectures (MLP, RealMLP, TabM), xRFM, HyperFast, and iLTM. For these models, the validation set remains separate for monitoring and model selection. 

In-context learning models (TabPFNv2.5, RealTabPFNv2.5, TabPFN-Wide, TabICLv2, and TabDPT) do not update pretrained weights during final evaluation. Instead, they rerun preprocessing and inference setup using the validation-selected configuration and the merged training-validation context.

xRFM follows a hybrid procedure: it reinitializes internal preprocessing components and retrains its kernel-based representation with the optimal configuration. Table \ref{tab:model_implementation} summarizes each model's implementation along with the computational resources used.

\begin{table}[h]
\centering
\small
\caption{Comparison of preprocessing pipelines, refitting strategies, and computational resource allocations across models. In the preprocessing columns, \cmark\ indicates a step applied by the shared pipeline and \xmark\ indicates no explicit step. ``Internal'' includes processing within the model-specific wrapper or library, and ``Native'' denotes native categorical support. Shared-pipeline imputation uses KNN unless specified otherwise. Shared-pipeline encoding uses one-hot encoding unless marked ``Label.'' xRFM and RealMLP add missingness indicators to mean-imputed numerical features. TabM additionally removes constant numerical features. In Refit, \cmark\ indicates train-validation merging for final refit and \xmark\ indicates that the split is preserved. CPU and GPU identify the job resource group.}
\label{tab:model_implementation}
\begin{tabular}{lcccccc}
\toprule
\textbf{Model} & \textbf{Imputation} & \textbf{Scaling} & \textbf{Encoding} & \textbf{Refit} & \textbf{CPU} & \textbf{GPU} \\
\midrule
LinearModel & \cmark & \cmark & \cmark & \cmark & \cmark & \\
LassoModel & \cmark & \cmark & \cmark & \cmark & \cmark & \\
RidgeModel & \cmark & \cmark & \cmark & \cmark & \cmark & \\
ElasticNet & \cmark & \cmark & \cmark & \cmark & \cmark & \\

KNN & \cmark & \cmark & \cmark & \cmark & \cmark & \\
SVM & \cmark & \cmark & \cmark & \cmark & \cmark & \\
NeoLsSVM & \cmark & \cmark & \cmark & \cmark & \cmark & \\

DecisionTree & \xmark & \xmark & \cmark & \cmark & \cmark & \\
RandomForest & \xmark & \xmark & \cmark & \cmark & \cmark & \\

XGBoost & \xmark & \xmark & Native & \xmark & & \cmark \\
CatBoost & \xmark & \xmark & Native (str) & \xmark & & \cmark \\
LightGBM & \xmark & \xmark & Native & \xmark & & \cmark \\

xRFM & Mean (internal) & Internal & Internal & \xmark & & \cmark \\
MLP & \cmark & \cmark & \cmark & \xmark & & \cmark \\
RealMLP & Mean (internal) & Internal & Internal & \xmark & & \cmark \\

TabM & Mean & Internal (Quantile) & Label & \xmark & & \cmark \\

TabPFNv2.5 & Internal & Internal & Internal & \cmark & & \cmark \\
TabPFN-Wide & Internal & Internal & Internal & \cmark & & \cmark \\
RealTabPFNv2.5 & Internal & Internal & Internal & \cmark & & \cmark \\

TabICLv2 & Internal & Internal & Internal & \cmark & & \cmark \\
TabDPT & Internal & Internal & Label & \cmark & & \cmark \\

HyperFast & Internal & Internal & Internal & \xmark & & \cmark \\
iLTM & Internal & Internal & Internal & \xmark & & \cmark \\

\bottomrule
\end{tabular}
\end{table}

\textbf{Post-Refit Calibration Diagnostics.} In addition to conventional predictive metrics, we include post-refit probabilistic diagnostic curves.

\begin{figure}[h]
    \centering
    \includegraphics[width=0.85
    \linewidth]{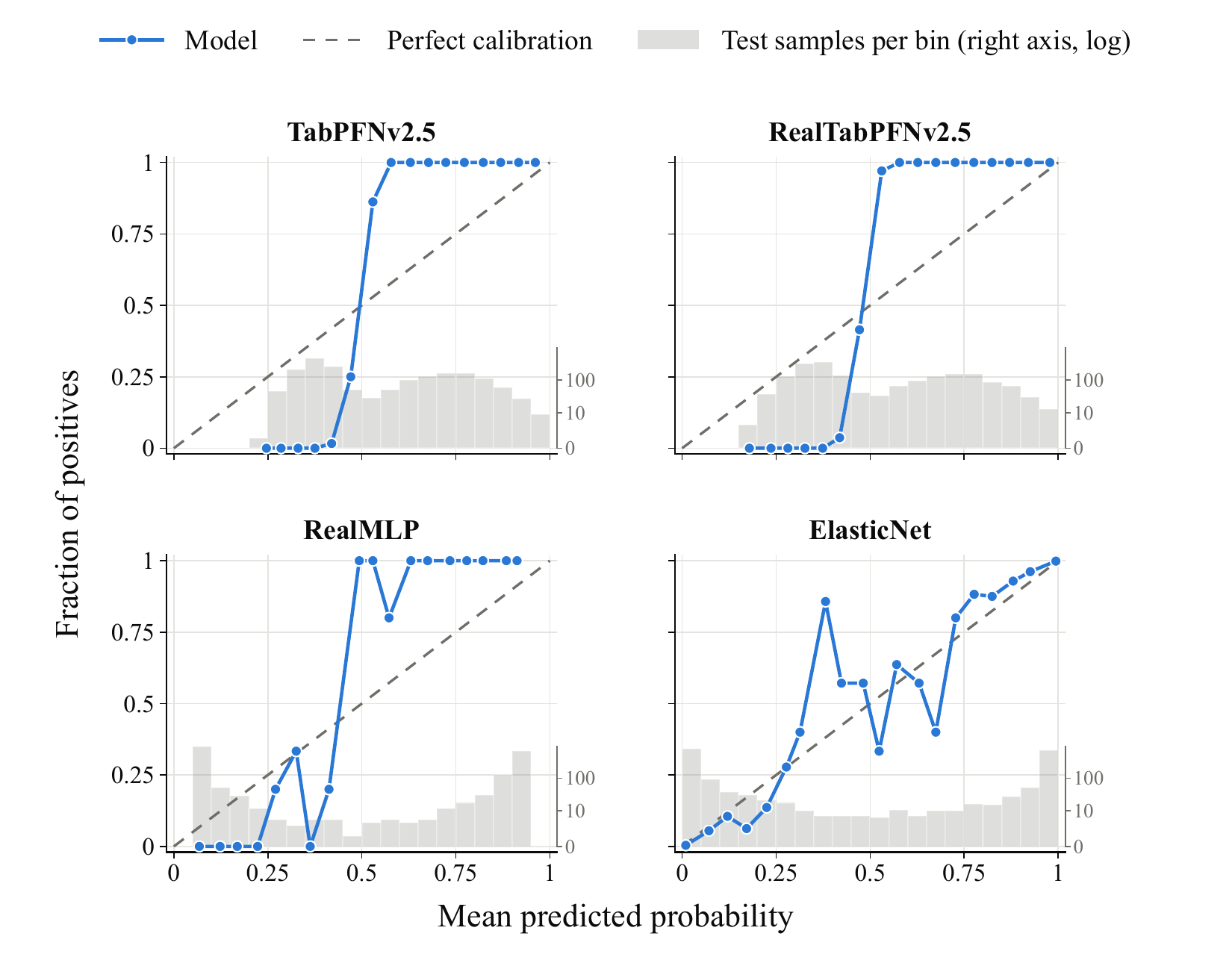}
    \caption{Calibration curves for TabPFNv2.5, RealTabPFNv2.5, RealMLP, and ElasticNet on the \texttt{sexcheck\_chrX\_50k} dataset, evaluated on the held-out test set ($n=1,812$) after final refitting. Each panel shows the observed fraction of positives against the mean predicted probability across 20 equal-width bins, with the dashed diagonal representing perfect calibration. Gray histograms show test-sample counts per probability bin on a $\log_{10}(1+\mathrm{count})$ scale (right axis).}
    \label{fig:calibration_family_comparison}
\end{figure}

Figure \ref{fig:calibration_family_comparison} presents representative calibration curves for four models, including TabPFNv2.5, RealTabPFNv2.5, RealMLP, and ElasticNet, evaluated on the held-out test set of the  \texttt{sexcheck\_chrX\_50k} dataset. The results demonstrate that good discriminative performance does not necessarily translate into well-calibrated predicted probabilities. In particular, \textbf{TabPFNv2.5} and \textbf{RealTabPFNv2.5} achieve exceptionally high test ROC-AUC values of 0.999, yet exhibit substantial miscalibration, with integrated calibration index (ICI) values of 0.26 and 0.245, respectively. Both models display sharp transitions in observed positive fractions across intermediate predicted probabilities, indicating that their probability estimates are strongly underconfident despite their ability to distinguish positive and negative samples.

Similarly, \textbf{RealMLP} achieves a test ROC-AUC of 1.000 but exhibits noticeable deviations from the ideal calibration diagonal (ICI = 0.08). In contrast, \textbf{ElasticNet} achieves a slightly lower, yet still exceptionally strong, test ROC-AUC of 0.995 while demonstrating substantially better probability calibration (ICI = 0.005). This distinction is particularly relevant in biomedical and clinical applications, where downstream risk assessment and decision-making often depend on the accuracy of predicted probabilities rather than discriminative performance alone.

Importantly, post-hoc calibration techniques such as temperature scaling, isotonic regression, or Platt scaling are outside the scope of the current benchmark and were not applied during evaluation.

\section{Benchmark Evaluation}
\label{apx:benchmark_evaluation}

We evaluate all models using task-specific primary metrics defined by the nature of the prediction problem:

\begin{itemize}
    \item Binary classification: Receiver Operating Characteristic Area Under the Curve (ROC-AUC, higher is better)
    \item Multiclass classification: Logarithmic Loss (log loss, lower is better)
    \item Regression and multi-target regression: Mean Squared Error (MSE, lower is better)
\end{itemize}

These metrics serve as the primary evaluation signal across all datasets and are used consistently in all downstream analyses.

\textbf{Fixed-Split Evaluation.} For datasets with predefined train/validation/test splits, models are trained on the training set, tuned on the validation set, and evaluated on the held-out test set. We report test performance corresponding to the selected configuration (default or hyperparameter-tuned), with the test set strictly reserved for final evaluation.

\textbf{Cross-Validation Evaluation.} For datasets evaluated under cross-validation, we use five outer folds. Hyperparameters are selected independently within each outer fold's non-test samples by the mean validation score across three inner folds, reduced to two when class or group support requires it. For automatically generated splits, the stored final-fit training, final-fit validation, and outer-test partitions contain approximately 70\%, 10\%, and 20\% of the dataset, respectively. Dataset-specific predefined splits are preserved. 
The selected configuration is then refitted on the combined training and validation data, except for models that require validation data during fitting, which retain the separate validation subset, and evaluated on the held-out outer test fold.
Final performance is reported as the mean test score across the five outer folds. Standard deviation across folds is also computed and reported where applicable, but is not used for aggregation or ranking. This yields a single dataset-level performance value per model configuration for aggregate comparisons.

\textbf{Normalized Scores.} To enable comparison across heterogeneous datasets and metrics, we apply a per-dataset min--max normalization. For each dataset $d$, let $s_{m,d}$ denote the test performance of model $m$, and let $\mathcal{M}_d$ be the set of evaluated models on that dataset. The normalized score $\tilde{s}_{m,d}$ is defined as:

For metrics where higher values are better:
\[
\tilde{s}_{m,d} =
\frac{s_{m,d} - \min_{m' \in \mathcal{M}_d} s_{m',d}}
{\max_{m' \in \mathcal{M}_d} s_{m',d} - \min_{m' \in \mathcal{M}_d} s_{m',d}}.
\]

For metrics where lower values are better:
\[
\tilde{s}_{m,d} =
\frac{\max_{m' \in \mathcal{M}_d} s_{m',d} - s_{m,d}}
{\max_{m' \in \mathcal{M}_d} s_{m',d} - \min_{m' \in \mathcal{M}_d} s_{m',d}}.
\]

This transformation yields scores in $[0,1]$, where $1$ corresponds to the best-performing model and $0$ corresponds to the worst-performing model on each dataset. The normalization is performed independently per dataset and is intended for comparative analysis within the benchmark rather than as an absolute performance measure.

\subsection{Statistical comparison across all datasets.}
\label{apx:cd_analysis}
The CD analysis includes all 94 datasets and the 21 models applicable to every task type. For each model and dataset, we use the Tuned test result when available, otherwise the Default test result. Models without HPO use Default results. For cross-validation datasets, each variant's test metric is averaged across all outer folds and is considered available only when every fold has a valid result. Successful results are min-max normalized within the 21-model comparison using the formulas above. When neither variant has a complete test result, the model receives a normalized score of 0 and ties with the worst successful model. Per-dataset ranks use average ties, and all datasets are retained. Friedman tests use these ranks, and two-sided Wilcoxon signed-rank tests use the normalized scores, with Holm-Bonferroni correction across all model pairs at $\alpha=0.05$.

\subsection{Elo analysis.}
\label{apx:elo_analysis}
The main Elo analysis evaluates Default and Tuned configurations as separate competitors for the same 21 models used in Table~\ref{tab:rank_summary}. LinearModel and NeoLsSVM are Default-only, giving 40 eligible configurations. For cross-validation datasets, a configuration's primary test metric is averaged across every expected outer fold before comparisons are formed. A configuration with an incomplete outer-fold evaluation has no dataset-level score. Each dataset contributes equally, and pairwise comparisons respect the task-specific metric direction, and ties count as half a win for each competitor.

The main analysis includes all 94 datasets, replacing each missing Default or Tuned score with the Default Random Forest score on the same dataset. This imputes 59 configuration--dataset entries. The complete-case analysis uses the 79 datasets with valid scores for every eligible configuration. Default and Tuned remain separate competitors under both analyses.

Ratings are fitted jointly under the Bradley-Terry model using logistic regression with weak $\ell_2$ regularization ($C=10^6$). We draw 200 bootstrap resamples of datasets with replacement, refit the ratings, and use the 2.5th and 97.5th percentiles for 95\% confidence intervals. The additive offset that anchors the Default Random Forest point estimate to 1000 Elo is also applied to the bootstrap interval endpoints.

The regime and feature-count figures use one competitor per model. For each dataset, they use a complete Tuned test result, otherwise the model's complete Default result. Only when both are unavailable is the Default Random Forest score used. The RandomForest model competitor is anchored at 1000 in these figures.

\section{Pretraining Data Overlap and Robustness}
\label{apx:pretraining_overlap}

\textbf{Source overlap audit.} We cross-referenced the 94 benchmark tasks with the tabular foundation models trained on real datasets. Table~\ref{tab:pretraining_overlap} lists all confirmed source overlaps: 11 tasks for iLTM, six for HyperFast, one for TabDPT, and one for RealTabPFNv2.5, comprising 15 distinct benchmark tasks. Overlap is defined at the source-data level, allowing different preprocessing or target definitions. In particular, TabDPT's Diabetes130US uses seven predictors from the same UCI source as our 40-feature Diabetes\_UCI task. The TabPFNv2.5, TabPFN-Wide, and TabICLv2 checkpoints use synthetic pretraining data \citep{grinsztajn2025tabpfn,kolberg2025tabpfnwide,qu2026tabiclv2}. The reported dataset overlaps are specific to the checkpoints currently available and used in this benchmark, and may change with future model releases and checkpoint versions.

\begin{table}[h]
\centering
\small
\caption{\textbf{Confirmed pretraining source overlaps within the 94-task benchmark.} A check mark indicates that the task shares source data with the model's pretraining or continued-training corpus. Counts overlap across model columns, and the union contains 15 tasks.}
\label{tab:pretraining_overlap}
\resizebox{\linewidth}{!}{%
\begin{tabular}{lcccc}
\toprule
Benchmark dataset & iLTM & HyperFast & TabDPT & RealTabPFNv2.5 \\
\midrule
\texttt{ALLAML} & \cmark & & & \\
\texttt{Cardiovascular-Disease-dataset} & \cmark & & & \cmark \\
\texttt{Diabetes\_UCI} & \cmark & & \cmark & \\
\texttt{GLI\_85} & \cmark & & & \\
\texttt{HeartDisease\_UCI} & \cmark & & & \\
\texttt{QSAR\_biodegradation} & & \cmark & & \\
\texttt{bioresponse} & & \cmark & & \\
\texttt{blood-transfusion-service} & & \cmark & & \\
\texttt{dna} & & \cmark & & \\
\texttt{ilpd\_patient\_data} & \cmark & \cmark & & \\
\texttt{pima\_diabetes} & \cmark & & & \\
\texttt{sick} & \cmark & \cmark & & \\
\texttt{spectf} & \cmark & & & \\
\texttt{thyroid} & \cmark & & & \\
\texttt{wisconsin-breast-cancer} & \cmark & & & \\
\midrule
Total & 11 & 6 & 1 & 1 \\
\bottomrule
\end{tabular}%
}
\end{table}

\begin{table}[htbp]
\centering
\small
\caption{\textbf{Ranking sensitivity after excluding all confirmed overlapping sources.} Task counts give the number before and after exclusion. Model comparisons contain the same 21 models on both sides, and configuration comparisons contain the same 40 Default/Tuned entries. Mean ranks follow the CD score-selection and failure rules, and Elo uses the missing-score policies of the main analysis. Maximum shift is the largest absolute change in leaderboard position.}
\label{tab:pretraining_overlap_robustness}
\resizebox{\linewidth}{!}{%
\begin{tabular}{llccccc}
\toprule
Coverage & Ranking & Tasks & Kendall $\tau_b$ & $p$ & Top-five overlap & Max.\ shift \\
\midrule
All datasets & Mean rank (21 models) & $94/79$ & 0.883 & $2.28\times10^{-8}$ & $4/5$ & 3.5 \\
RF-imputed & Elo (21 models) & $94/79$ & 0.924 & $1.10\times10^{-13}$ & $4/5$ & 3 \\
RF-imputed & Elo (40 configurations) & $94/79$ & 0.918 & $7.30\times10^{-17}$ & $3/5$ & 6 \\
Complete-case & Mean rank (21 models) & $79/65$ & 0.962 & $4.06\times10^{-16}$ & $5/5$ & 2 \\
Complete-case & Elo (21 models) & $79/65$ & 0.952 & $2.00\times10^{-15}$ & $5/5$ & 3 \\
Complete-case & Elo (40 configurations) & $79/65$ & 0.921 & $5.99\times10^{-17}$ & $4/5$ & 7 \\
\bottomrule
\end{tabular}%
}
\end{table}

\begin{figure}[htbp!]
    \centering
    \includegraphics[width=\linewidth]{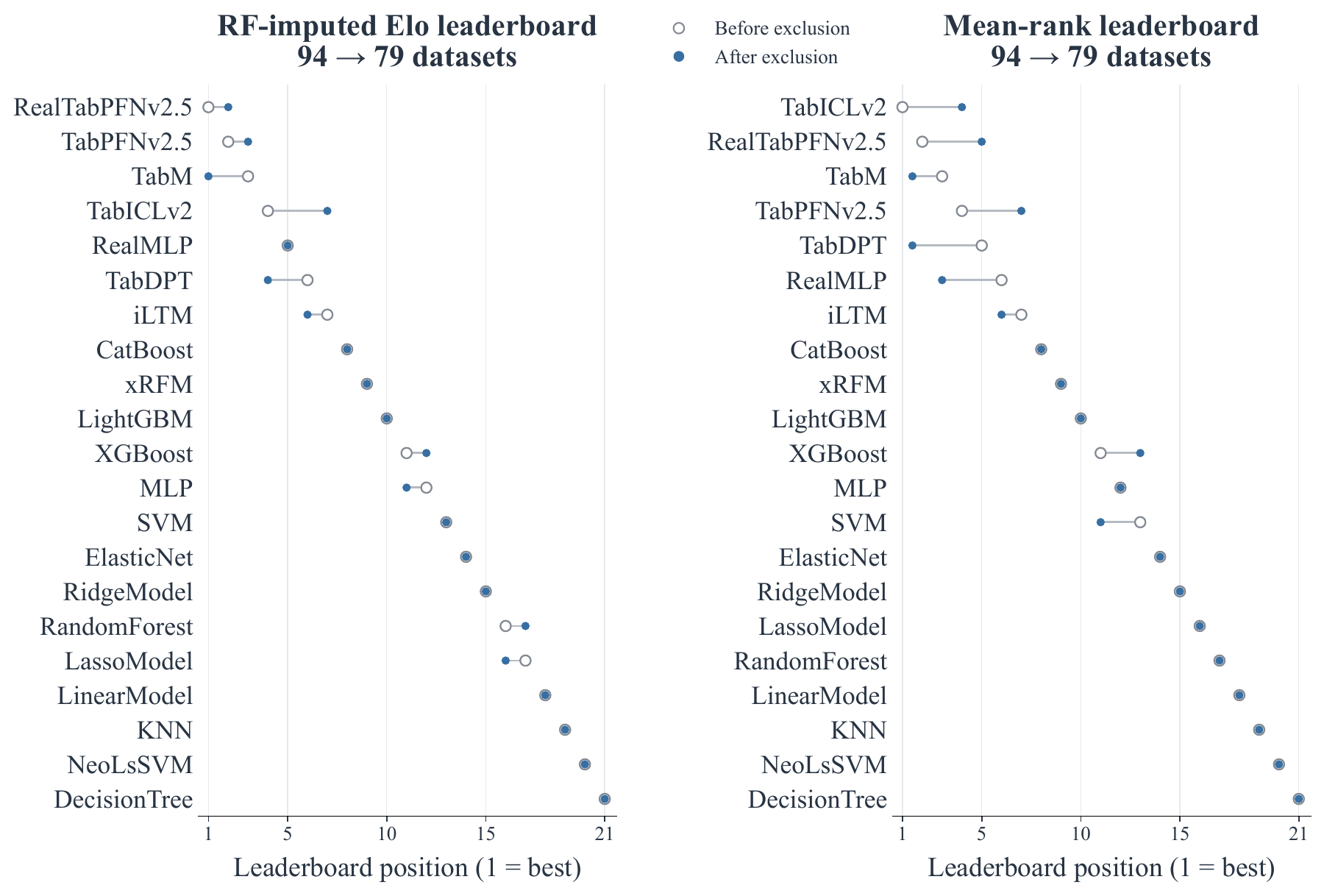}
    \caption{\textbf{Ranking sensitivity to pretraining source overlap.} Leaderboard positions before and after excluding the 15 overlapping datasets, reducing the benchmark from 94 to 79 tasks. Left: RF-imputed Elo leaderboard. Right: mean rank leaderboard.}
    \label{fig:pretraining_overlap_rank_shifts}
\end{figure}

We recomputed the mean-rank and Elo leaderboards excluding the union of overlapping tasks for every model, leaving 79 tasks outside the overlap set. The mean-rank and RF-imputed Elo leaderboards each retain four of their original top five models (Table~\ref{tab:pretraining_overlap_robustness}). After exclusion, TabDPT rises from fifth to joint first in mean rank, tying with TabM, and from sixth to fourth in RF-imputed Elo (Figure~\ref{fig:pretraining_overlap_rank_shifts}). iLTM improves from seventh to sixth in both rankings even after excluding the set of datasets that includes the 11 tasks overlapping with its pretraining sources. Complete-case comparisons retain all five leading models under both ranking methods, using 65 of the original 79 datasets after exclusion. The rankings show substantial agreement after overlap exclusion, while individual positions remain sensitive to dataset composition.

\section{Dependence Among Related Datasets}
\label{apx:source_dependence}

Several TabBioMed tasks are derived from the same observations or from the same source collection. Examples include the twelve genotype tasks built from the 1KG+HGDP callset or the endpoints selected from MoleculeNet. Treating all 94 tasks as independent evidence could therefore give additional weight to some sources. We identified these relations in the processed data, verified them against the benchmark results, and recomputed the leaderboards with group weighting and leave-one-group-out exclusion.

\paragraph{Dataset groups.} We distinguish two levels of dependence. \emph{Same-X groups} contain tasks derived from shared or overlapping observations (individuals, cells, molecules, patients, or sequence pairs), potentially differing in prediction targets, feature representations, or retained samples. \emph{Source families} group tasks by data-generation project, curated collection, or domain-specific repository, without requiring shared observations. This broader grouping tests sensitivity to source composition. Generic multi-domain hosts (UCI, OpenML, TALENT, PMLB, PhysioNet, and Hugging Face) are not treated as families. Each grouping is a partition of the 94 tasks in which ungrouped tasks are singletons, giving 71 same-X units and 42 source-family units. The two levels are analyzed separately. Table~\ref{tab:source_groups} lists every group.

\begin{table}[htbp]
\centering
\caption{\textbf{Dataset groups used in the dependence analysis.} X denotes a same-X group and F a source family; groups marked X/F are both. Families also contain the same-X groups nested in them, and all other tasks are singletons.}
\label{tab:source_groups}
\footnotesize
\setlength{\tabcolsep}{3pt}
\renewcommand{\arraystretch}{1.12}
\begin{tabular}{>{\raggedright\arraybackslash}p{2.2cm}c>{\raggedright\arraybackslash}p{6.3cm}>{\raggedright\arraybackslash}p{3.8cm}}
\toprule
Group & Level & Tasks & Shared observations or source \\
\midrule
1KG+HGDP & X/F & \texttt{continental\_1k}, \texttt{continental\_50k}, \texttt{continental\_50k\_missing}, \texttt{coords\_1k}, \texttt{coords\_50k}, \texttt{pca\_1k}, \texttt{pca\_50k}, \texttt{subpop\_1000\_consec}, \texttt{subpop\_50k}, \texttt{sexcheck\_chrX\_50}, \texttt{sexcheck\_chrX\_500}, \texttt{sexcheck\_chrX\_50k} (12) & Same individuals; identical $X$ for the \texttt{continental}/\texttt{coords} pairs \\
QM8 & X & \texttt{QM8\_E1-CC2}, \texttt{QM8\_f1-CC2} & Identical Coulomb matrices \\
QM9 & X & \texttt{QM9\_g298}, \texttt{QM9\_gap} & Identical Coulomb matrices \\
SIDER & X & \texttt{SIDER\_gastro}, \texttt{SIDER\_nervous} & Identical fingerprints \\
Tox21 & X & \texttt{Tox21\_NRAhR}, \texttt{Tox21\_NRER}, \texttt{Tox21\_SRMMP} & 87--91\% shared molecules \\
Tahoe cells & X & \texttt{tahoe\_cell\_name\_500}, \texttt{tahoe\_cell\_name\_1000}, \texttt{tahoe\_cell\_name\_5000}, \texttt{tahoe\_g2m\_score} & Same cells and splits; nested gene panels \\
AVIDa-hTNFa & X & \texttt{avida\_htnfa\_3mer}, \texttt{avida\_htnfa\_esm2}, \texttt{avida\_htnfa\_onehot} & Same pairs, labels, and splits \\
mouse kidney & X & \texttt{sc\_muris\_age}, \texttt{sc\_muris\_celltype} & Same 1,817 cells \\
Garavan thyroid & X/F & \texttt{sick}, \texttt{thyroid} & 87\% of complete \texttt{sick} records recur in \texttt{thyroid} \\
\midrule
MoleculeNet & F & QM8, QM9, SIDER, and Tox21 groups, \texttt{ESOL}, \texttt{HIV} (11) & Same molecular benchmark collection \\
Tahoe-100M & F & Tahoe cells, \texttt{tahoe\_batch\_integration\_14plates} (5) & Same screen \\
AVIDa (COGNANO) & F & AVIDa-hTNFa, \texttt{avida\_hil6\_onehot}, \texttt{avida\_sarscov2\_3mer} (5) & Same assay platform \\
MLOmics & F & \texttt{MultiOmics\_GS-BRCA}, \texttt{-COAD}, \texttt{-GBM}, \texttt{-LGG}, \texttt{-OV} (5) & TCGA multi-omics pipeline \\
MetaboLights & F & \texttt{MTBLS136}, \texttt{MTBLS161}, \texttt{MTBLS404} & Same repository \\
Metabolomics Workbench & F & \texttt{ST000369}, \texttt{ST000496}, \texttt{ST001000}, \texttt{ST001047} & Same repository \\
TabPFN-VOCS & F & \texttt{Gum1\_s}, \texttt{Noc\_s}, \texttt{PT-Mush21} & Same PTR-ToF-MS collection \\
TRUE-CDS (PEER) & F & \texttt{true\_betalactamase\_unique}, \texttt{true\_fluorescence}, \texttt{true\_melting\_point} & Same protein-task collection \\
scikit-feature & F & \texttt{ALLAML}, \texttt{CLL\_SUB\_111}, \texttt{GLIOMA}, \texttt{GLI\_85}, \texttt{Prostate\_GE}, \texttt{SMK\_CAN\_187}, \texttt{TOX\_171}, \texttt{lung} (8) & Same microarray collection \\
scUnified & F & mouse kidney, \texttt{sc\_muraro\_celltype} (3) & Same harmonized collection \\
\bottomrule
\end{tabular}
\end{table}

\paragraph{Weighting and exclusion.} We use the 21 models supporting all 94 tasks. For weighting, each task in a group $g$ receives weight $1/|g|$ in the pairwise comparisons, so every group, including singletons, counts as a single task, and bootstrap intervals resample whole groups. For the exclusion analysis, we remove the tasks of one group at a time, for each of the 19 groups, and recompute the unweighted ratings. We compare every variant with the unweighted ranking using Kendall's $\tau_b$ and the overlap between their five leading models.

\begin{figure}[htbp]
    \centering
    \includegraphics[width=\linewidth]{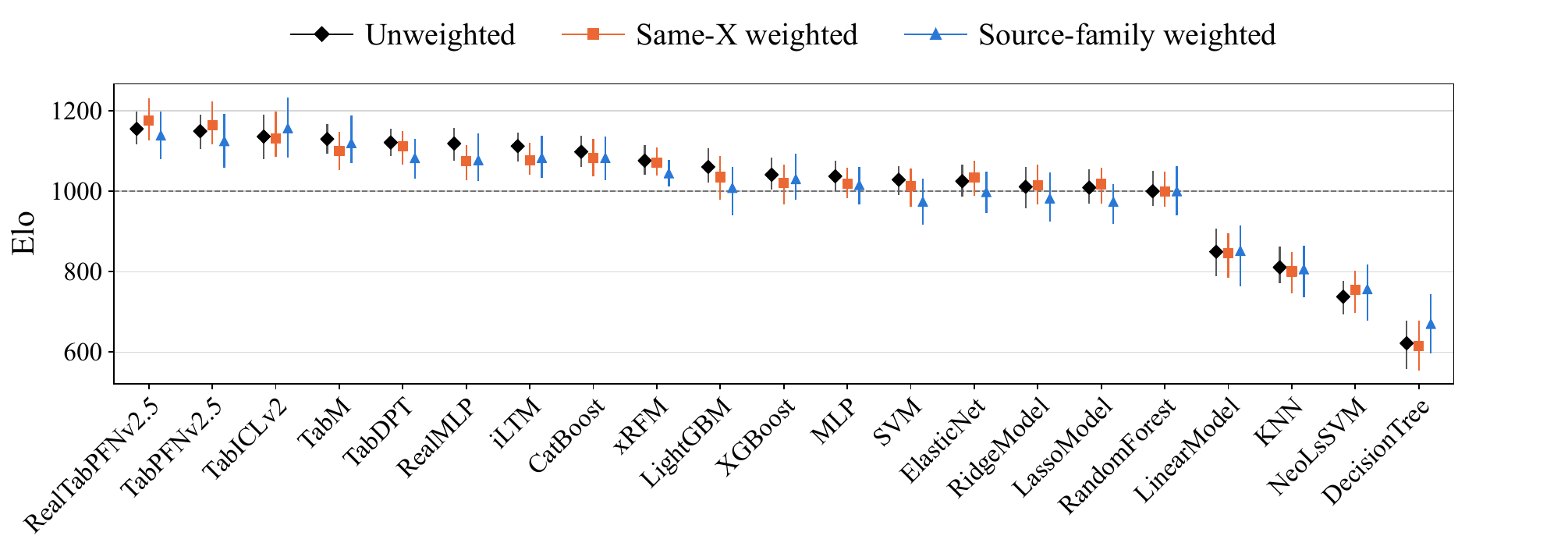}
    \caption{\textbf{Elo ratings under dependence-aware weighting.} Elo of 21 models supporting all 94 tasks without weighting and under same-X and source-family weighting. Scores use Tuned results, falling back to the model's own Default and then RF Default. The RandomForest model is anchored to 1000 (dashed line). Bars are 95\% intervals from 200 dataset-bootstrap (unweighted) or cluster-bootstrap (weighted) resamples. Models are ordered by unweighted Elo.}
    \label{fig:source_dependence}
\end{figure}

\paragraph{Results.} Same-X weighting retains the unweighted top five, with $\tau_b=0.905$. Source-family weighting retains four of five, with $\tau_b=0.867$, and iLTM replaces TabDPT in the top five (Figure~\ref{fig:source_dependence}). RealTabPFNv2.5 has the highest Elo point estimate without weighting and under same-X weighting, while TabICLv2 leads under source-family weighting. Across the 19 group exclusions, mean $\tau_b=0.962$ (minimum $0.905$), and at least four of the original top five are retained.

\section{Cross-Benchmark Ranking Consistency with TabArena}
\label{app:tabarena_comparison}

To assess whether model rankings obtained on general-purpose tabular benchmarks transfer to biomedical datasets, we compare TabBioMed against TabArena using the 13 models shared between the two benchmarks. The comparison uses the TabBioMed mean ranks from Table~\ref{tab:rank_summary}. For TabArena, we use tuned and ensembled entries, except for TabICLv2, for which we use the Default entry. We examine overall ranking agreement, individual model rank changes, and consistency across prediction tasks. Additionally we quantify agreement using Kendall's rank correlation coefficient ($\tau$)  and Spearman's rank correlation coefficient ($\rho$).

\paragraph{Overall Ranking Agreement.} The two benchmarks exhibit strong overall ranking agreement, with Kendall's $\tau =0.769$ and Spearman's $\rho =0.923$. Among the 13 shared models, both benchmarks have the same top-five set: RealTabPFNv2.5, TabICLv2, TabM, TabDPT, and RealMLP. This complete top-five overlap indicates substantial consistency in the identification of leading tabular learning approaches across general-purpose and biomedical datasets.
Table~\ref{tab:tabarena_model_comparison} compares the rankings of all 13 shared models.

\begin{table}[t]
\centering
\caption{Comparison of model rankings between TabArena and TabBioMed for the 13 shared models, ordered by their TabBioMed positions. $\Delta$ denotes the change in ordinal position (TabBioMed minus TabArena), where positive values indicate a lower placement in TabBioMed. TabBioMed mean ranks follow Table~\ref{tab:rank_summary}, with selection and failure handling as in Appendix~\ref{apx:cd_analysis}. Average ranks are reported as originally computed within each benchmark and are not directly comparable in magnitude.}
\label{tab:tabarena_model_comparison}
\small
\setlength{\tabcolsep}{5pt}
\begin{tabular}{lrrrrr}
\toprule
\multirow{2}{*}[-0.5ex]{\textbf{Model}} &
\multicolumn{2}{c}{\textbf{TabArena}} &
\multicolumn{2}{c}{\textbf{TabBioMed}} &
\multirow{2}{*}[-0.5ex]{\textbf{$\Delta$}} \\
\cmidrule(lr){2-3}
\cmidrule(lr){4-5}
& \textbf{Order} & \textbf{Avg. rank}
& \textbf{Order} & \textbf{Mean rank} & \\
\midrule
TabICLv2 & 1 & 30.34 & 1 & 8.01 & 0 \\
RealTabPFNv2.5 & 2 & 30.52 & 2 & 8.05 & 0 \\
TabM & 5 & 43.61 & 3 & 8.17 & -2 \\
TabDPT & 4 & 42.49 & 4 & 8.37 & 0 \\
RealMLP & 3 & 38.39 & 5 & 8.48 & +2 \\
iLTM & 8 & 47.78 & 6 & 8.65 & -2 \\
CatBoost & 7 & 46.02 & 7 & 9.06 & 0 \\
xRFM & 10 & 52.68 & 8 & 9.62 & -2 \\
LightGBM & 6 & 45.05 & 9 & 10.03 & +3 \\
XGBoost & 9 & 50.45 & 10 & 10.55 & +1 \\
RandomForest & 11 & 70.20 & 11 & 11.62 & 0 \\
LinearModel & 13 & 86.80 & 12 & 15.21 & -1 \\
KNN & 12 & 86.51 & 13 & 15.91 & +1 \\
\bottomrule
\end{tabular}
\end{table}

\paragraph{Ranking Agreement Across Prediction Tasks.}

We further investigate whether cross-benchmark ranking agreement varies across prediction tasks. Table~\ref{tab:tabarena_task_agreement} reports Kendall's $\tau$, Spearman's $\rho$, and the overlap among the five highest-ranked models for each comparable task category.

The strongest agreement is observed in the overall comparison ($\tau$=0.769, $\rho$=0.923), with all five leading models shared between the benchmarks. Among individual prediction tasks, multiclass classification exhibits moderately strong agreement ($\tau$=0.513, $\rho$=0.610), as does binary classification ($\tau$=0.487, $\rho$=0.648). Regression shows comparatively weaker agreement ($\tau$=0.436, $\rho$=0.555).

The top-five overlap decreases from five models in the overall comparison to three models within each individual prediction task. Specifically, RealTabPFNv2.5, TabICLv2, and TabM are shared among the five highest-ranked models for binary classification; LightGBM, RealMLP, and TabM for multiclass classification; and RealMLP, TabDPT, and TabICLv2 for regression. 

\begin{table}[t]
\centering
\caption{Cross-benchmark ranking agreement between TabArena and TabBioMed across prediction tasks, evaluated using the 13 shared models. $N$ denotes the number of TabBioMed datasets in each category. Top-5 overlap reports the number of models appearing among the five highest-ranked models in both benchmarks. No separate multi-target regression comparison is reported because TabArena does not include such tasks.}
\label{tab:tabarena_task_agreement}
\small
\setlength{\tabcolsep}{5pt}
\begin{tabular}{lrrrr}
\toprule
\textbf{Task} &
\textbf{$N$} &
\textbf{Kendall $\tau$} &
\textbf{Spearman $\rho$} &
\textbf{Top-5} \\
\midrule
Overall & 94 & 0.769 & 0.923 & 5/5 \\
Binary classification & 48 & 0.487 & 0.648 & 3/5 \\
Multiclass classification & 30 & 0.513 & 0.610 & 3/5 \\
Regression & 12 & 0.436 & 0.555 & 3/5 \\
\bottomrule
\end{tabular}
\end{table}

\section{Additional Results and Analysis}
\label{apx:additional_results}

We present additional analyses to complement the main benchmark results, covering robustness across tasks, efficiency trade-offs, and structural patterns in model performance. These analyses operate on the same underlying evaluation framework described in Section~\ref{sec:evaluation_metrics} and Appendix~\ref{apx:benchmark_evaluation}, but differ in aggregation level (global, task-specific, and domain-specific views).

\subsection{Realized HPO Budgets and Overtuning Diagnostics}
\label{apx:hpo_budget_sensitivity}
\paragraph{Realized search effort.}
The six-hour fixed-split and two-hour outer-fold limits provide equal wall-clock allowances, not equally exhaustive searches. Models differ in fitting cost, failed attempts and finite search spaces. Table~\ref{tab:hpo_realized_overtuning} reports attempted and successfully scored configurations under the official 100-attempt limit, or 25 recommended configurations for iLTM. For CV, counts are averaged across the five outer searches before taking dataset-level medians and quartiles. The Default configuration remains eligible for selection within the same time limit.

\begin{figure}[p]
\centering
\includegraphics[width=\linewidth]{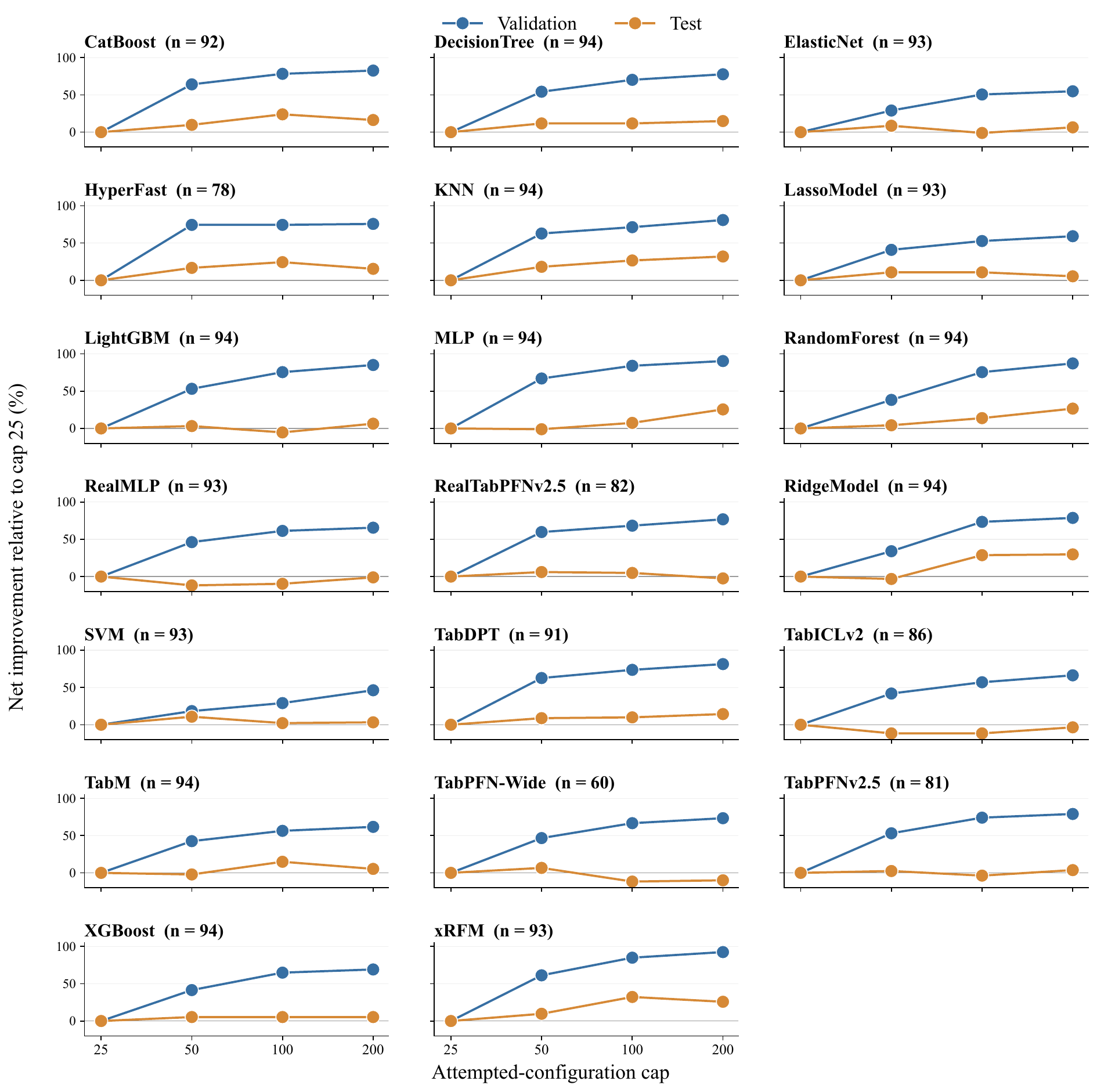}
\caption{Validation and test changes under attempted-configuration caps of 25, 50, 100 and 200. Each circular marker represents one cap. Net improvement is the percentage of datasets improving minus the percentage worsening relative to cap 25. Panel counts give the matched dataset cohort for that model. Flat curves can reflect unchanged selections, finite search spaces or the wall-clock limit. These are within-model comparisons, with different cohorts across models.}
\label{fig:hpo_budget_comparison}
\end{figure}

\paragraph{Configuration-cap comparison.}
We compare validation-only selections at caps of 25, 50, 100 and 200 using the saved trajectories, the original wall-clock limits and the corresponding saved final test evaluations. The 200-attempt continuation is supplementary and does not change the official 100-attempt benchmark. Among 1,933 model--dataset pairs with recorded trajectories, all outer searches reach the four caps in 89.3\%, 81.5\%, 75.4\%, 65.9\% of cases, respectively. An unreached cap uses its available in-budget prefix, rather than implying additional trials. In particular, iLTM has at most 25 recommended configurations. Each model's curves retain the same datasets at all four caps, requiring every outer fold and successful validation-selected test evaluations. Figure~\ref{fig:hpo_budget_comparison} summarizes changes relative to cap 25 as the percentage of datasets that improve minus the percentage that worsen, respecting each primary metric's direction. This avoids averaging ROC-AUC, log loss and MSE in incompatible units.

\begin{figure}[p]
\centering
\includegraphics[width=\linewidth]{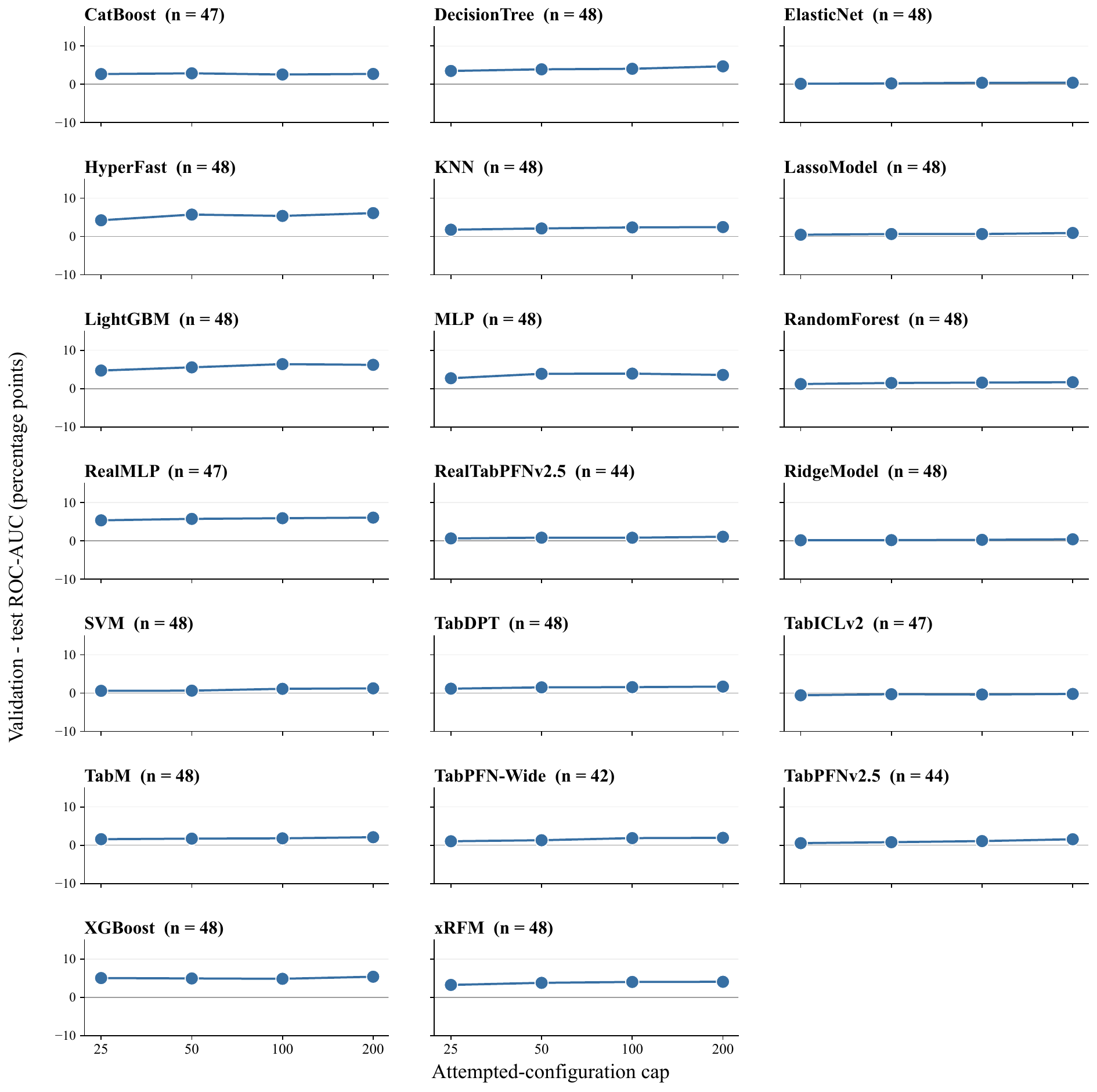}
\caption{Mean validation--test ROC-AUC gaps across the four caps on binary datasets. Each circular marker is a configuration cap, and zero indicates equal validation and test ROC-AUC. Panel counts give matched datasets with all four evaluations. Validation comes from the selection partitions, while test performance is measured after final refitting. Gaps therefore also reflect split differences and final-fit training size, rather than HPO overfitting alone.}
\label{fig:hpo_validation_test_gaps}
\end{figure}

\paragraph{Validation--test gaps and overtuning.}
Figure~\ref{fig:hpo_validation_test_gaps} shows the mean validation minus test ROC-AUC on binary datasets, in percentage points, using a fixed four-cap cohort per model. Positive gaps mean validation performance is higher. Log-loss gaps are reported separately below in their native units. Across all task types, our overtuning diagnostic flags a strict validation improvement over cap 25 accompanied by strictly worse test performance. Table~\ref{tab:hpo_realized_overtuning} reports this rate among datasets with complete paired evaluations at 25 and the indicated cap, including unchanged selections in the denominator. The diagnostic flags 497/1882 pairs at cap 100 and 539/1881 at cap 200.

\begin{table}[htbp]
\centering\small
\caption{Realized official HPO counts and per-model overtuning diagnostics. Counts are medians [25th, 75th percentiles] across datasets, using mean counts per outer fold for CV. Attempted counts include failed configurations. Scored counts require an eligible validation score. Unrecorded counts are omitted. The final two columns report flagged/paired datasets (percentage), where validation performance improves but test performance worsens relative to cap 25.}
\label{tab:hpo_realized_overtuning}
\resizebox{\linewidth}{!}{%
\begin{tabular}{lrrrr}\toprule
Model & Attempted & Scored & Cap 100 vs 25 & Cap 200 vs 25 \\
\midrule
CatBoost & 100 [100, 100] & 100 [100, 100] & 24/93 (25.8\%) & 29/92 (31.5\%) \\
DecisionTree & 100 [100, 100] & 100 [100, 100] & 27/94 (28.7\%) & 29/94 (30.9\%) \\
ElasticNet & 100 [34, 100] & 100 [33, 100] & 22/93 (23.7\%) & 21/93 (22.6\%) \\
HyperFast & 100 [41, 100] & 100 [40, 100] & 19/78 (24.4\%) & 23/78 (29.5\%) \\
KNN & 100 [100, 100] & 100 [100, 100] & 21/94 (22.3\%) & 23/94 (24.5\%) \\
LassoModel & 100 [50, 100] & 100 [49, 100] & 18/93 (19.4\%) & 23/93 (24.7\%) \\
LightGBM & 100 [100, 100] & 100 [100, 100] & 38/94 (40.4\%) & 37/94 (39.4\%) \\
MLP & 100 [100, 100] & 100 [100, 100] & 35/94 (37.2\%) & 29/94 (30.9\%) \\
RandomForest & 100 [100, 100] & 100 [100, 100] & 29/94 (30.9\%) & 28/94 (29.8\%) \\
RealMLP & 100 [61.8, 100] & 100 [46.7, 100] & 33/93 (35.5\%) & 31/93 (33.3\%) \\
RealTabPFNv2.5 & 100 [47.8, 100] & 100 [39.5, 100] & 24/82 (29.3\%) & 32/82 (39.0\%) \\
RidgeModel & 100 [100, 100] & 100 [100, 100] & 17/94 (18.1\%) & 20/94 (21.3\%) \\
SVM & 100 [86.5, 100] & 100 [85.8, 100] & 12/93 (12.9\%) & 19/93 (20.4\%) \\
TabDPT & 100 [100, 100] & 100 [97, 100] & 27/91 (29.7\%) & 29/91 (31.9\%) \\
TabICLv2 & 100 [64, 100] & 100 [63, 100] & 26/86 (30.2\%) & 25/86 (29.1\%) \\
TabM & 100 [23.8, 100] & 100 [22, 100] & 19/94 (20.2\%) & 26/94 (27.7\%) \\
TabPFN-Wide & 100 [29, 100] & 100 [14.5, 100] & 23/60 (38.3\%) & 25/60 (41.7\%) \\
TabPFNv2.5 & 100 [56.2, 100] & 100 [52.5, 100] & 31/81 (38.3\%) & 29/81 (35.8\%) \\
XGBoost & 100 [100, 100] & 100 [100, 100] & 28/94 (29.8\%) & 30/94 (31.9\%) \\
iLTM & 25 [20, 25] & 25 [19, 25] & 0/94 (0.0\%) & 0/94 (0.0\%) \\
xRFM & 100 [100, 100] & 100 [100, 100] & 24/93 (25.8\%) & 31/93 (33.3\%) \\
\bottomrule\end{tabular}}
\end{table}

\paragraph{Default versus Tuned.}
Table~\ref{tab:default_tuned_frequencies} compares the official 100-cap Tuned test result with Default on datasets where both variants have complete test evaluations. CV requires all five outer folds, averaged before comparison. No missing scores are imputed here. Differences of at most $10^{-12}$ in the primary metric count as ties. These are test-performance frequencies, distinct from the validation-improves/test-worsens diagnostic relative to cap 25 above.

\begin{table}[htbp]
\centering\small
\caption{Official Tuned versus Default test outcomes by model. Entries give counts (percentages) among paired datasets, using ROC-AUC for binary classification, log loss for multiclass classification and MSE for regression. Default-only models are excluded.}
\label{tab:default_tuned_frequencies}
\begin{tabular}{lrrrr}\toprule
Model & Paired & Tuned better & Tie & Default better \\
\midrule
CatBoost & 93 & 59 (63.4\%) & 5 (5.4\%) & 29 (31.2\%) \\
DecisionTree & 94 & 79 (84.0\%) & 2 (2.1\%) & 13 (13.8\%) \\
ElasticNet & 93 & 50 (53.8\%) & 23 (24.7\%) & 20 (21.5\%) \\
HyperFast & 78 & 45 (57.7\%) & 11 (14.1\%) & 22 (28.2\%) \\
KNN & 94 & 86 (91.5\%) & 1 (1.1\%) & 7 (7.4\%) \\
LassoModel & 93 & 51 (54.8\%) & 17 (18.3\%) & 25 (26.9\%) \\
LightGBM & 94 & 70 (74.5\%) & 3 (3.2\%) & 21 (22.3\%) \\
MLP & 94 & 70 (74.5\%) & 1 (1.1\%) & 23 (24.5\%) \\
RandomForest & 94 & 61 (64.9\%) & 11 (11.7\%) & 22 (23.4\%) \\
RealMLP & 93 & 74 (79.6\%) & 1 (1.1\%) & 18 (19.4\%) \\
RealTabPFNv2.5 & 81 & 38 (46.9\%) & 11 (13.6\%) & 32 (39.5\%) \\
RidgeModel & 94 & 63 (67.0\%) & 15 (16.0\%) & 16 (17.0\%) \\
SVM & 93 & 87 (93.5\%) & 1 (1.1\%) & 5 (5.4\%) \\
TabDPT & 94 & 48 (51.1\%) & 6 (6.4\%) & 40 (42.6\%) \\
TabICLv2 & 93 & 64 (68.8\%) & 7 (7.5\%) & 22 (23.7\%) \\
TabM & 94 & 67 (71.3\%) & 5 (5.3\%) & 22 (23.4\%) \\
TabPFN-Wide & 60 & 34 (56.7\%) & 5 (8.3\%) & 21 (35.0\%) \\
TabPFNv2.5 & 80 & 34 (42.5\%) & 9 (11.2\%) & 37 (46.2\%) \\
XGBoost & 94 & 79 (84.0\%) & 1 (1.1\%) & 14 (14.9\%) \\
iLTM & 94 & 61 (64.9\%) & 1 (1.1\%) & 32 (34.0\%) \\
xRFM & 92 & 77 (83.7\%) & 4 (4.3\%) & 11 (12.0\%) \\
\bottomrule\end{tabular}
\end{table}

Table~\ref{tab:default_tuned_degradation} quantifies the losses when Default is better. ROC-AUC decreases are percentage points and log-loss increases retain their native units. MSE increases are relative to Default, avoiding averages across incompatible target scales.

\begin{table}[htbp]
\centering\small
\caption{Metric-specific Tuned degradation relative to Default, pooled descriptively over model--dataset pairs. Magnitudes condition on strictly worse Tuned test performance. MSE degradation is $100(\mathrm{MSE}_{T}/\mathrm{MSE}_{D}-1)$, excluding a zero Default MSE from magnitude summaries. No failures are imputed.}
\label{tab:default_tuned_degradation}
\resizebox{\linewidth}{!}{%
\begin{tabular}{lrrrr}\toprule
Primary metric & Paired & Tuned worse & Median [Q1, Q3] & Maximum \\
\midrule
ROC-AUC decrease (pp) & 993 & 311 (31.3\%) & 1.12 [0.372, 2.42] & 24.4 \\
Multiclass log-loss increase & 601 & 101 (16.8\%) & 0.0161 [0.00519, 0.0386] & 0.492 \\
Relative MSE increase (\%) & 295 & 40 (13.6\%) & 9.17 [2.05, 18.5] & 64.4 \\
\bottomrule\end{tabular}}
\end{table}

\begin{figure}[p]
\centering
\includegraphics[width=\linewidth]{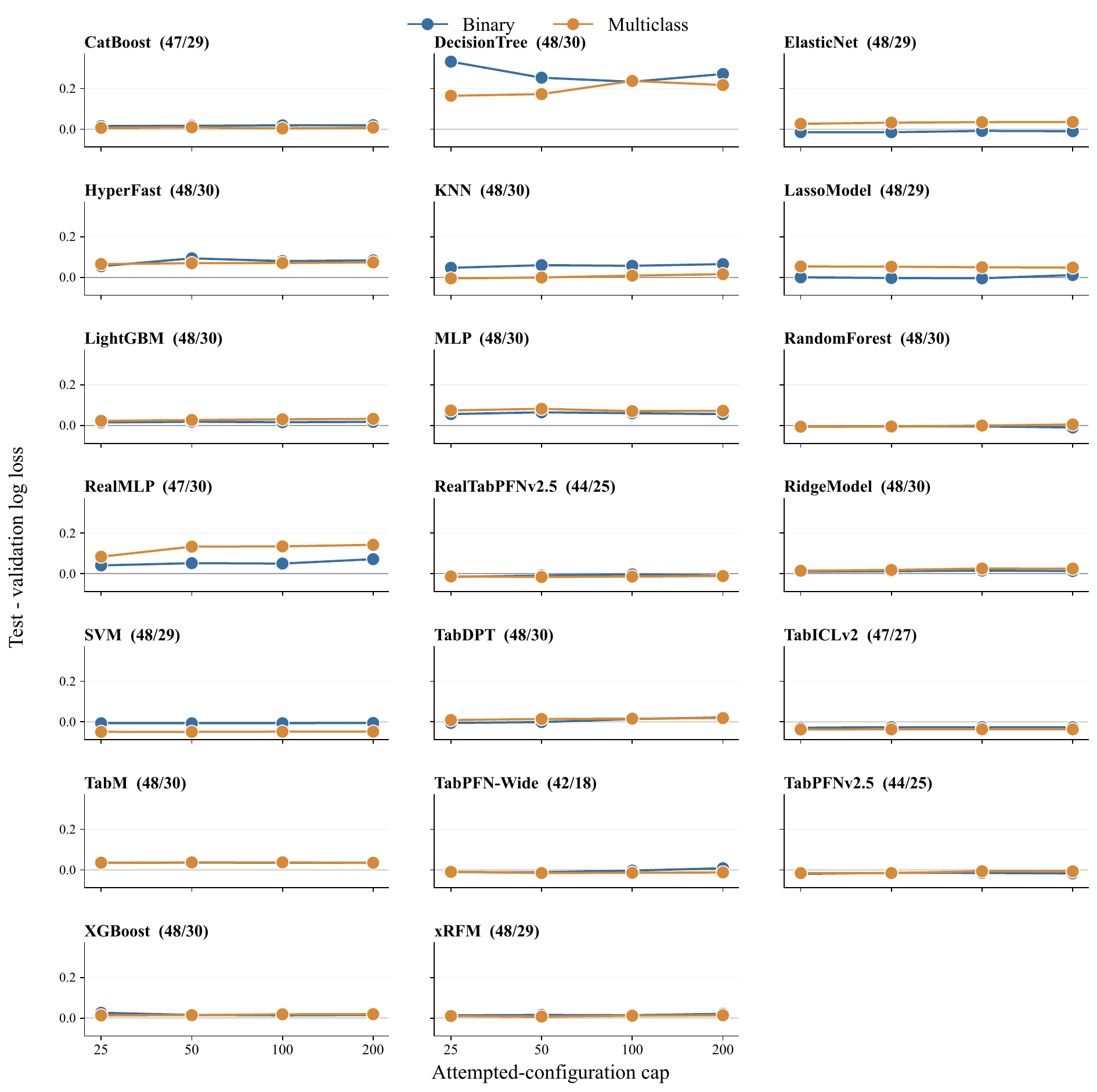}
\caption{Mean test minus validation log loss at each configuration cap, in native log-loss units. Circular markers denote caps 25, 50, 100 and 200. Panel counts are binary/multiclass datasets, with each cohort held fixed across caps. As with ROC-AUC gaps, partition differences and final refitting contribute to the observed gaps.}
\label{fig:hpo_logloss_gaps}
\end{figure}

\paragraph{Log-loss validation--test gaps.}
Figure~\ref{fig:hpo_logloss_gaps} adds binary and multiclass log-loss magnitudes. Positive values indicate higher loss on test than on validation. Binary selections still optimize ROC-AUC, while multiclass selections optimize log loss. Each curve uses datasets with complete validation and test log loss at all four caps, without imputation.

\begin{figure}[p]
\centering
\includegraphics[width=.95\linewidth]{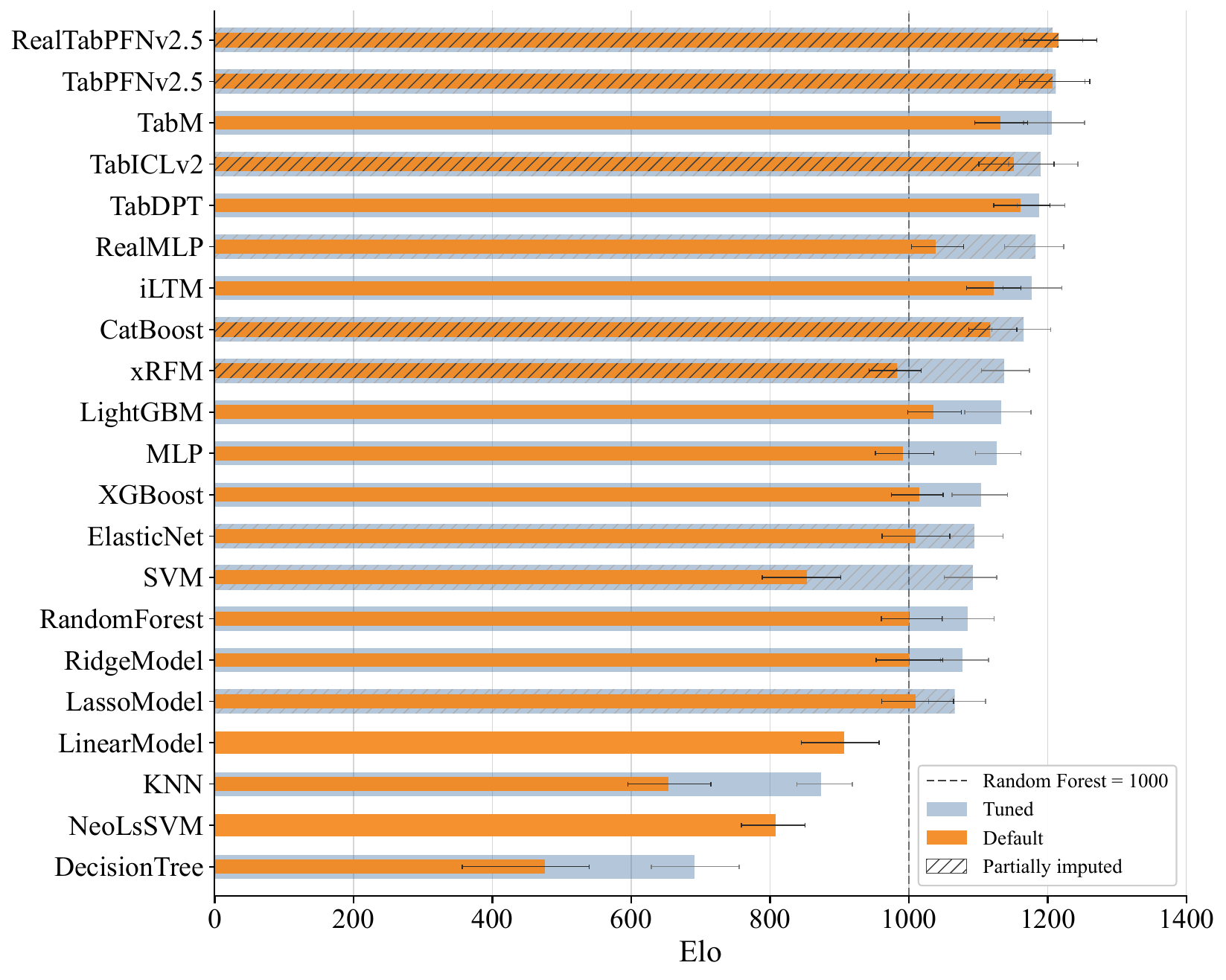}
\caption{Supplementary RF-imputed Elo leaderboard at the 200-attempt cap on all 94 datasets. Default and Tuned share one comparison pool, with RandomForest Default anchored at 1,000. Hatching identifies entries with any RF-imputed score. Intervals are dataset-bootstrap 95\% intervals. LinearModel and NeoLsSVM remain Default-only, and iLTM retains its finite 25-configuration space.}
\label{fig:hpo_200_elo}
\end{figure}

\paragraph{Supplementary 200-cap leaderboard.}
Figure~\ref{fig:hpo_200_elo} repeats the Main RF-imputed Elo procedure on all 94 datasets using the saved validation-selected test result at cap 200 for Tuned. Default results are unchanged. The joint pool contains 40 Default/Tuned entries for the 21 models applicable to every task. Complete outer-fold means precede pairwise comparisons, missing entries use RandomForest Default, ties contribute half a win, and 200 dataset bootstraps provide 95\% intervals. The six-hour/two-hour search limits and finite spaces remain unchanged, so cap 200 does not imply 200 completed configurations. There are 60 imputed entries out of 3760. This supplementary analysis does not replace the official 100-cap leaderboard.

\paragraph{Limitations.}
Optimizing noisy validation estimates can select configurations that generalize worse, as studied by \citet{schneider2025overtuning}. Our paired rates are descriptive overtuning proxies, and they depend on the cap-25 reference. Validation--test gaps also reflect sampling variation, inner versus outer partitions and final-refit protocols. These diagnostics do not use test performance to alter HPO selection, and higher caps are not guaranteed to improve generalization.

\FloatBarrier

\subsection{Failure Reporting and Completion Sensitivity}
\label{apx:failure_reporting}
Among 4,072 applicable Default or Tuned entries, 95 lack complete test results (2.3\%, comprising 45 Default and 50 Tuned entries). An entry is one model--dataset--configuration combination and requires all five outer folds for CV. Unsupported tasks, PCA experiments and A100 retries are excluded. Tables~\ref{tab:failure_models} and~\ref{tab:failure_domains} summarize coverage and recorded causes.

\begin{table}[htbp]
\centering\small
\caption{Main failure counts by model and failure-as-worst mean ranks. Missing Default and Tuned entries are counted separately. Usable results select complete Tuned results, falling back to complete Default results. Ranks cover the 21 models applicable to all 94 datasets, with every failure below all successful models and failures tied with one another. A dash denotes an inapplicable entry.}
\label{tab:failure_models}
\begin{tabular}{lrrrr}
\toprule
Model & \makecell{Missing\\Default} & \makecell{Missing\\Tuned} & \makecell{Usable\\results} & \makecell{Failure-as-worst\\mean rank} \\
\midrule
TabICLv2 & 1 & 1 & 93/94 & 8.016 \\
RealTabPFNv2.5 & 11 & 12 & 84/94 & 8.101 \\
TabM & 0 & 0 & 94/94 & 8.170 \\
TabPFNv2.5 & 12 & 13 & 83/94 & 8.303 \\
TabDPT & 0 & 0 & 94/94 & 8.372 \\
RealMLP & 0 & 1 & 94/94 & 8.484 \\
iLTM & 0 & 0 & 94/94 & 8.649 \\
CatBoost & 1 & 1 & 93/94 & 9.064 \\
xRFM & 2 & 1 & 93/94 & 9.628 \\
LightGBM & 0 & 0 & 94/94 & 10.027 \\
XGBoost & 0 & 0 & 94/94 & 10.548 \\
MLP & 0 & 0 & 94/94 & 10.628 \\
SVM & 0 & 1 & 94/94 & 10.787 \\
ElasticNet & 0 & 1 & 94/94 & 10.936 \\
RidgeModel & 0 & 0 & 94/94 & 11.213 \\
LassoModel & 0 & 1 & 94/94 & 11.362 \\
RandomForest & 0 & 0 & 94/94 & 11.622 \\
LinearModel & 0 & -- & 94/94 & 15.176 \\
KNN & 0 & 0 & 94/94 & 15.883 \\
NeoLsSVM & 0 & -- & 94/94 & 17.191 \\
DecisionTree & 0 & 0 & 94/94 & 18.840 \\
\midrule
HyperFast & 0 & 0 & 78/78 & -- \\
TabPFN-Wide & 18 & 18 & 60/78 & -- \\
\bottomrule\end{tabular}
\end{table}

\begin{table}[htbp]
\centering\small
\caption{Missing Main Default/Tuned entries by domain and recorded failure type. Mixed denotes GPU OOM and segmentation faults recorded during the same search. Counts describe unavailable final test results, not individual failed HPO trials.}
\label{tab:failure_domains}
\begin{tabular}{lrrrrrr}
\toprule
Domain & \makecell{Missing/\\applicable} & \makecell{GPU\\OOM} & \makecell{RAM\\OOM} & \makecell{Time\\limit} & Segfault & Mixed \\
\midrule
Clinical and healthcare & 8/968 & 0 & 0 & 8 & 0 & 0 \\
Drug discovery & 13/552 & 10 & 0 & 2 & 1 & 0 \\
Genomics & 8/600 & 7 & 0 & 0 & 1 & 0 \\
Metabolomics & 0/432 & 0 & 0 & 0 & 0 & 0 \\
Proteomics & 15/472 & 11 & 0 & 0 & 2 & 2 \\
Single-cell & 49/432 & 20 & 2 & 11 & 13 & 3 \\
Systems biology & 0/220 & 0 & 0 & 0 & 0 & 0 \\
Transcriptomics & 2/396 & 2 & 0 & 0 & 0 & 0 \\
\midrule
Total & 95/4072 & 50 & 2 & 21 & 17 & 5 \\
\bottomrule\end{tabular}
\end{table}

Placing failed evaluations strictly below all successful evaluations, rather than tying them with the worst success, leaves the 21-model ordering unchanged (Kendall $\tau_b=1.000$).

Using complete Tuned results, with complete Default results as fallback, completion is 100\% for DecisionTree, ElasticNet, KNN, LassoModel, LightGBM, LinearModel, MLP, NeoLsSVM, RandomForest, RealMLP, RidgeModel, SVM, TabDPT, TabM, XGBoost and iLTM (94/94 datasets) and HyperFast (78/78 classification datasets). The remaining completion rates are CatBoost, TabICLv2 and xRFM (93/94, 98.9\%), RealTabPFNv2.5 (84/94, 89.4\%), TabPFNv2.5 (83/94, 88.3\%) and TabPFN-Wide (60/78, 76.9\%). TabPFN-Wide is evaluated on classification datasets only.

\begin{figure}[htbp]
\centering
\includegraphics[width=\linewidth]{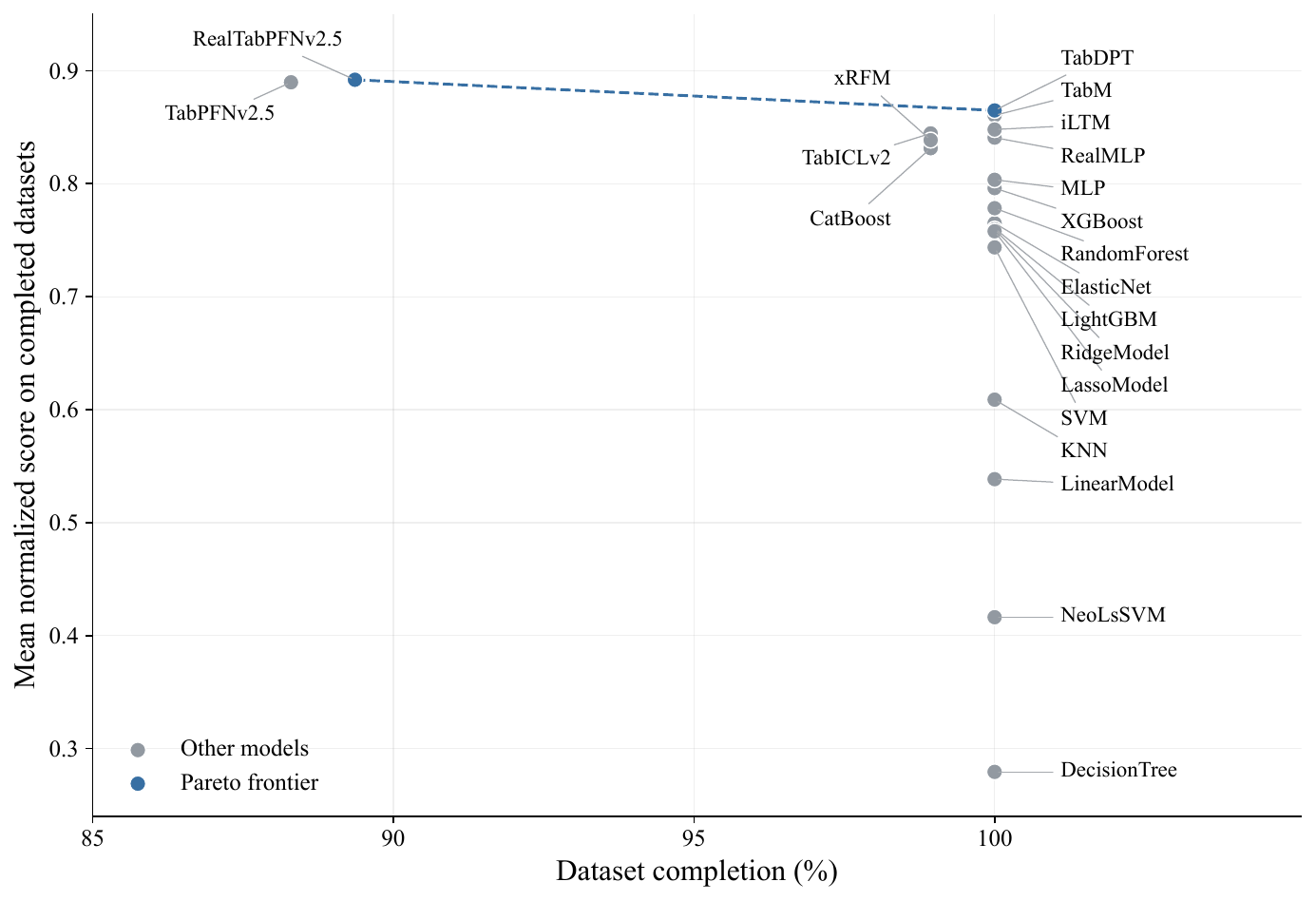}
\caption{Main performance--completion Pareto analysis on the 94-dataset scope. Both axes are maximized, and the dashed line joins nondominated models. Completion and performance use complete Tuned results with complete Default results as fallback. The performance mean uses only each model's completed datasets.}
\label{fig:accuracy_completion_pareto}
\end{figure}

\paragraph{Performance versus completion.}
Figure~\ref{fig:accuracy_completion_pareto} plots completion against mean normalized primary-metric performance on completed datasets for the 21 models applicable to all tasks. RealTabPFNv2.5 and TabDPT form the frontier. Scores are min--max normalized within each dataset over successful models, with higher values better. The means condition on each model's completed datasets, so differences in dataset coverage also affect this descriptive frontier. Failures are not imputed in this plot.

\FloatBarrier

\subsection{GPU-Memory Sensitivity Analysis}
\label{apx:a100_sensitivity}
We assess ranking sensitivity to larger GPU memory using 80\,GB A100 retries for model--dataset pairs affected by GPU out-of-memory errors in the 24\,GB A5500 Main runs. We retain available Main results and fill missing Default or Tuned results with the corresponding successful retry. The retries recover 16 test evaluations across 10 pairs. Mean-rank and Elo calculations follow Appendix~\ref{apx:benchmark_evaluation}.

\begin{figure}[!htbp]
    \centering
    \includegraphics[width=\linewidth]{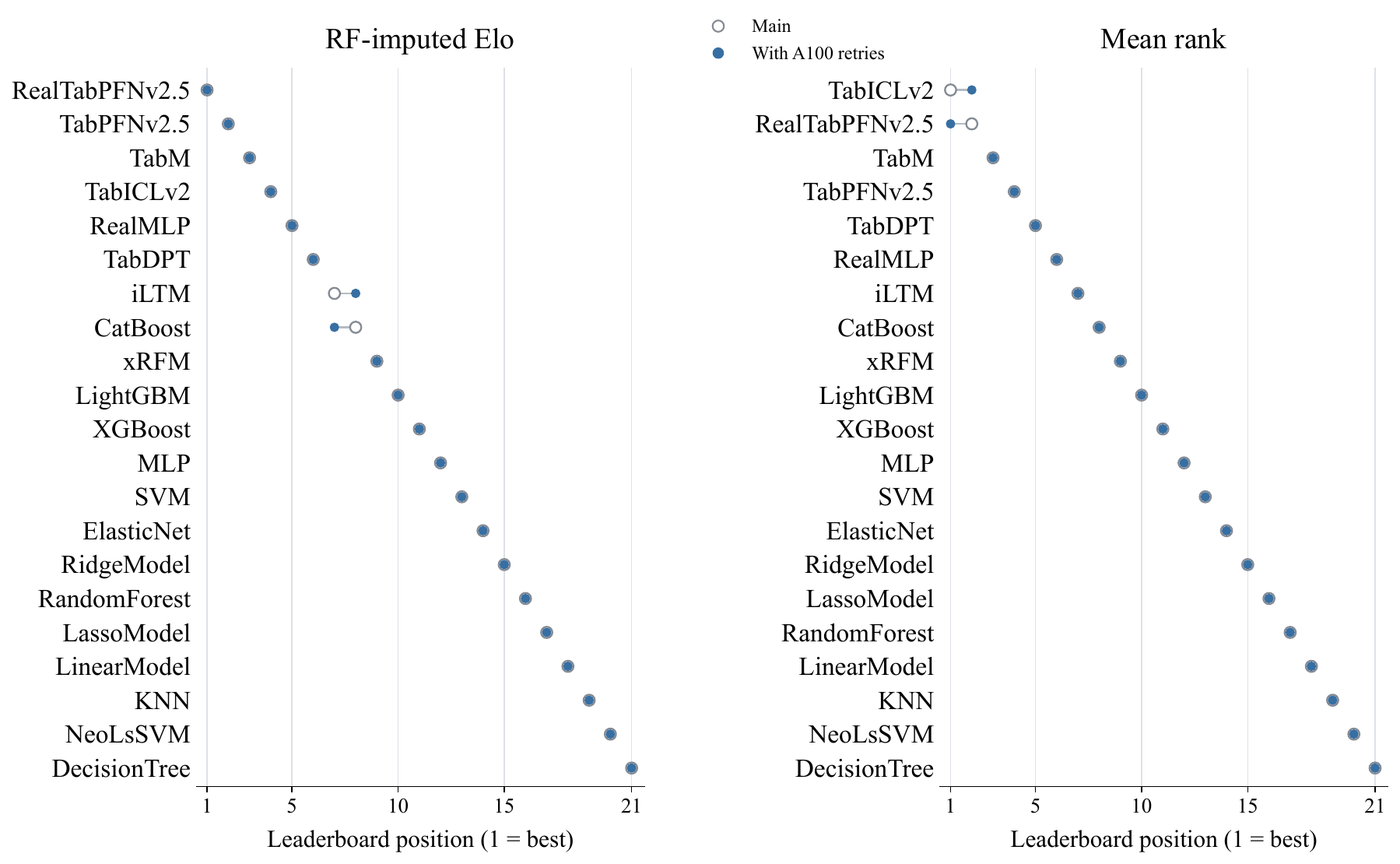}
    \caption{\textbf{A100 retry sensitivity across all 94 datasets and 21 models.} Leaderboard positions using Main results (open markers) and after incorporating successful A100 retry results (filled markers). Left: RF-imputed Elo. Right: mean rank.}
    \label{fig:a100_all_tasks}
\end{figure}

\begin{figure}[!htbp]
    \centering
    \includegraphics[width=\linewidth]{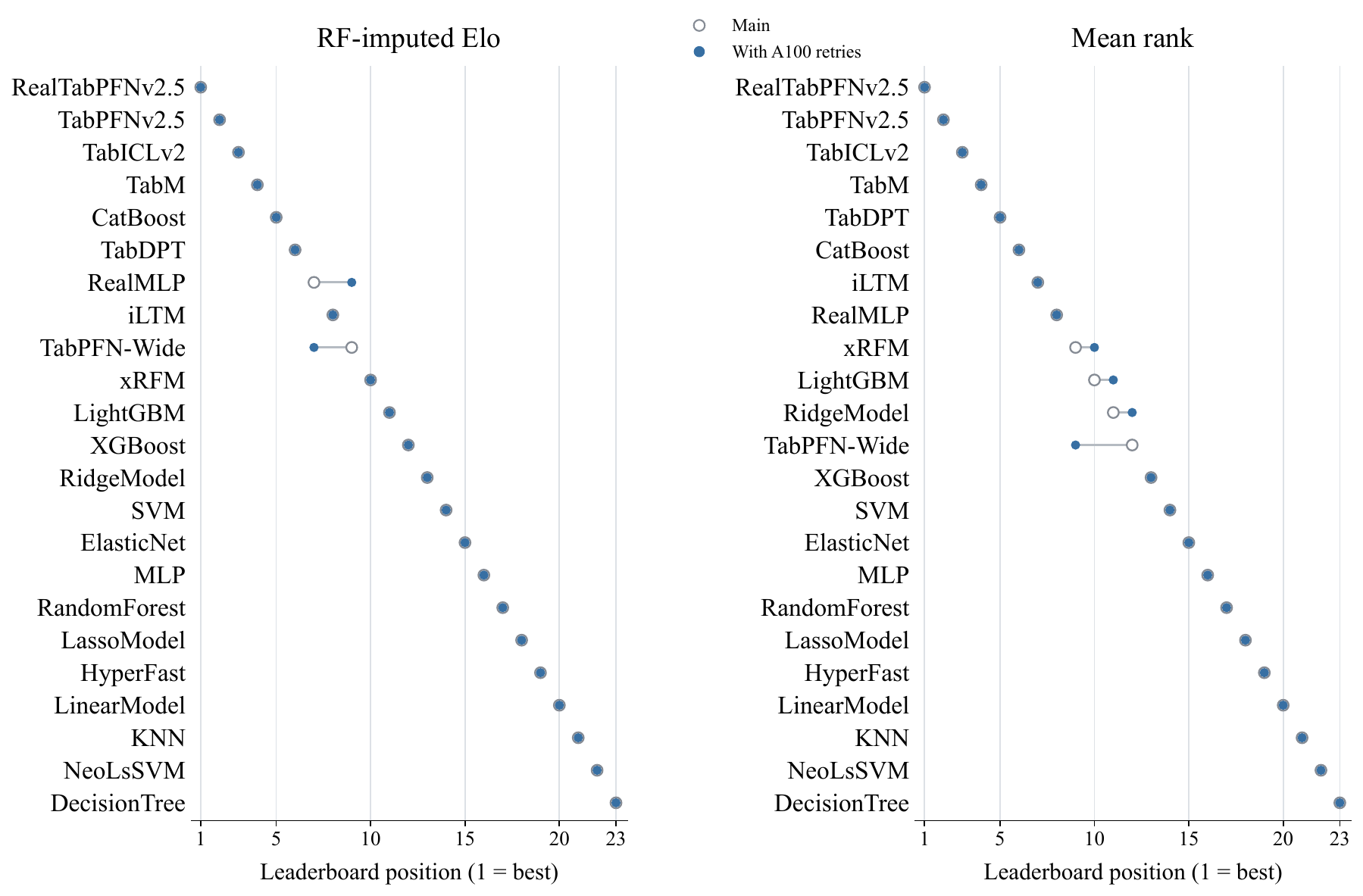}
    \caption{\textbf{A100 retry sensitivity across 78 classification datasets and 23 models.} Open markers denote Main results and filled markers incorporate successful A100 retry results. Left: RF-imputed Elo. Right: mean rank.}
    \label{fig:a100_classification}
\end{figure}

Figures~\ref{fig:a100_all_tasks} and~\ref{fig:a100_classification} compare rankings across all 94 datasets with 21 models and across the 78 classification datasets with 23 models, respectively. Top-five membership is unchanged under both ranking measures in both scopes. Across all 94 datasets, Kendall's $\tau_b=0.990$ for both rankings, with positions shifting by at most one place. RealTabPFNv2.5 and TabICLv2 exchange first and second place by mean rank. In classification, TabPFN-Wide rises from twelfth to ninth by mean rank and from ninth to seventh by Elo, with $\tau_b=0.976$ for both rankings.

\FloatBarrier

\subsection{Cross-Dataset and Global Analyses}
\label{appendix_global_analysis}

We first analyze performance aggregated across all datasets in the benchmark to obtain a global view of model competitiveness under heterogeneous task conditions.

\begin{figure}[h]
    \centering
    \includegraphics[width=1\linewidth]{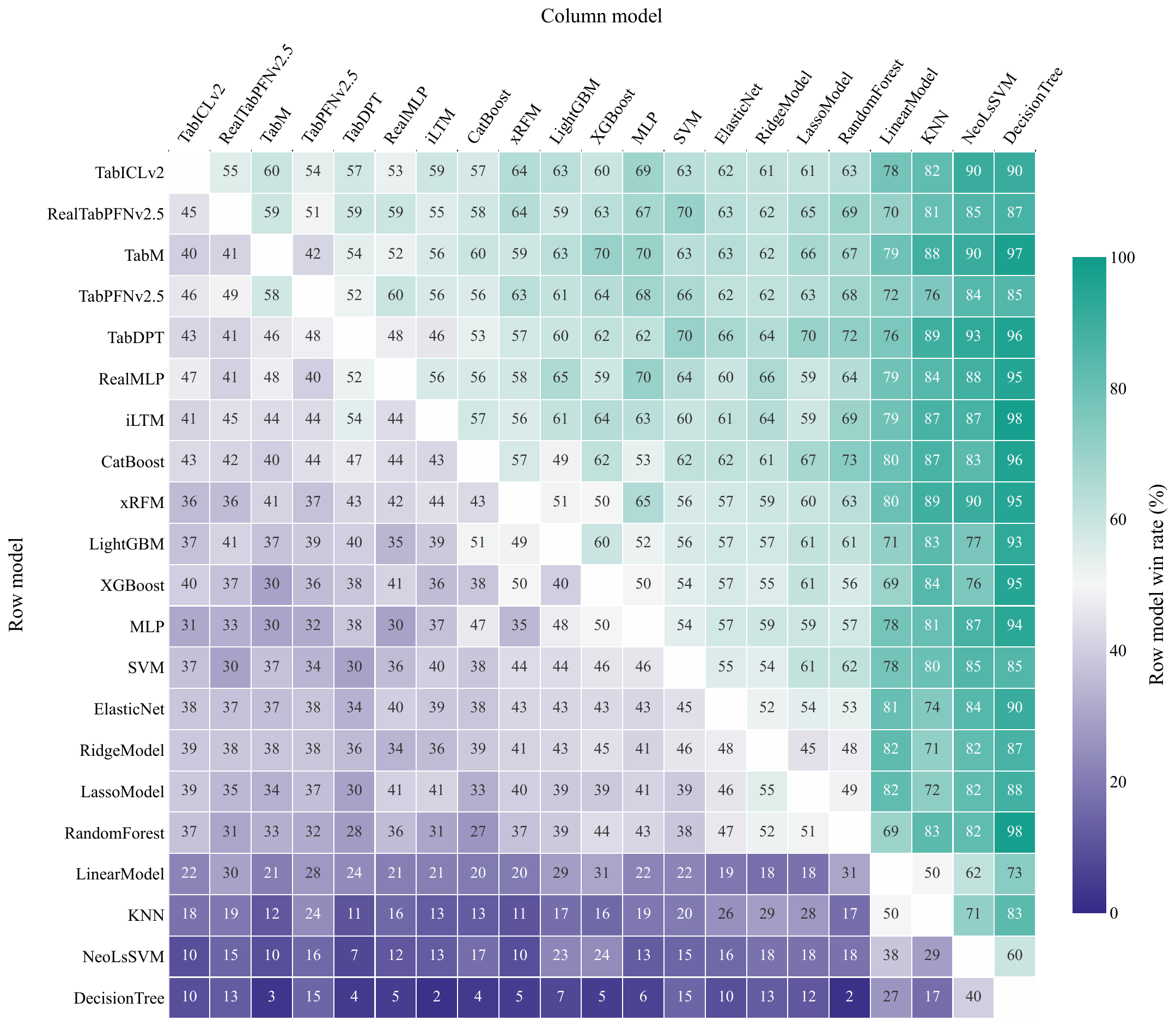}
    \caption{\textbf{Pairwise win-rate matrix across benchmark datasets.} Each cell reports the percentage of datasets on which the row model outperforms the column model under the task-specific primary evaluation metric, with half credit for ties. Higher values indicate stronger relative performance for the row model. Models are ordered by their overall pairwise performance.}
    \label{fig:win_rate_matrix_global}
\end{figure}

\textbf{Global win-rate matrix and pairwise comparisons.}
We construct a pairwise win-rate matrix (Figure~\ref{fig:win_rate_matrix_global}) using the task-specific primary evaluation metric for each dataset. Each entry $(i,j)$ represents the percentage of datasets on which model $i$ outperforms model $j$, with half credit for ties. We use Tuned results when available and Default otherwise. A successful evaluation beats a failure, and two failures are treated as a tie. 

Figure~\ref{fig:win_rate_matrix_global} reveals a clear performance hierarchy across the evaluated models. Foundation-style tabular architectures, particularly \textbf{RealTabPFNv2.5}, \textbf{TabPFNv2.5}, and \textbf{TabICLv2}, occupy the leading positions alongside \textbf{TabM} and achieve consistently high pairwise win rates against most competing methods. \textbf{RealTabPFNv2.5} achieves win rates of 51.1\% against TabPFNv2.5, 45.2\% against TabICLv2, and 58.5\% against TabDPT. These results indicate a strong overall advantage across heterogeneous biomedical prediction tasks, although the margins are narrower among the highest-performing models.

The matrix also reveals performance relationships that are not fully captured by aggregate mean-rank statistics. First, the leading foundation-style models exhibit relatively close pairwise competition, with substantially smaller margins against one another than against lower-performing methods. Second, classical ensemble methods, including \textbf{LightGBM}, \textbf{RandomForest}, \textbf{CatBoost}, and \textbf{XGBoost}, remain competitive against the MLP baseline, despite generally trailing the strongest foundation-style models. This suggests that tree-based ensembles continue to provide effective predictive baselines across diverse biomedical tabular datasets.

In contrast, simpler models, including \textbf{LinearModel}, \textbf{KNN}, \textbf{NeoLsSVM}, and \textbf{DecisionTree}, generally exhibit lower pairwise win rates against higher-performing approaches. Nevertheless, their nonzero win rates indicate that no single model universally dominates across all datasets.

\begin{figure}
    \centering
    \includegraphics[width=\linewidth]{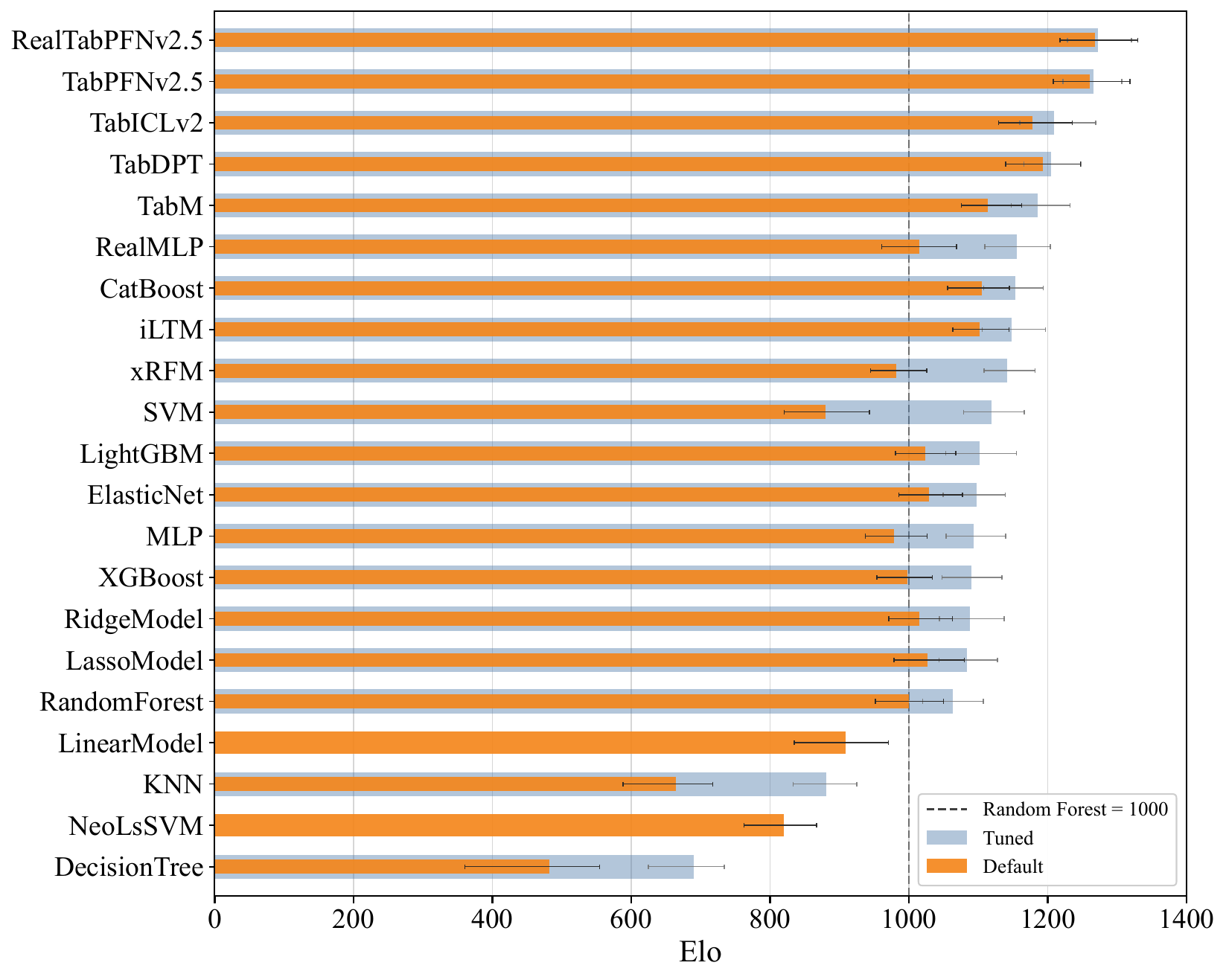}
    \caption{\textbf{TabBioMed leaderboard based on Elo ratings.} Results for 21 models across the 79 datasets with all applicable Default and Tuned evaluations available, without imputation. Error bars denote 95\% dataset-bootstrap confidence intervals.}
    \label{fig:elo_c_complete}
\end{figure}

\textbf{Elo-based global ranking analysis.} In addition to the analysis in Figure \ref{fig:elo_rankings_global}, we compute Elo-style ratings under complete-case evaluation on the 79 datasets with results for all applicable Default and Tuned configurations (Figure~\ref{fig:elo_c_complete}).

The resulting Elo analysis supports the same broad leading group: RealTabPFNv2.5 and TabPFNv2.5 have the highest Tuned Elo point estimates in both analyses, and the nine highest-ranked models are the same. Their ordering changes, with TabM moving from fifth under complete-case evaluation to third under RF imputation, while RealMLP replaces TabDPT in the top five. Four of the top five models are shared. Model-level rankings, using Tuned ratings for HPO models and Default ratings for Default-only models, have Spearman correlation $\rho=0.975$. Because both dataset coverage and missing-score handling differ, this comparison assesses their combined effect.

\subsubsection{Dataset Size and Dimensionality Regimes Across All Tasks}
\label{apx:elo_regime_all_tasks}

We extend the classification-only analysis in Figure~\ref{fig:elo_regimes_classification_23} to all 94 tasks using the 21 models that support classification, regression, and multi-target regression. TabPFN-Wide and HyperFast are excluded. We recompute the median cutpoints over all 94 datasets, yielding 24, 23, 22, and 25 tasks in Q1--Q4, respectively, including 2, 1, 5, and 8 regression or multi-target regression tasks. 
TabICLv2 has the highest Elo point estimate in Q1, followed by RealTabPFNv2.5 and TabPFNv2.5, and in Q3, followed by RealMLP, TabM, RealTabPFNv2.5, and iLTM. In Q2, RealTabPFNv2.5 leads, followed by TabPFNv2.5, TabDPT, and CatBoost. In Q4, RealMLP ranks first, closely followed by iLTM, then TabM.

\begin{figure}[h]
    \centering\includegraphics[width=\linewidth]{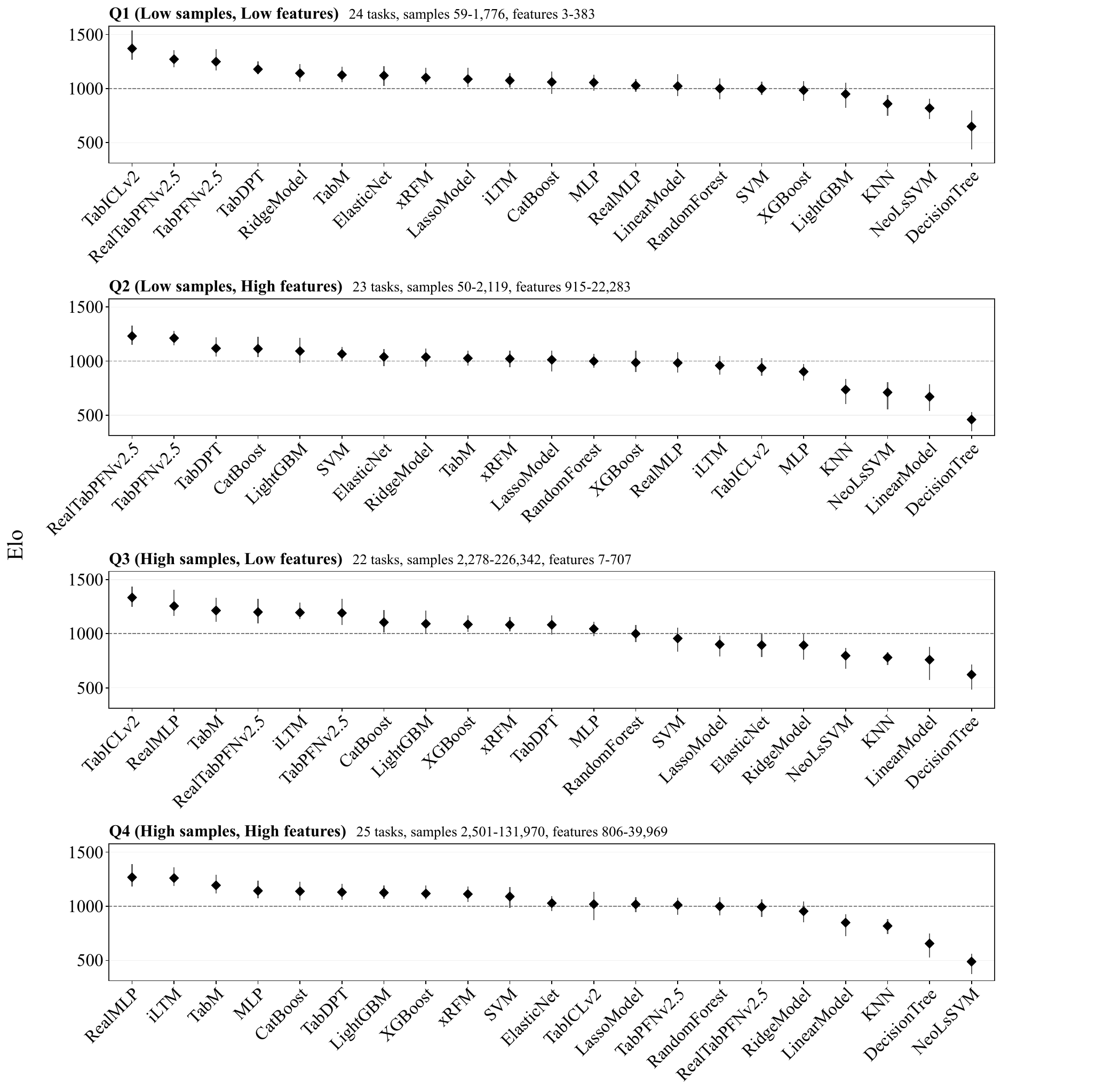}
    \caption{\textbf{Elo across dataset size and dimensionality regimes for 21 models on all 94 tasks.} Median splits over the full benchmark define Q1--Q4, containing 24, 23, 22, and 25 tasks. Each model uses Tuned results when available, falling back to its own Default and then to Default RandomForest imputation. Diamonds show Elo point estimates, bars show 95\% confidence intervals from 200 dataset bootstrap resamples, and the dashed line marks the RandomForest model anchored at 1,000.}
    \label{fig:elo_regimes_all_tasks_21}
\end{figure}

\begin{figure}[H]
    \centering
    \includegraphics[width=0.49\linewidth]{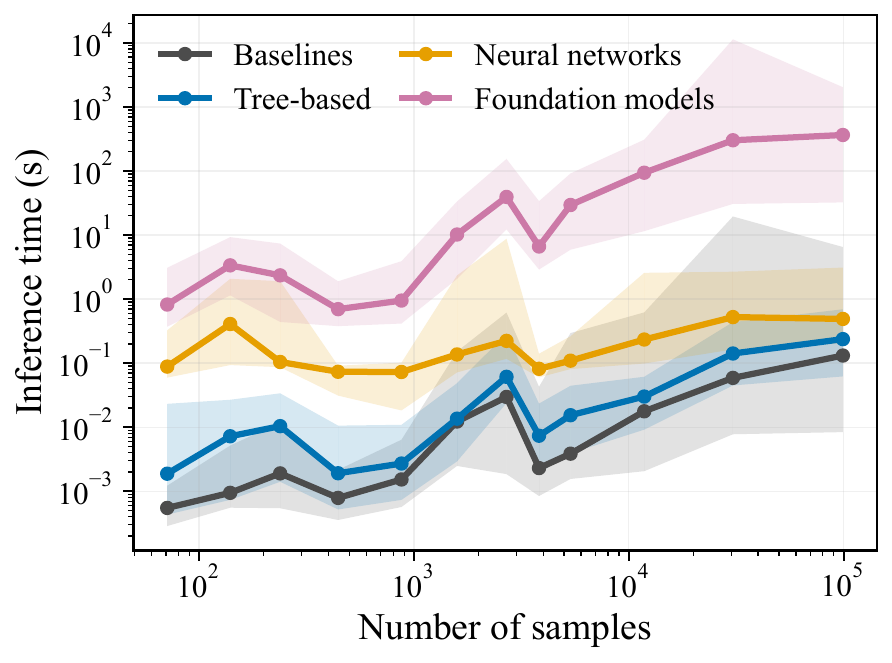}
    \hfill
    \includegraphics[width=0.49\linewidth]{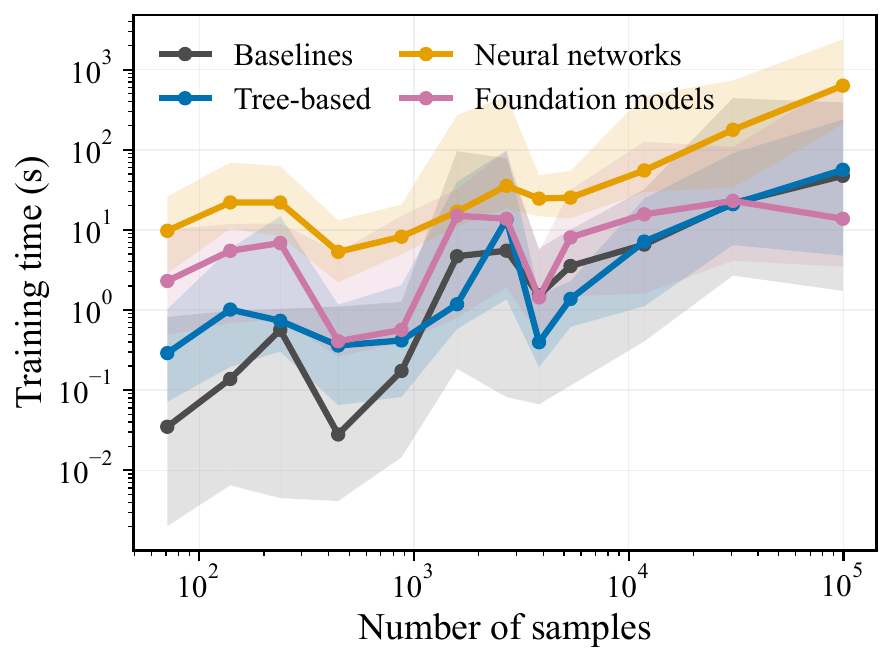}
    \caption{\textbf{Inference vs. training scaling behavior.}
    Comparison of computational cost as a function of the number of samples for inference (left) and training (right) across all model families. Training includes the selected fit and validation prediction. Lines show medians and shading shows interquartile ranges across model--dataset pairs within sample-count bins.}
    \label{fig:inference_training_scaling}
\end{figure}

\textbf{Runtime behavior across dataset size.}
We further investigate how inference and training times vary with dataset size across different model families. Figure~\ref{fig:inference_training_scaling} presents inference time (left) and training time (right) as functions of the number of samples, highlighting differences in computational scaling across model classes.

For inference, baseline, tree-based, and neural network models generally have relatively low median computational costs, whereas foundation models exhibit substantially higher inference times, increasing notably at larger sample sizes. Many of these foundation models perform in-context learning at prediction time.

Training-time behavior exhibits a different pattern. Neural networks have the highest median training times across sample sizes. Baseline and tree-based methods generally remain computationally efficient. Foundation models have intermediate median training times on smaller datasets, becoming comparable to or faster than baselines and trees in the largest sample-size bins.
The shaded regions further indicate considerable variability in computational costs across dataset sizes and model families.

\subsection{Model Performance Across Feature-Count Regimes}
\label{apx:feature_count_analysis}

To further investigate the influence of feature dimensionality on model performance, we stratified the 94 datasets into three groups according to their number of input features: low-feature (3--150 features; $n=31$), medium-feature (167--3,068 features; $n=31$), and high-feature (3,312--39,969 features; $n=32$). Each feature-count group includes datasets with widely varying sample sizes. This analysis complements the joint feature-count and sample-size analysis presented in Figure~\ref{fig:elo_regimes_all_tasks_21} by examining how relative model performance and execution reliability vary across feature-dimensionality regimes.

\begin{figure*}[ht]
    \centering
    \includegraphics[width=0.95\textwidth]{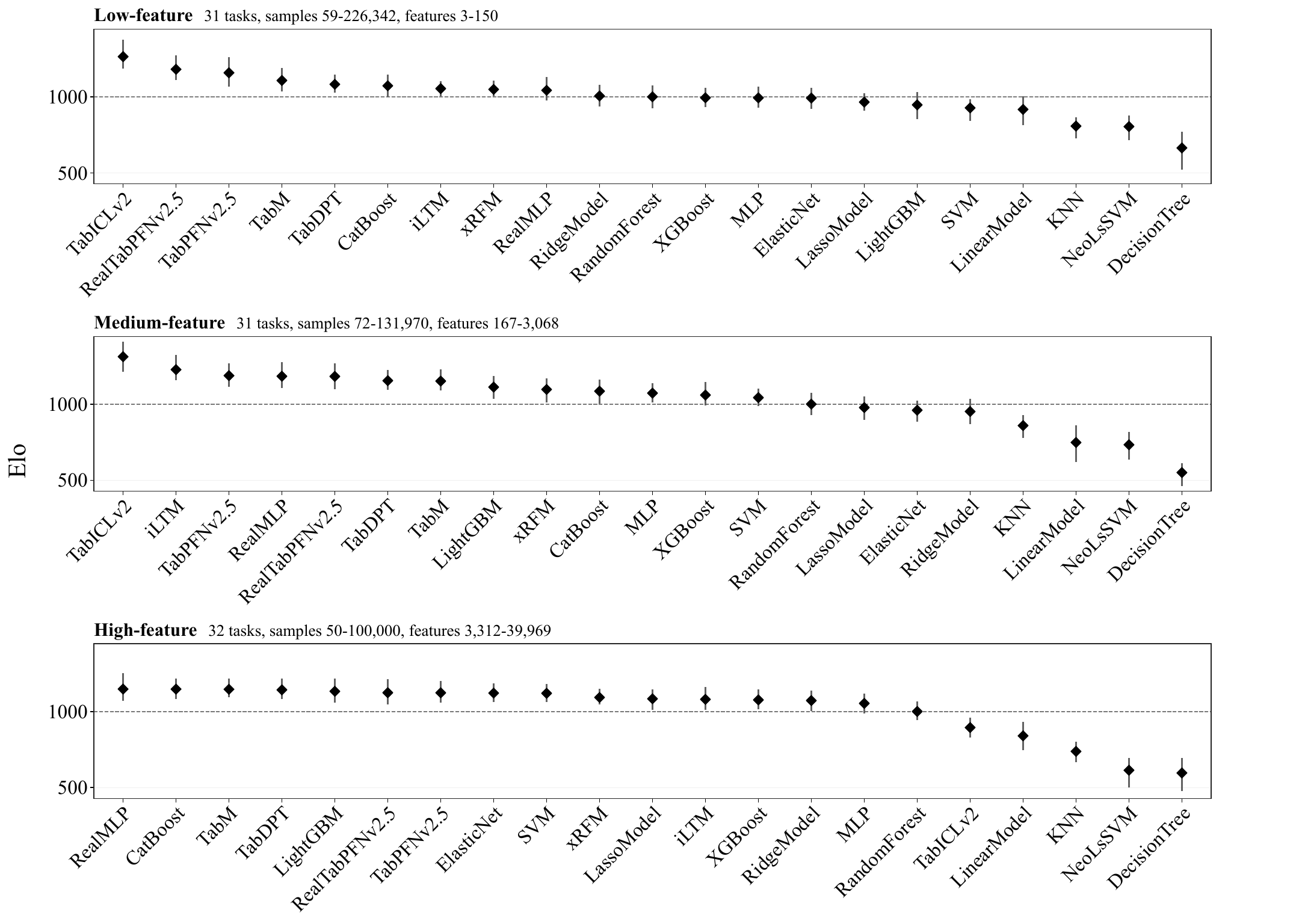}
    \caption{\textbf{Elo across feature-dimensionality regimes for 21 models on all 94 tasks.} Datasets are divided into low-feature (3--150 features; $n=31$), medium-feature (167--3,068 features; $n=31$), and high-feature (3,312--39,969 features; $n=32$) terciles. Within each regime, models are ranked by their Elo ratings, with higher ratings indicating stronger relative predictive performance. Each model uses Tuned results when available, falling back to its own Default and then to Default Random Forest imputation. Diamonds show Elo point estimates, bars show 95\% confidence intervals from 200 dataset bootstrap resamples, and the dashed line marks the RandomForest model rating of 1,000.}
    \label{fig:feature_tercile_performance}
\end{figure*}

Figure~\ref{fig:feature_tercile_performance} presents the calibrated Elo ratings of the 21 evaluated models across the three feature regimes. On low-feature datasets, tabular foundation models occupy the leading positions, with TabICLv2 achieving the highest Elo rating (1,263.9), followed by RealTabPFNv2.5 (1,180.9) and TabPFNv2.5 (1,157.8). TabICLv2 completes all 31 datasets, while RealTabPFNv2.5 and TabPFNv2.5 each achieve a completion rate of 96.77\%. A similar pattern emerges in the medium-feature regime, where TabICLv2 maintains the highest Elo rating (1,311.7), followed by iLTM (1,226.8), TabPFNv2.5 (1,187.4), RealMLP (1,183.3), and RealTabPFNv2.5 (1,181.9). TabICLv2, iLTM and RealMLP achieve full completion in this regime, whereas TabPFNv2.5 and RealTabPFNv2.5 each complete 83.87\% of datasets. These results indicate that tabular foundation models remain highly competitive on datasets with low to moderate feature dimensionality.

In the high-feature regime, the relative ordering changes substantially, with neural networks and tree-based methods becoming more prominent. RealMLP achieves the highest Elo rating (1,147.5), followed by CatBoost (1,146.8), TabM (1,145.6), TabDPT (1,141.8), and LightGBM (1,132.5). RealMLP, TabM, TabDPT and LightGBM complete all 32 datasets, while CatBoost completes 31 of 32. In comparison, TabPFNv2.5 and RealTabPFNv2.5 obtain Elo ratings of 1,122.7 and 1,123.3, respectively, with completion rates of 84.38\% and 87.50\%.

To assess the stability of model rankings across feature-dimensionality regimes, we additionally computed pairwise Kendall rank correlations between the Elo-based orderings of all 21 models. Ranking agreement is strongest between low- and medium-feature datasets ($\tau=0.667$, $p<0.0001$), followed by medium- and high-feature datasets ($\tau=0.448$, $p=0.0040$). Agreement is weaker between low- and high-feature datasets ($\tau=0.400$, $p=0.0109$). 

Foundation models occupy the leading positions in the low- and medium-feature regimes, while the high-feature regime has a more diverse set of leading neural, tree-based and foundation models.

\subsection{Effect of PCA Preprocessing on High-Dimensional Tabular Foundation Models}
\label{apx:pca_tfm}

We compare external PCA followed by TabICLv2, TabPFNv2.5, or RealTabPFNv2.5 with the corresponding model without external PCA on all 36 benchmark datasets with more than 2,000 features (range: 2,018--39,969).

\paragraph{Preprocessing and tuning.}
We use at most 500 components, limited further by the training-row count and number of nonconstant input columns. This cap matches the 500-feature per-estimator subsampling threshold used by TabPFNv2.5 and RealTabPFNv2.5 \citep{grinsztajn2025tabpfn}. Missing values use training-column means, with zero for entirely unobserved columns. Imputation and PCA are fitted within each training partition, including inner HPO folds, and applied to held-out rows.

We retain matched model-specific search spaces, test splits, seeds, and HPO budgets to the original non-PCA runs: six hours per fixed split or two hours per outer fold, with at most 100 sampled configurations. We compare results retaining native preprocessing. TabDPT and iLTM already include native PCA mechanisms \citep{ma2024tabdpt,bonet2025iltm}.

\begin{figure}[t]
    \centering
    \includegraphics[width=0.7\linewidth]{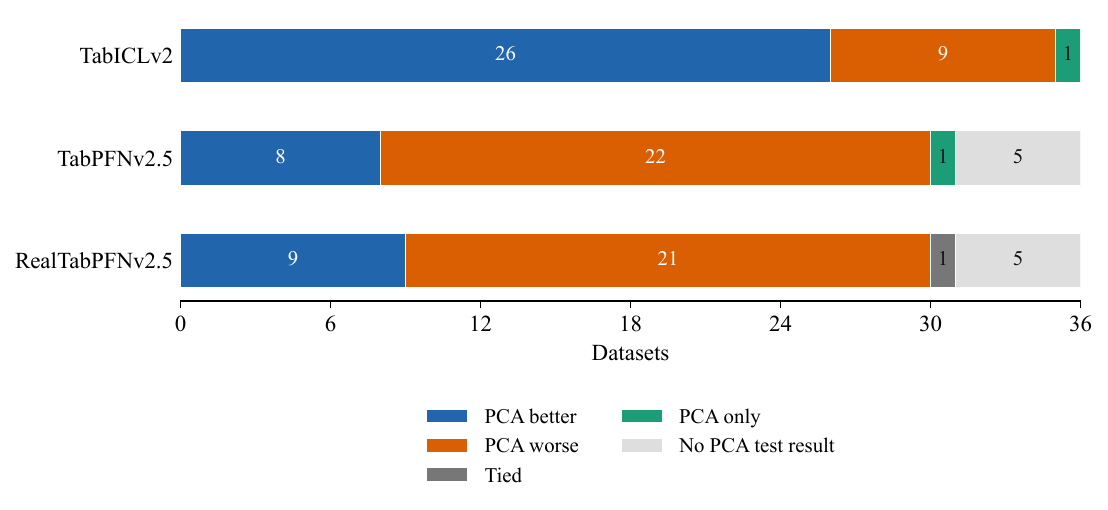}
    \caption{\textbf{Effect and coverage of external PCA across 36 high-dimensional datasets.} Paired comparisons improve (blue), degrade (orange), or tie (dark gray) with PCA. Green denotes successful PCA-only evaluations, and light gray denotes unavailable results.}
    \label{fig:pca_paired_outcomes}
\end{figure}

\paragraph{Evaluation protocol.}
Binary classification uses ROC-AUC, multiclass classification uses log loss, and regression and multitarget regression use MSE. For each model, we use Tuned results when available and its own Default otherwise, independently for the PCA and non-PCA runs. We orient differences so positive values favor PCA (Table~\ref{tab:pca_tfm_deltas}). Figure~\ref{fig:pca_paired_outcomes} summarizes outcomes and coverage.

\begingroup
\small
\setlength{\tabcolsep}{1.5pt}
\renewcommand{\arraystretch}{1.08}
\begin{longtable}{@{}lcrrrrr@{}}
\caption{Effect of external PCA on test performance relative to the same model without PCA. 
Positive $\Delta$ favors PCA: PCA minus non-PCA for ROC-AUC, and non-PCA minus
PCA for log loss and MSE. The asterisks on \texttt{hnoca\_cells30} and
\texttt{tahoe\_g2m\_score} mark comparisons against RandomForest Default because
the corresponding non-PCA results are unavailable. These references are excluded from paired counts.
Dashes indicate unavailable results.}
\label{tab:pca_tfm_deltas}\\
\toprule
Dataset & Task & Samples & Features &
\makecell{TabICLv2~$\Delta$} &
\makecell{TabPFNv2.5~$\Delta$} &
\makecell{RealTabPFNv2.5~$\Delta$} \\
\midrule
\endfirsthead
\multicolumn{7}{l}{\textit{Table continued from previous page}}\\
\toprule
Dataset & Task & Samples & Features &
\makecell{TabICLv2~$\Delta$} &
\makecell{TabPFNv2.5~$\Delta$} &
\makecell{RealTabPFNv2.5~$\Delta$} \\
\midrule
\endhead
\midrule
\multicolumn{7}{r}{\textit{Continued on next page}}\\
\endfoot
\bottomrule
\endlastfoot
\texttt{SIDER\_gastro} & Bin & 1,427 & 2,018 & -0.0570 & -0.0543 & 0.0000 \\
\texttt{SIDER\_nervous} & Bin & 1,427 & 2,018 & 0.0205 & -0.1428 & -0.1125 \\
\texttt{ST001000} & Bin & 121 & 2,177 & -0.0185 & -0.0200 & -0.0107 \\
\texttt{true\_fluorescence} & Reg & 51,738 & 3,068 & 0.8172 & -- & -- \\
\texttt{lung} & MultiC & 203 & 3,312 & 0.0079 & 0.0050 & -0.0176 \\
\texttt{GLIOMA} & MultiC & 50 & 4,434 & 0.0152 & -0.3004 & -0.3875 \\
\texttt{avida\_htnfa\_3mer} & Bin & 5,580 & 5,000 & -0.0199 & -0.0135 & 0.0140 \\
\texttt{tahoe\_batch\_integr.} & MultiC & 28,000 & 5,000 & 0.4841 & -- & -- \\
\texttt{tahoe\_cell\_name\_5000} & MultiC & 21,426 & 5,000 & 1.0084 & -- & -- \\
\texttt{tahoe\_g2m\_score} & Reg & 21,426 & 5,000 & 0.0153 & 0.0180* & 0.0196 \\
\texttt{TOX\_171} & MultiC & 171 & 5,748 & -0.3759 & -0.2524 & -0.3067 \\
\texttt{Prostate\_GE} & Bin & 102 & 5,966 & 0.0602 & -0.0104 & -0.0102 \\
\texttt{avida\_sarscov2\_3mer} & Bin & 26,979 & 7,072 & 0.1135 & -0.0354 & -0.0614 \\
\texttt{ALLAML} & Bin & 72 & 7,129 & 0.0027 & -0.0240 & -0.0240 \\
\texttt{proteinea\_solubility} & Bin & 71,421 & 8,000 & 0.1267 & -- & -- \\
\texttt{sc\_gastrulation\_stage} & Reg & 10,000 & 10,000 & -0.1259 & -0.1472 & -0.1383 \\
\texttt{CLL\_SUB\_111} & MultiC & 111 & 11,340 & 0.0990 & -0.0166 & -0.0033 \\
\texttt{cptac\_histology} & Bin & 101 & 12,592 & 0.0394 & -0.0206 & -0.0100 \\
\texttt{MultiOmics\_GS-LGG} & MultiC & 247 & 13,242 & -0.0283 & -0.0524 & -0.0841 \\
\texttt{MultiOmics\_GS-OV} & MultiC & 284 & 13,501 & 0.0916 & -0.1053 & -0.0796 \\
\texttt{MultiOmics\_GS-GBM} & MultiC & 244 & 14,640 & 0.2139 & -0.0170 & -0.0211 \\
\texttt{MultiOmics\_GS-BRCA} & MultiC & 671 & 15,200 & -0.0874 & -0.1016 & -0.1378 \\
\texttt{MultiOmics\_GS-COAD} & MultiC & 260 & 15,200 & 0.0694 & -0.0413 & -0.0505 \\
\texttt{sc\_muris\_age} & MultiC & 1,817 & 16,436 & 0.5459 & -0.0786 & -0.0852 \\
\texttt{sc\_muris\_celltype} & MultiC & 1,817 & 16,472 & 0.5517 & 0.0748 & 0.1104 \\
\texttt{sexcheck\_chrX\_50k} & Bin & 12,078 & 17,118 & -0.0354 & -0.0411 & -0.0339 \\
\texttt{sc\_muraro\_celltype} & MultiC & 2,119 & 18,508 & 0.7748 & 0.0940 & 0.1101 \\
\texttt{SMK\_CAN\_187} & Bin & 187 & 19,993 & 0.0325 & -0.0368 & -0.0334 \\
\texttt{metabric\_grade} & MultiC & 1,892 & 20,603 & -0.0374 & -0.0134 & -0.0103 \\
\texttt{hnoca\_cells30} & MultiC & 100,000 & 22,000 & 1.7134* & -- & -- \\
\texttt{GLI\_85} & Bin & 85 & 22,283 & 0.0027 & -0.0142 & -0.0045 \\
\texttt{continental\_50k\_missing} & MultiC & 3,309 & 25,921 & 1.2850 & 0.3879 & 0.4134 \\
\texttt{subpop\_50k} & MultiC & 2,501 & 32,339 & 1.9858 & 0.7388 & 0.7690 \\
\texttt{pca\_50k} & MultiReg & 2,609 & 34,635 & 45.9369 & 23.9975 & 23.6644 \\
\texttt{continental\_50k} & MultiC & 3,309 & 39,969 & 1.0743 & 0.1821 & 0.1913 \\
\texttt{coords\_50k} & MultiReg & 3,309 & 39,969 & 559.4015 & 314.0823 & 337.1634 \\
\end{longtable}
\endgroup

TabICLv2 improves on 26 of 35 paired datasets (74.3\%) and degrades on 9. It also yields a test result for \texttt{hnoca\_cells30}, where its non-PCA baseline failed. TabPFNv2.5 improves on 8 of 30 paired datasets (26.7\%), yields a test result for \texttt{tahoe\_g2m\_score}, and degrades on 22, while RealTabPFNv2.5 improves on 9 of 31 (29.0\%) and degrades on 21, and ties on one. The TabPFN variants agree on the direction of change on 27 of their 30 shared paired datasets (90.0\%). 
TabICLv2's lack of native PCA/SVD, compared with the TabPFN variants' feature subsampling and optional SVD features, may explain its greater benefit from external PCA, which remains a hypothesis since the comparison does not isolate the mechanism or establish an optimal PCA component count.

\subsection{Domain-wise Analysis}
\label{sec:domain_wise_analysis_appendix}

We compare the same 21 models within each domain using Elo leaderboards. Panels report dataset counts and sample and feature ranges. 
Scores use Tuned test results when available, otherwise the model's own Default results. If neither is available, the missing score is imputed using the same task's Default RandomForest score. Point estimates weight datasets equally. Where shown, the 95\% intervals use 200 bootstrap resamples, with datasets sharing samples kept together (Appendix~\ref{apx:source_dependence}).

The domain-wise Pareto analyses use the same 21 models and RF-imputation policy as the main Elo analysis, with ratings fitted separately within each domain. Missing evaluations inherit RF Default scores and timings.

\subsubsection{Clinical and Healthcare}
\label{sec:domain_ehr}

\begin{figure}[!htbp]
    \centering
    \includegraphics[width=\linewidth]{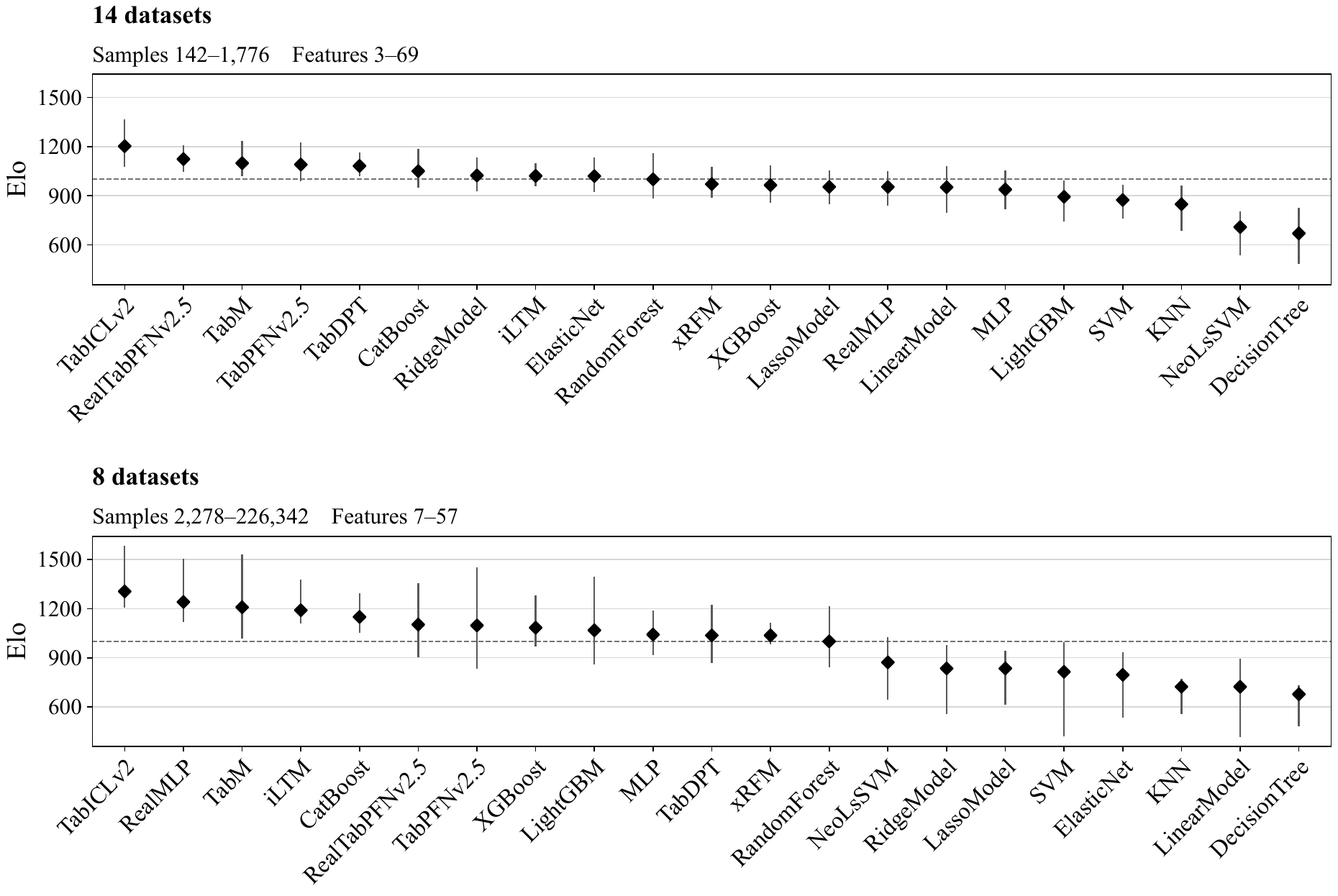}
    \caption{\textbf{Model-level Elo for Clinical and Healthcare datasets.} The two panels contain 14 tasks with 142--1,776 samples and eight with 2,278--226,342 samples, comparing the same 21 models. Diamonds show point estimates, bars show 95\% cluster-bootstrap intervals, and the dashed line marks the RandomForest reference of 1,000.}
    \label{fig:domain_norm_ehr}
\end{figure}

We show results for Clinical and Healthcare datasets  divided at 2,000 samples into groups of 14 and eight tasks. TabICLv2 achieves the highest Elo point estimate in both clinical sample-size groups (Figure~\ref{fig:domain_norm_ehr}). RealMLP, TabM and iLTM follow in the large sample size group, while RealTabPFNv2.5 and TabM follow in the small sample size group.

\begin{figure}[!htbp]
    \centering
    \includegraphics[width=\linewidth]{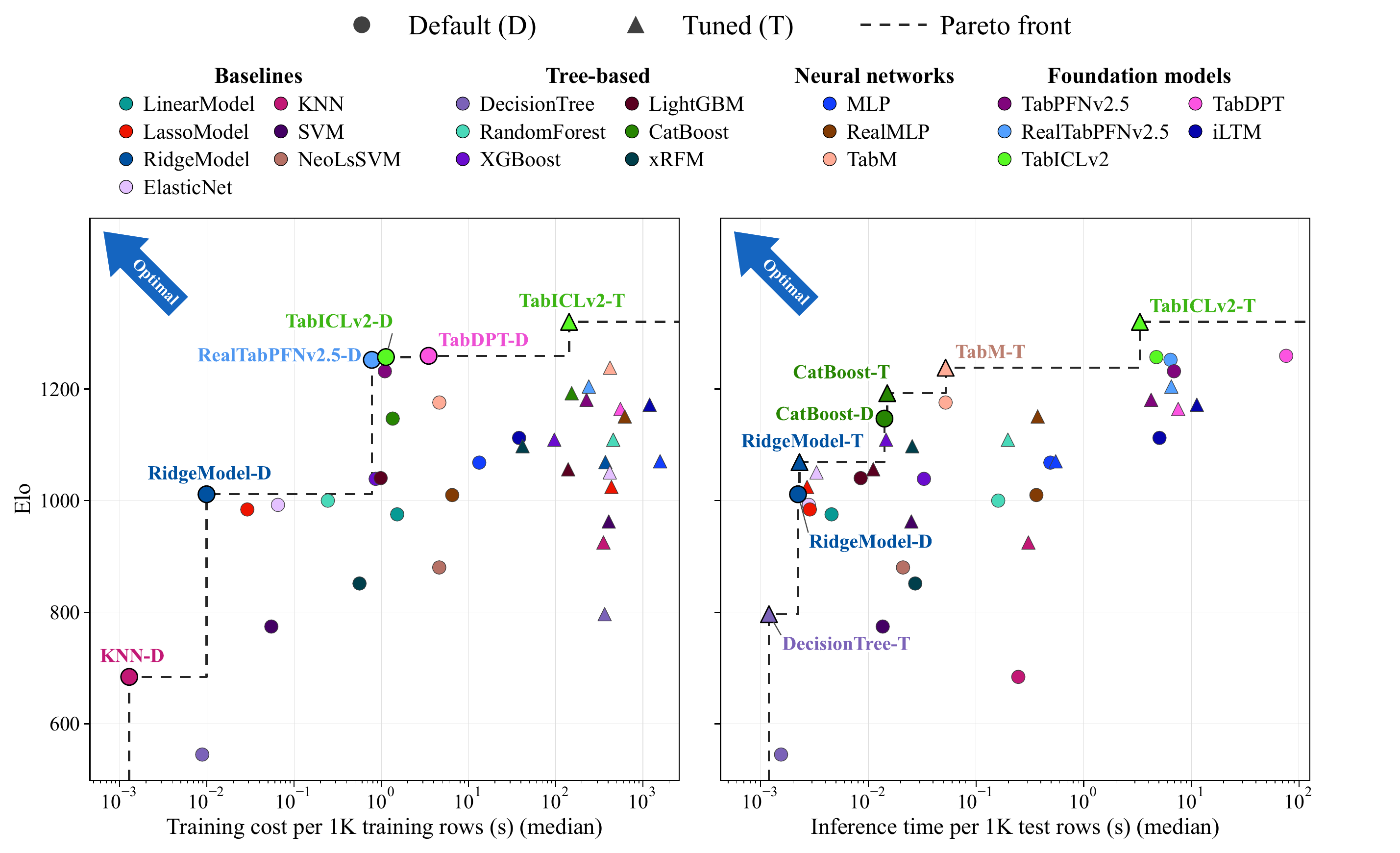}
    \caption{\textbf{Pareto trade-offs for Clinical and Healthcare datasets.} The left panel shows training cost trade-offs, while the right panel shows inference-time trade-offs.}
    \label{fig:pareto_ehr}
\end{figure}

Tuned TabICLv2 reaches the highest Elo at high runtimes (Figure~\ref{fig:pareto_ehr}). Default foundation configurations (RealTabPFNv2.5, TabICLv2) offer lower training costs, while RidgeModel provides low inference latency and CatBoost and TabM offer intermediate performance--latency trade-offs.

\FloatBarrier
\subsubsection{Drug Discovery and Molecular Properties}
\label{sec:domain_drug}

\begin{figure}[!htbp]
    \centering
    \includegraphics[width=\linewidth,height=0.78\textheight,keepaspectratio]{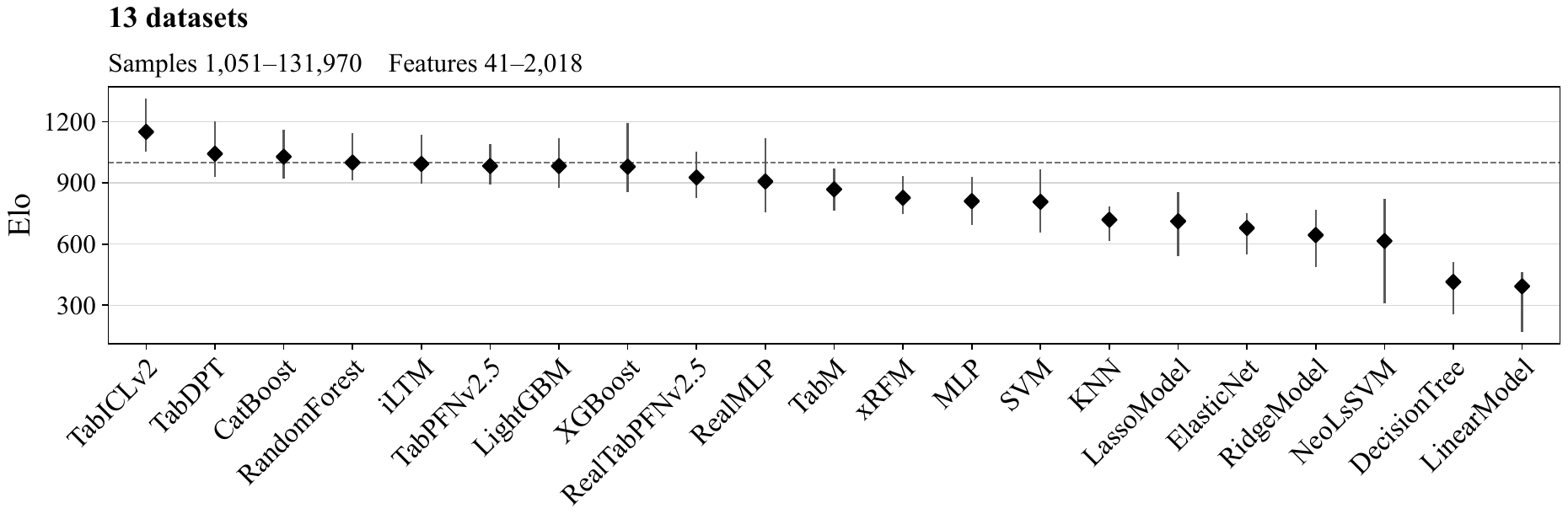}
    \caption{\textbf{Model-level Elo for Drug Discovery and Molecular Properties datasets.} The 21 models are compared across all 13 tasks. Diamonds show point estimates, bars show 95\% cluster-bootstrap intervals, and the dashed line marks the RandomForest reference of 1,000.}
    \label{fig:domain_norm_drug}
\end{figure}

Across all 13 drug-discovery tasks, TabICLv2 also achieves the highest Elo point estimate (Figure~\ref{fig:domain_norm_drug}). TabDPT and CatBoost rank next. The pooled comparison covers eight binary classification and five regression tasks.

\begin{figure}[!htbp]
    \centering
    \includegraphics[width=\linewidth]{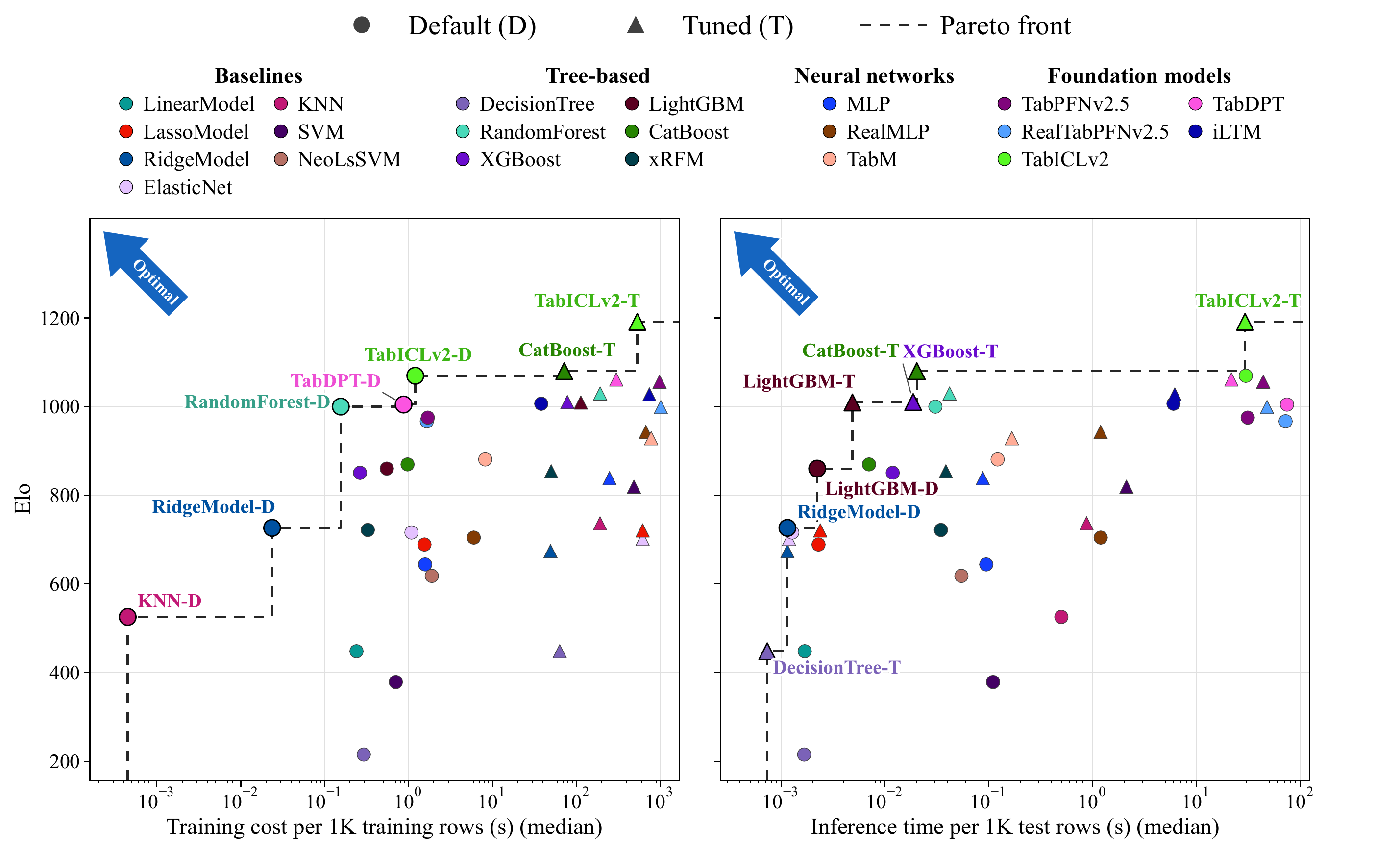}
    \caption{\textbf{Pareto trade-offs for Drug Discovery datasets.} The left panel shows training cost trade-offs, while the right panel shows inference-time trade-offs.}
    \label{fig:pareto_drug}
\end{figure}

Default TabDPT and TabICLv2 offer training-cost trade-offs alongside Tuned CatBoost and TabICLv2 (Figure~\ref{fig:pareto_drug}). Tuned LightGBM, XGBoost and CatBoost provide intermediate inference-cost trade-offs, while Tuned TabICLv2 reaches the highest Elo rating at greater inference cost.

\FloatBarrier
\subsubsection{Genomics}
\label{sec:domain_genom}

\begin{figure}[!htbp]
    \centering
    \includegraphics[width=\linewidth]{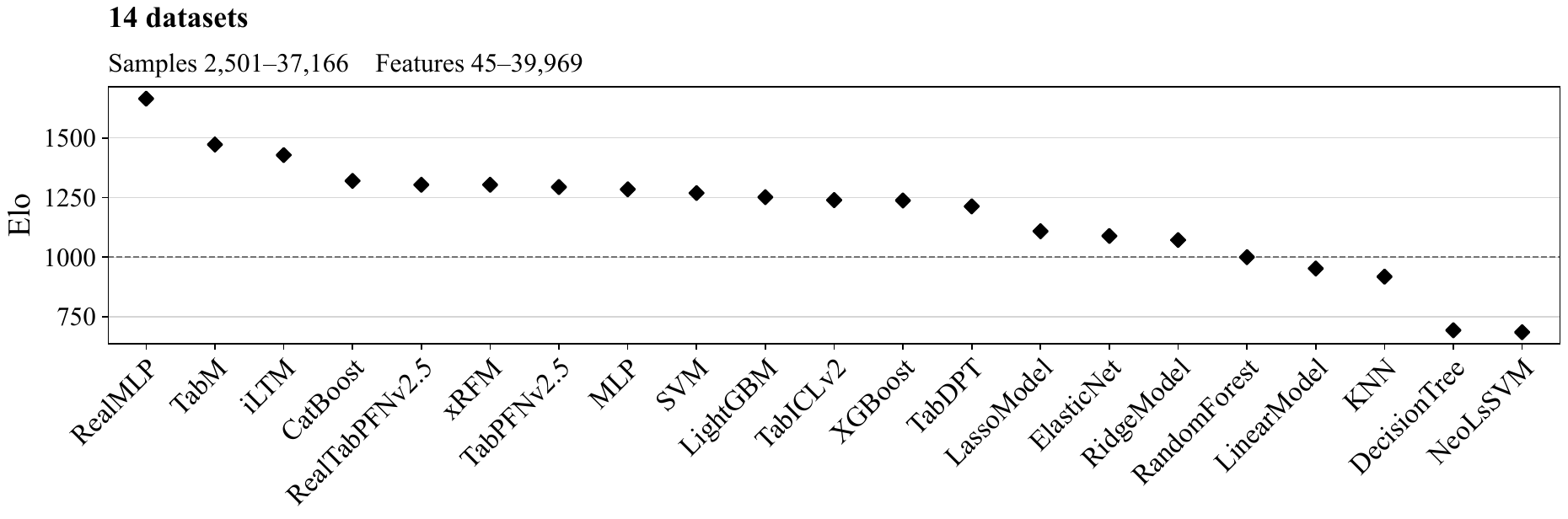}
    \caption{\textbf{Model-level Elo for Genomics datasets.} Point estimates for 21 models across 14 tasks. Twelve tasks derive from the same 1KG+HGDP individuals. The dashed line marks the RandomForest reference of 1,000.}
    \label{fig:domain_norm_genom}
\end{figure}

RealMLP achieves the highest Elo point estimate across the genomic tasks, followed by TabM and iLTM (Figure~\ref{fig:domain_norm_genom}). 

\begin{figure}[!htbp]
    \centering
    \includegraphics[width=\linewidth]{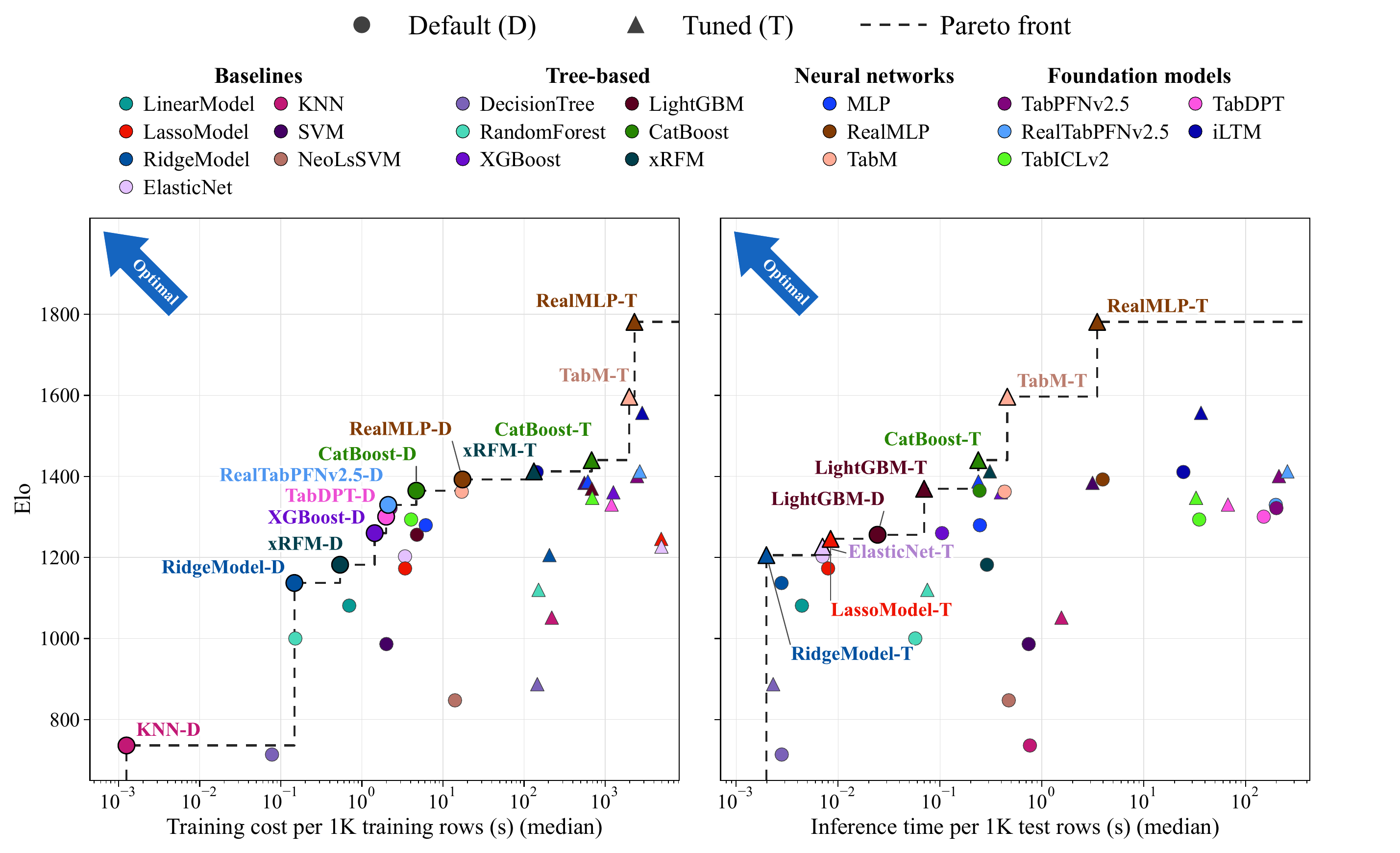}
    \caption{\textbf{Pareto trade-offs for Genomics datasets.} The left panel shows training cost trade-offs, while the right panel shows inference-time trade-offs.}
    \label{fig:pareto_genom}
\end{figure}

Tuned RealMLP reaches the highest Elo rating in the Pareto analysis, with greater training and inference costs (Figure~\ref{fig:pareto_genom}). Linear models offer low inference latency, while tuned GBDTs (LightGBM and CatBoost) and TabM provide intermediate trade-offs.

\FloatBarrier
\subsubsection{Metabolomics}
\label{sec:domain_mtbls}

\begin{figure}[!htbp]
    \centering
    \includegraphics[width=\linewidth]{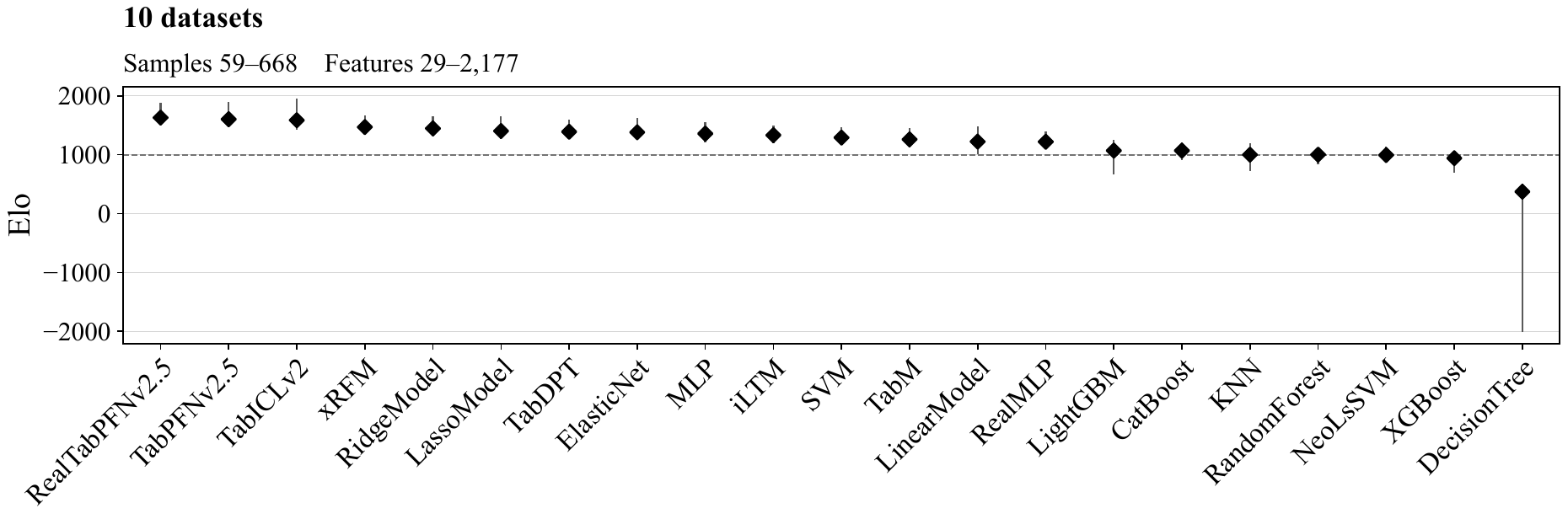}
    \caption{\textbf{Model-level Elo for Metabolomics datasets.} The 21 models are compared across all 10 tasks. Diamonds show point estimates, bars show 95\% cluster-bootstrap intervals, and the dashed line marks the RandomForest reference of 1,000.}
    \label{fig:domain_norm_mtbls}
\end{figure}

The 10 metabolomics datasets contain 59--668 samples (Figure~\ref{fig:domain_norm_mtbls}). The two TabPFN variants and TabICLv2 occupy the leading positions, with xRFM and RidgeModel providing competitive alternatives.

\begin{figure}[!htbp]
    \centering
    \includegraphics[width=\linewidth]{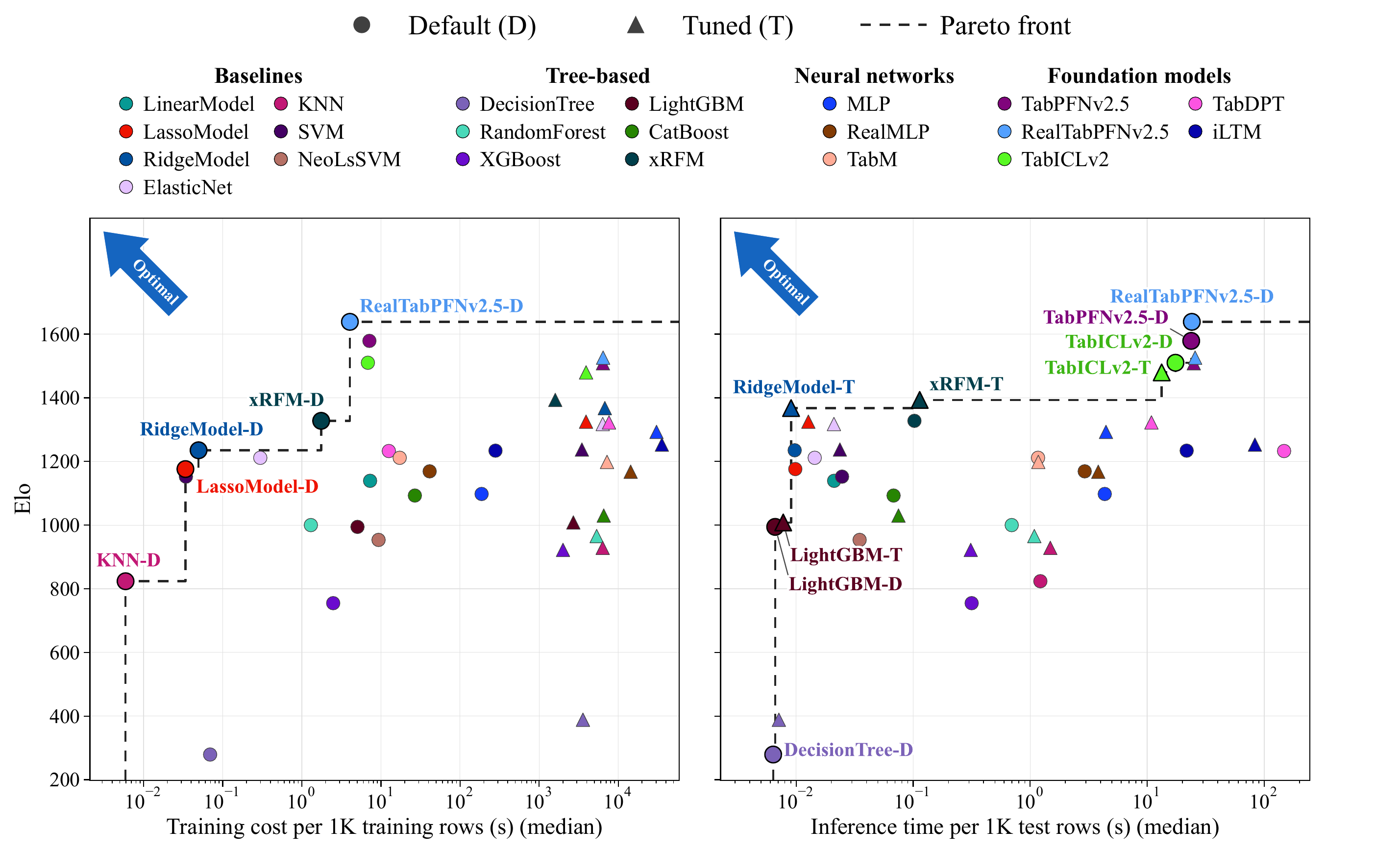}
    \caption{\textbf{Pareto trade-offs for Metabolomics datasets.} The left panel shows training cost trade-offs, while the right panel shows inference-time trade-offs.}
    \label{fig:pareto_mtbls}
\end{figure}

The training-cost frontier consists entirely of Default configurations, with RealTabPFNv2.5 achieving the highest Elo rating (Figure~\ref{fig:pareto_mtbls}). Linear baselines offer lower-cost alternatives, and tuned xRFM provides an intermediate inference-cost trade-off before the higher-performing foundation model configurations.

\FloatBarrier
\subsubsection{Proteomics}
\label{sec:domain_ptomics}

\begin{figure}[!htbp]
    \centering
    \includegraphics[width=\linewidth]{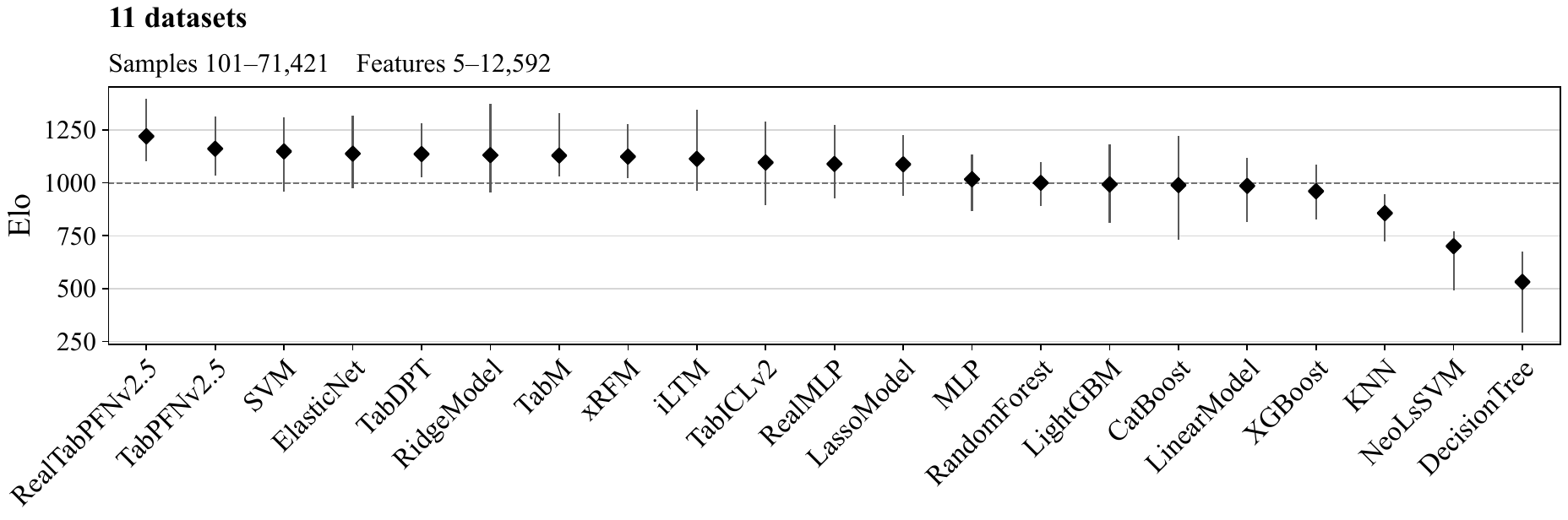}
    \caption{\textbf{Model-level Elo for Proteomics datasets.} The 21 models are compared across all 11 tasks. Diamonds show point estimates, bars show 95\% cluster-bootstrap intervals, and the dashed line marks the RandomForest reference of 1,000.}
    \label{fig:domain_norm_ptomics}
\end{figure}

The proteomics comparison spans 101--71,421 samples and 5--12,592 features (Figure~\ref{fig:domain_norm_ptomics}). RealTabPFNv2.5 has the highest Elo point estimate, followed by TabPFNv2.5, while SVM and ElasticNet achieve ratings close to the latter.

\begin{figure}[!htbp]
    \centering
    \includegraphics[width=\linewidth]{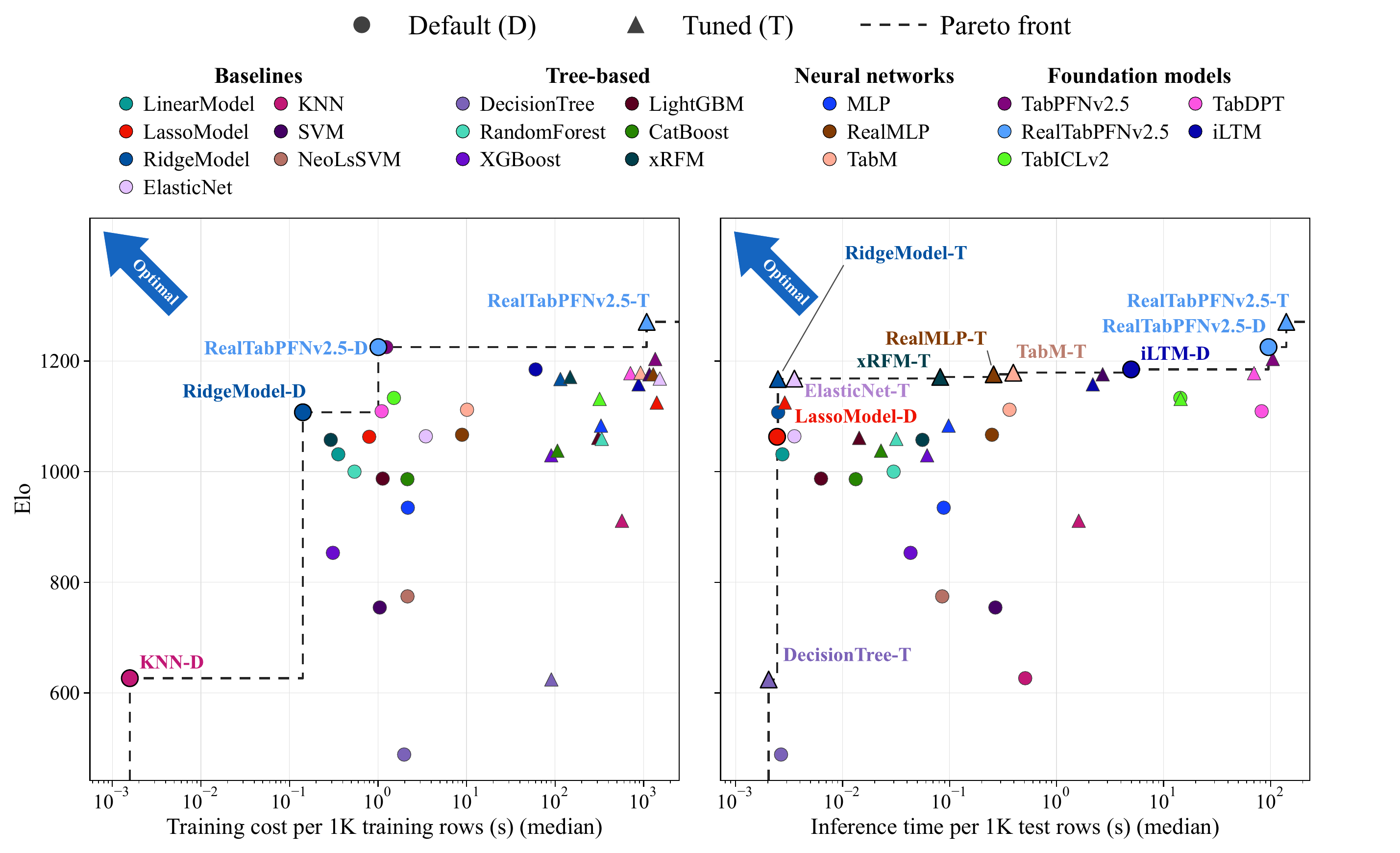}
    \caption{\textbf{Pareto trade-offs for Proteomics datasets.} The left panel shows training cost trade-offs, while the right panel shows inference-time trade-offs.}
    \label{fig:pareto_ptomics}
\end{figure}

The default RealTabPFNv2.5 configuration lies on both training and inference frontiers (Figure~\ref{fig:pareto_ptomics}). RidgeModel provides a lower-cost alternative on the training frontier, while xRFM, RealMLP, TabM, and iLTM offer intermediate trade-offs with lower inference latency.

\FloatBarrier
\subsubsection{Single-Cell}
\label{sec:domain_sc}

\begin{figure}[!htbp]
    \centering
    \includegraphics[width=\linewidth]{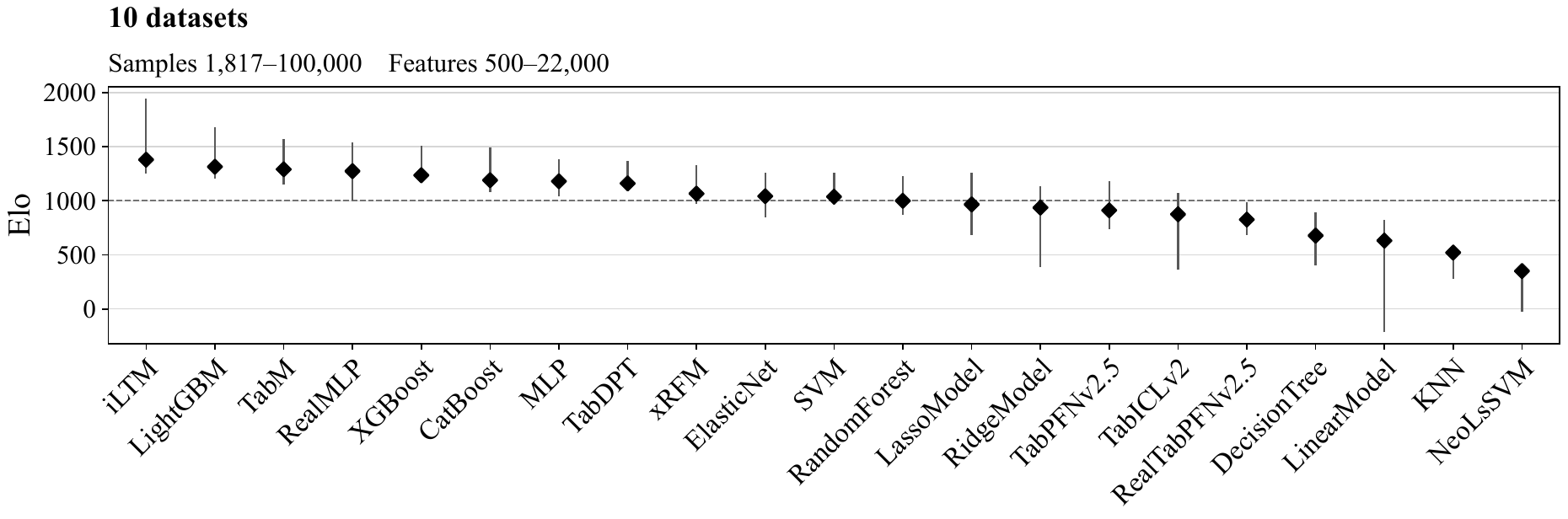}
    \caption{\textbf{Model-level Elo for Single-Cell datasets.} The 21 models are compared across all 10 tasks. Diamonds show point estimates, bars show 95\% cluster-bootstrap intervals, and the dashed line marks the RandomForest reference of 1,000.}
    \label{fig:domain_norm_sc}
\end{figure}

Across the 10 single-cell tasks, iLTM has the highest Elo point estimate, followed by LightGBM, TabM and RealMLP (Figure~\ref{fig:domain_norm_sc}). The comparison covers 1,817--100,000 samples and 500--22,000 features, with related Tahoe and mouse-kidney tasks resampled together. Notably, iLTM is the only foundation model among the seven highest-ranked models, while TabDPT ranks below MLP.

\begin{figure}[!htbp]
    \centering
    \includegraphics[width=\linewidth]{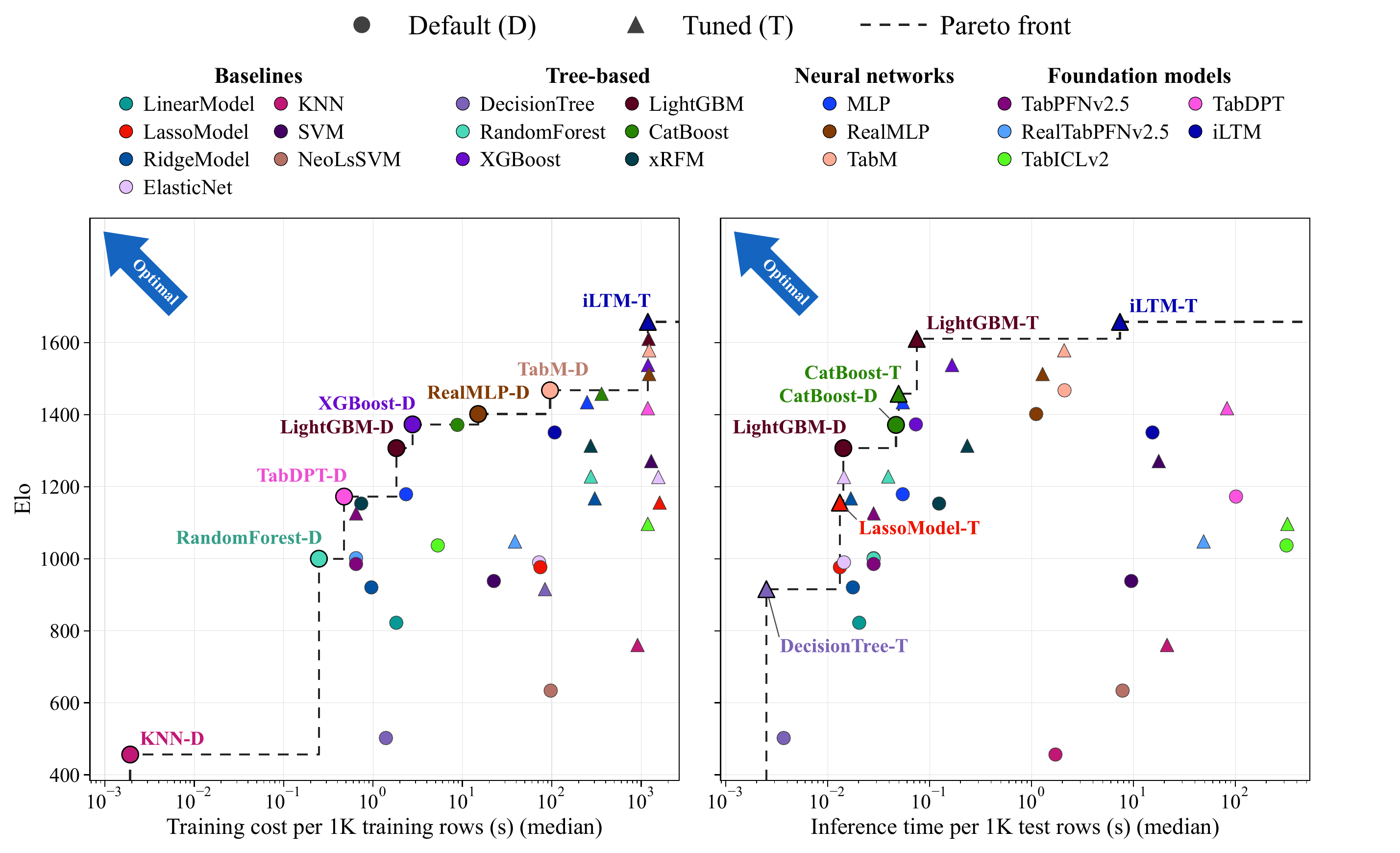}
    \caption{\textbf{Pareto trade-offs for Single-Cell datasets.} The left panel shows training cost trade-offs, while the right panel shows inference-time trade-offs.}
    \label{fig:pareto_sc}
\end{figure}

Tuned iLTM reaches the highest Elo rating and lies on both Pareto frontiers (Figure~\ref{fig:pareto_sc}). Boosting trees and neural networks offer lower training costs, while boosting trees offer lower inference costs.

\FloatBarrier
\subsubsection{Systems Biology}
\label{sec:domain_sb}

\begin{figure}[!htbp]
    \centering
    \includegraphics[width=\linewidth]{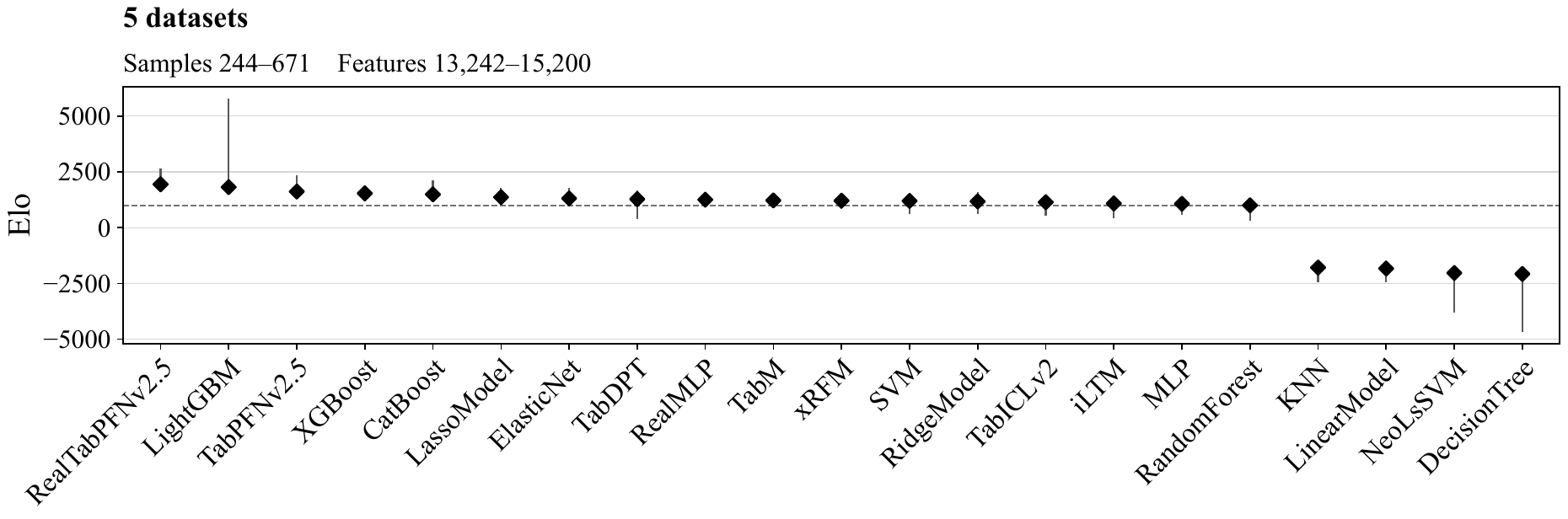}
    \caption{\textbf{Model-level Elo for Systems Biology datasets.} The 21 models are compared across all five tasks. Diamonds show point estimates, bars show 95\% cluster-bootstrap intervals, and the dashed line marks the RandomForest reference of 1,000.}
    \label{fig:domain_norm_sb}
\end{figure}

The five Systems Biology datasets contain 244--671 samples and 13,242--15,200 features (Figure~\ref{fig:domain_norm_sb}). RealTabPFNv2.5 has the highest Elo point estimate, followed by LightGBM.

\begin{figure}[!htbp]
    \centering
    \includegraphics[width=\linewidth]{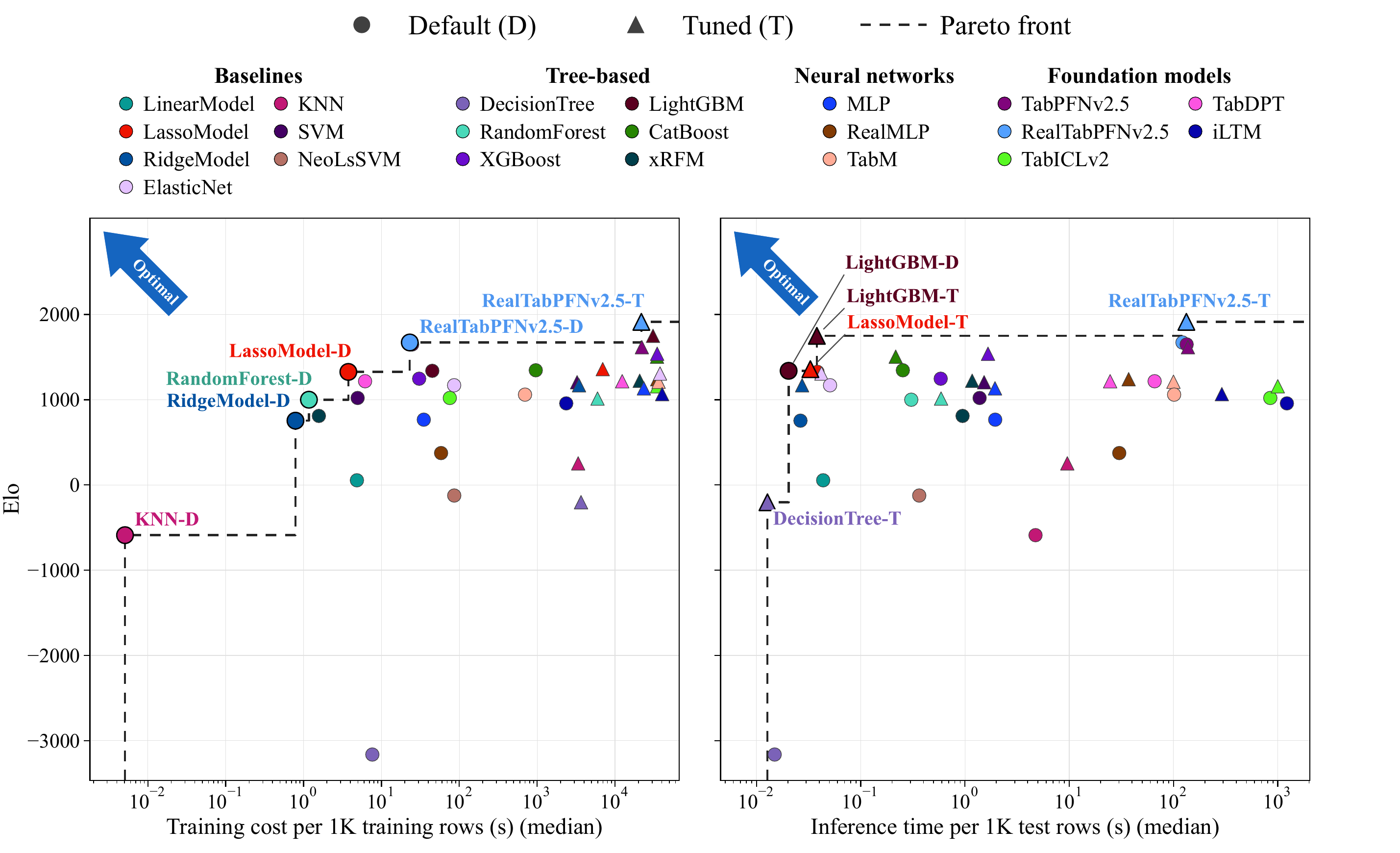}
    \caption{\textbf{Pareto trade-offs for Systems Biology datasets.} The left panel shows training cost trade-offs, while the right panel shows inference-time trade-offs.}
    \label{fig:pareto_sb}
\end{figure}

Both the Lasso and RealTabPFNv2.5 lie on the training and inference-cost frontiers, while LightGBM is competitive and provides similar inference cost as the Lasso (Figure~\ref{fig:pareto_sb}).

\FloatBarrier
\subsubsection{Transcriptomics}
\label{sec:domain_trans}

\begin{figure}[!htbp]
    \centering
    \includegraphics[width=\linewidth]{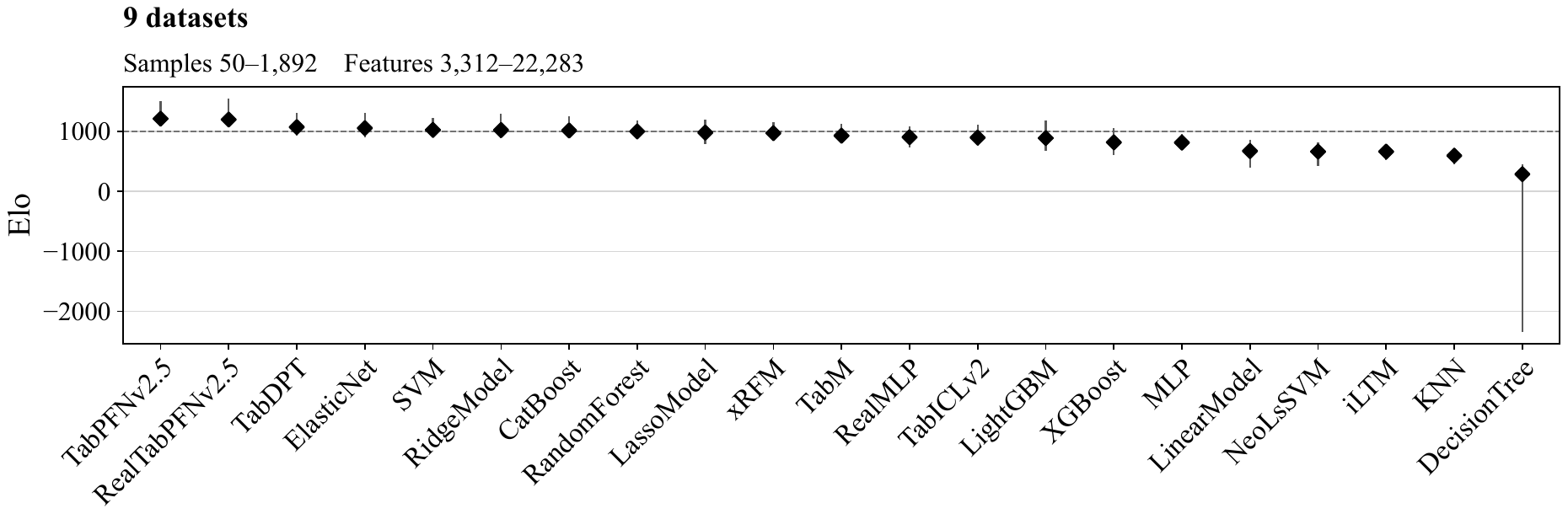}
    \caption{\textbf{Model-level Elo for Transcriptomics datasets.} The 21 models are compared across all nine tasks. Diamonds show point estimates, bars show 95\% cluster-bootstrap intervals, and the dashed line marks the RandomForest reference of 1,000.}
    \label{fig:domain_norm_trans}
\end{figure}

The nine transcriptomic datasets contain 50--1,892 samples and 3,312--22,283 features (Figure~\ref{fig:domain_norm_trans}). The TabPFN variants have the highest Elo point estimates, while ElasticNet, SVM and RidgeModel are the strongest classical baselines.

\begin{figure}[!htbp]
    \centering
    \includegraphics[width=\linewidth]{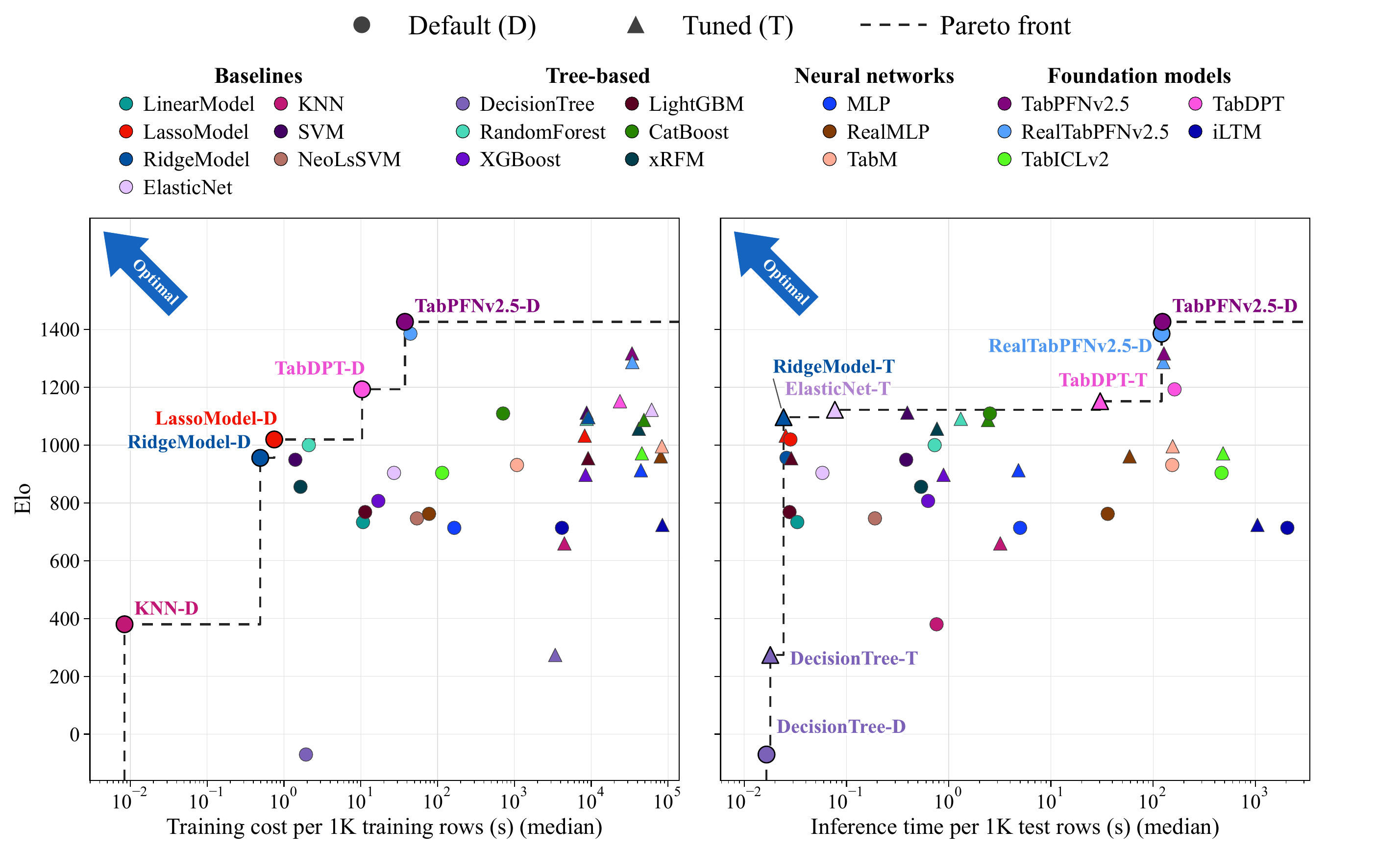}
    \caption{\textbf{Pareto trade-offs for Transcriptomics datasets.} The left panel shows training cost trade-offs, while the right panel shows inference-time trade-offs.}
    \label{fig:pareto_trans}
\end{figure}

The training-cost frontier consists entirely of Default configurations, with classical baselines on the lower training cost end and TabDPT and TabPFNv2.5 reaching the highest Elo ratings (Figure~\ref{fig:pareto_trans}). RidgeModel and ElasticNet offer very low-latency inference, while TabDPT and RealTabPFNv2.5 provide intermediate trade-offs before TabPFNv2.5 at the high-performance end.

\section{Alternative Binary Classification Metrics: Ranking Agreement and Calibration Analysis}
\label{apx:binary_metric_robustness}

\subsection{Motivation and Evaluation Protocol}
\label{apx:binary_metric_protocol}

The primary evaluation metric for binary classification in TabBioMed is the area under the receiver operating characteristic curve (ROC-AUC), consistent with established tabular benchmarking protocols. However, ROC-AUC primarily measures discriminative ability and does not directly assess the quality of predicted probabilities. In biomedical classification tasks, particularly those involving substantial class imbalance, complementary metrics can provide additional information about minority-class performance and the reliability of predicted probabilities.

To examine the sensitivity of benchmark conclusions to the choice of evaluation metric, we extend the binary classification analysis to include the area under the precision-recall curve (PR-AUC), binary log loss, Brier score, and Expected Calibration Error (ECE). PR-AUC, computed as average precision, provides a complementary discrimination-oriented assessment that is particularly informative for imbalanced classification tasks. Binary log loss and Brier score are strictly proper scoring rules that evaluate the quality of predicted probabilities, whereas ECE provides a descriptive estimate of probability calibration. 
For each model and dataset, we use complete Tuned test results when available, otherwise complete Default results, and use the same variant for all five metrics. Cross-validation results require all five outer test folds and are averaged across folds. Model--dataset pairs without either complete variant are excluded from this analysis.

For each evaluation metric, models are ranked separately within each dataset. Higher values indicate better performance for ROC-AUC and PR-AUC, whereas lower values are preferred for binary log loss, Brier score, and ECE. Model performance is subsequently summarized using mean and median ranks across datasets.

To assess ranking consistency, we compare the model rankings obtained under each alternative metric against those obtained using ROC-AUC. Agreement is quantified using Spearman's rank correlation coefficient ($\rho$) and Kendall's rank correlation coefficient ($\tau$). We additionally report the frequency of changes in the dataset-level top-ranked model and the Jaccard similarity between the sets of the three highest-ranked models under each pair of metrics.

The analysis includes 48 binary classification datasets. To investigate the influence of class imbalance, datasets are stratified according to their class imbalance ratio (IR), using a threshold of $\mathrm{IR}=2$. This yields 25 relatively balanced datasets ($\mathrm{IR}\leq 2$) and 23 imbalanced datasets ($\mathrm{IR}>2$).

\paragraph{Calibration Evaluation.}

Expected Calibration Error (ECE) is computed using the predicted probabilities of the positive class. Predictions are partitioned into $B=15$ equal-width bins over $[0,1]$. Within each bin, calibration error is the absolute difference
between the observed positive-class frequency and the mean predicted probability. ECE is the sample-weighted average of these discrepancies:
\begin{equation}
\mathrm{ECE}
=
\sum_{\substack{b=1 \\ |S_b|>0}}^{B}
\frac{|S_b|}{N}
\left|
\frac{1}{|S_b|}
\sum_{i\in S_b} y_i
-
\frac{1}{|S_b|}
\sum_{i\in S_b} \hat{p}_i
\right|,
\label{eq:binary_ece}
\end{equation}
where $S_b$ is the set of samples assigned to bin $b$, $N$ is the total number of samples, $y_i\in\{0,1\}$ is the observed binary label, and $\hat{p}_i\in[0,1]$ is the predicted probability of the positive class. Empty bins contribute zero.

ECE is used as a complementary descriptive calibration diagnostic rather than as a standalone measure of overall probabilistic performance. Accordingly, conclusions regarding predicted probability quality are primarily supported by the strictly proper scoring rules, binary log loss and Brier score, with ECE providing additional evidence about estimated calibration.

\subsection{Model Rankings Under Alternative Evaluation Metrics}
\label{apx:binary_metric_ranks}

Table~\ref{tab:binary_metric_ranks} presents the mean and median ranks of the evaluated models across ROC-AUC, PR-AUC, binary log loss, Brier score, and ECE, with models ordered according to their ROC-AUC mean rank.

\begin{table}[ht]
\centering
\caption{
Mean and median model ranks across alternative binary classification
evaluation metrics. Each cell reports the mean rank followed by the
median rank (Mean / Median), with lower ranks indicating better
relative performance. Models are ordered by increasing ROC-AUC
mean rank. The lowest mean rank for each evaluation metric is
highlighted in bold. $N$ denotes the number of datasets with complete test results for each model.
}
\label{tab:binary_metric_ranks}

\scriptsize
\setlength{\tabcolsep}{3.5pt}
\renewcommand{\arraystretch}{1.12}

\begin{tabular}{lcccccc}
\toprule
\textbf{Model} &
\textbf{$N$} &
\textbf{ROC-AUC} &
\textbf{PR-AUC} &
\textbf{Log loss} &
\textbf{Brier score} &
\textbf{ECE} \\
\midrule

RealTabPFNv2.5 & 46 &
\textbf{6.30 / 5.0} &
\textbf{5.85 / 5.0} &
8.91 / 7.0 &
8.07 / 6.0 &
12.02 / 12.0 \\

TabPFNv2.5 & 46 &
6.87 / 5.25 &
6.29 / 5.0 &
8.89 / 8.0 &
8.52 / 7.0 &
12.26 / 12.0 \\

TabICLv2 & 48 &
7.76 / 7.0 &
7.90 / 6.5 &
\textbf{6.83 / 4.5} &
\textbf{7.27 / 4.5} &
9.73 / 9.0 \\

TabPFN-Wide & 42 &
9.46 / 8.0 &
9.13 / 8.0 &
11.07 / 10.5 &
11.57 / 10.0 &
14.62 / 17.0 \\

TabDPT & 48 &
9.52 / 9.5 &
9.36 / 8.0 &
10.96 / 10.5 &
10.19 / 10.0 &
12.40 / 13.0 \\

CatBoost & 48 &
10.17 / 8.0 &
10.33 / 9.0 &
9.54 / 8.0 &
11.65 / 12.0 &
12.92 / 11.5 \\

TabM & 48 &
10.27 / 11.5 &
9.85 / 10.0 &
7.83 / 8.0 &
8.63 / 8.5 &
8.79 / 8.0 \\

iLTM & 48 &
10.38 / 11.5 &
10.38 / 11.5 &
12.60 / 12.5 &
13.48 / 14.0 &
14.98 / 17.0 \\

xRFM & 48 &
10.57 / 9.0 &
10.38 / 10.0 &
9.58 / 9.0 &
10.04 / 9.5 &
8.50 / 7.0 \\

RandomForest & 48 &
11.19 / 11.0 &
11.53 / 11.5 &
10.33 / 11.0 &
12.38 / 12.5 &
14.88 / 16.5 \\

RealMLP & 48 &
11.19 / 11.0 &
11.03 / 11.0 &
11.08 / 11.0 &
10.08 / 9.5 &
11.60 / 11.0 \\

RidgeModel & 48 &
11.33 / 11.0 &
12.09 / 12.0 &
14.35 / 16.0 &
13.35 / 14.0 &
12.67 / 14.0 \\

ElasticNet & 48 &
11.73 / 12.5 &
12.34 / 13.5 &
11.94 / 12.5 &
10.92 / 11.0 &
11.04 / 11.0 \\

LassoModel & 48 &
11.86 / 12.0 &
12.63 / 13.5 &
14.44 / 16.0 &
13.60 / 15.0 &
12.77 / 12.5 \\

SVM & 48 &
12.31 / 11.5 &
11.77 / 11.0 &
7.79 / 6.5 &
9.00 / 9.0 &
8.54 / 8.0 \\

HyperFast & 48 &
12.64 / 13.0 &
12.42 / 13.0 &
11.58 / 12.5 &
11.94 / 12.0 &
14.38 / 15.0 \\

MLP & 48 &
12.94 / 12.5 &
11.90 / 11.5 &
9.63 / 9.5 &
10.56 / 11.0 &
10.73 / 10.5 \\

XGBoost & 48 &
13.36 / 14.0 &
12.56 / 12.0 &
15.71 / 18.0 &
17.13 / 20.0 &
16.21 / 19.0 \\

LightGBM & 48 &
14.16 / 15.0 &
13.73 / 14.5 &
13.13 / 15.0 &
15.04 / 17.0 &
10.48 / 10.0 \\

LinearModel & 48 &
14.89 / 17.0 &
15.34 / 16.5 &
17.19 / 19.5 &
15.83 / 17.0 &
13.54 / 15.5 \\

KNN & 48 &
16.29 / 17.0 &
16.40 / 18.0 &
16.04 / 17.5 &
14.60 / 16.0 &
13.04 / 14.0 \\

NeoLsSVM & 48 &
17.10 / 18.0 &
18.85 / 19.5 &
14.52 / 16.0 &
12.15 / 12.5 &
\textbf{5.94 / 6.0} \\

DecisionTree & 48 &
20.77 / 22.0 &
20.92 / 22.0 &
19.50 / 22.0 &
17.48 / 19.0 &
12.15 / 12.0 \\

\bottomrule
\end{tabular}
\end{table}

The aggregate rankings show that RealTabPFNv2.5 achieves the lowest mean rank under both ROC-AUC (6.30) and PR-AUC (5.85), indicating that the leading model remains unchanged across the two discrimination-oriented metrics. TabPFNv2.5 also maintains competitive rankings under both metrics, with mean ranks of 6.87 and 6.29, respectively. These findings are consistent with the strong ranking agreement observed between ROC-AUC and PR-AUC in Table~\ref{tab:binary_metric_agreement}.

In contrast, the rankings under binary log loss reveal a different model preference. TabICLv2 achieves the lowest
aggregate mean rank (6.83), followed by SVM (7.79) and TabM (7.83), whereas RealTabPFNv2.5 obtains a mean rank of 8.91. This shift indicates that the model leading under discrimination-oriented metrics does not necessarily maintain its position when evaluated using predicted probability quality.

A similar change in the aggregate leading model is observed under Brier score. TabICLv2 achieves the lowest mean rank (7.27), followed by RealTabPFNv2.5 (8.07), TabPFNv2.5 (8.52), and TabM (8.63). Thus, although RealTabPFNv2.5 remains competitive under Brier score, TabICLv2 exhibits the most favorable aggregate ranking under both proper scoring rules, binary log loss and Brier score.

The most pronounced change in aggregate model ordering is observed under ECE. NeoLsSVM achieves the lowest mean rank (5.94), despite obtaining substantially less favorable rankings under ROC-AUC (17.10) and PR-AUC (18.85).
Conversely, RealTabPFNv2.5, which leads under the discrimination-oriented metrics, obtains an ECE mean rank of 12.02. These results indicate that models with strong discriminative performance do not necessarily exhibit the lowest estimated calibration error.

\subsection{Agreement Between Alternative Evaluation Metrics}
\label{apx:binary_metric_agreement}

To evaluate the robustness of model rankings to the choice of evaluation metric, we compare rankings obtained using PR-AUC, binary log loss, Brier score, and Expected Calibration Error (ECE) against the primary ROC-AUC-based rankings.

\begin{table}[ht]
\centering
\caption{
Agreement between ROC-AUC and alternative evaluation metrics across
48 binary classification datasets. Rank correlations are computed
across models within each dataset and summarized using the median
across datasets. Top-1 changes indicate datasets for which the
highest-ranked model differs from that selected by ROC-AUC.
Top-3 Jaccard similarity quantifies the overlap between the sets
of the three highest-ranked models under the two metrics.
}
\label{tab:binary_metric_agreement}

\small
\setlength{\tabcolsep}{5pt}

\begin{tabular}{lcccc}
\toprule
\textbf{Statistic} &
\textbf{PR-AUC} &
\textbf{Log loss} &
\textbf{Brier score} &
\textbf{ECE} \\
\midrule

Median Spearman $\rho$
& 0.926 & 0.608 & 0.674 & 0.130 \\

Median Kendall $\tau$
& 0.800 & 0.452 & 0.520 & 0.083 \\

Top-1 changes
& 19/48 & 28/48 & 26/48 & 44/48 \\

Top-1 change rate (\%)
& 39.6 & 58.3 & 54.2 & 91.7 \\

Median top-3 Jaccard similarity
& 0.500 & 0.200 & 0.200 & 0.000 \\

\midrule
\multicolumn{5}{l}{
\textit{Relatively balanced datasets
($\mathrm{IR}\leq 2$, $n=25$)}
} \\

Median Spearman $\rho$
& 0.949 & 0.705 & 0.774 & 0.166 \\

Top-1 changes
& 7/25 & 13/25 & 10/25 & 23/25 \\

Top-1 change rate (\%)
& 28.0 & 52.0 & 40.0 & 92.0 \\

\midrule
\multicolumn{5}{l}{
\textit{Imbalanced datasets
($\mathrm{IR}>2$, $n=23$)}
} \\

Median Spearman $\rho$
& 0.917 & 0.489 & 0.498 & 0.087 \\

Top-1 changes
& 12/23 & 15/23 & 16/23 & 21/23 \\

Top-1 change rate (\%)
& 52.2 & 65.2 & 69.6 & 91.3 \\

\bottomrule
\end{tabular}
\end{table}

ROC-AUC and PR-AUC exhibit strong ranking agreement, with a median Spearman correlation of $\rho=0.926$ and Kendall correlation of $\tau=0.800$. This indicates that models performing well under ROC-AUC generally maintain favorable relative rankings under PR-AUC. The dataset-level top-ranked model changes on 19 of the 48 datasets (39.6\%), indicating that strong overall ranking agreement does not necessarily imply identical model selection for individual datasets. The median top-3 Jaccard similarity of 0.500 further indicates partial overlap among the highest-ranked models.

Agreement is more moderate when ROC-AUC is compared with binary log loss and Brier score. Binary log loss yields median correlations of $\rho=0.608$ and $\tau=0.452$, while Brier score produces $\rho=0.674$ and $\tau=0.520$. The top-ranked model changes on 28 of the 48 datasets (58.3\%) under binary log loss and 26 datasets (54.2\%) under Brier score. Both metrics exhibit a median top-3 Jaccard similarity of 0.200, indicating limited overlap among the highest-ranked models. These findings suggest that discrimination-based rankings do not fully capture differences in predicted probability quality.

The greatest divergence is observed for ECE, which exhibits weak agreement with ROC-AUC ($\rho=0.130$, $\tau=0.083$). The top-ranked model differs on 44 of the 48 datasets (91.7\%), and the median top-3 Jaccard similarity is zero.

\paragraph{Influence of Class Imbalance on Ranking Agreement.}

\begin{figure*}[ht]
    \centering
    \includegraphics[width=\textwidth]{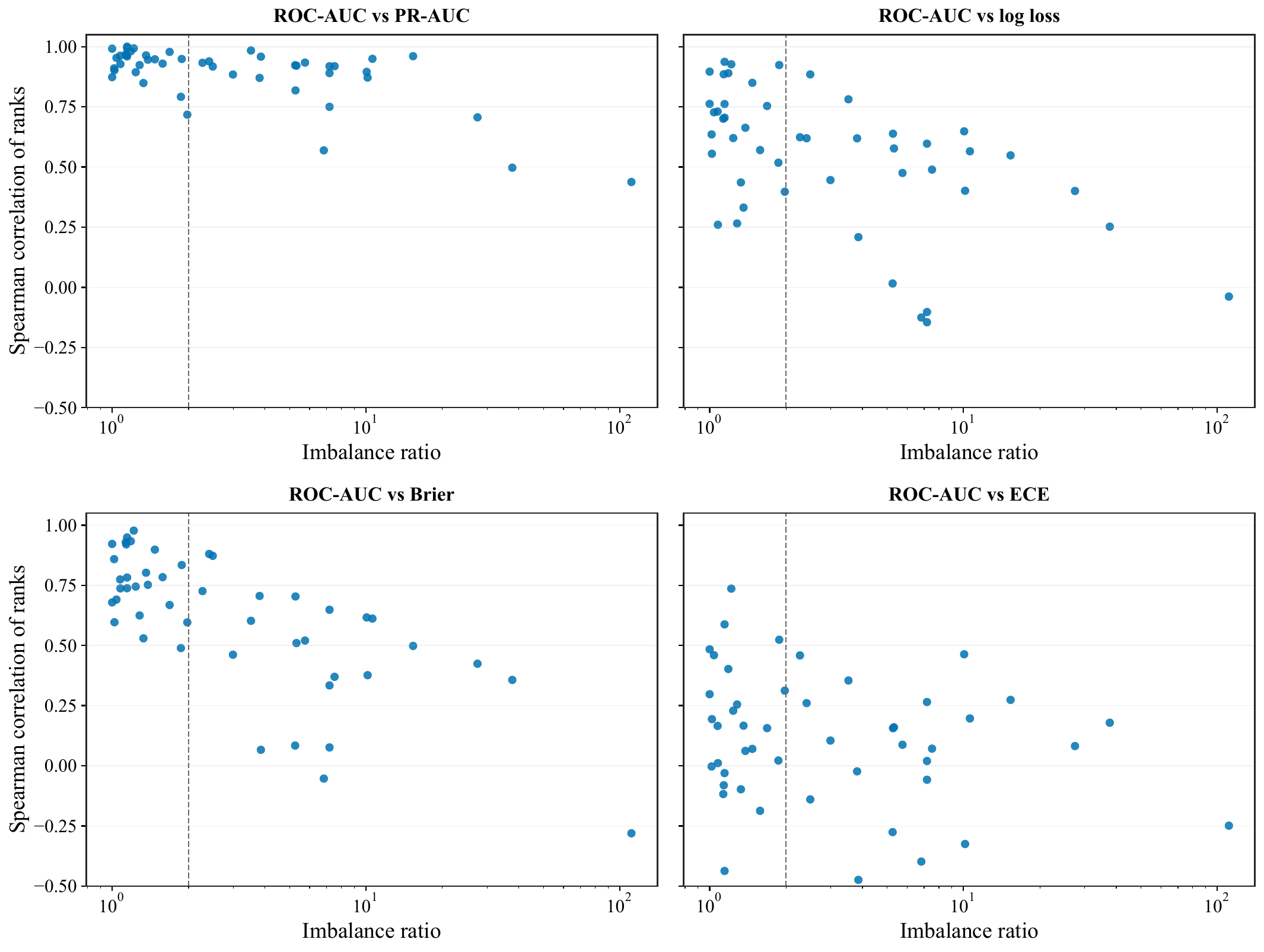}
    \caption{
    \textbf{Relationship between class imbalance and model ranking
    agreement across binary classification datasets.}
    Each point represents one dataset, with the horizontal
    axis showing the class imbalance ratio on a logarithmic
    scale and the vertical axis showing the dataset-level
    Spearman rank correlation between ROC-AUC and
    (a) PR-AUC, (b) binary log loss, (c) Brier score,
    and (d) Expected Calibration Error (ECE).
    The dashed vertical line indicates the threshold
    $\mathrm{IR}=2$, separating relatively balanced
    ($\mathrm{IR}\leq 2$, $n=25$) and imbalanced
    ($\mathrm{IR}>2$, $n=23$) datasets.
    Higher correlations indicate greater agreement
    between the model rankings obtained under ROC-AUC
    and the corresponding alternative metric.
    }
    \label{fig:binary_imbalance_agreement}
\end{figure*}

To further examine the relationship between class imbalance and ranking consistency, Figure~\ref{fig:binary_imbalance_agreement} shows the dataset-level Spearman rank correlations between ROC-AUC and each alternative evaluation metric as a function of the class imbalance ratio.

ROC-AUC and PR-AUC maintain high median ranking agreement in both relatively balanced and imbalanced datasets (0.949 and 0.917, respectively), although Figure~\ref{fig:binary_imbalance_agreement} shows lower agreement for several highly imbalanced datasets. Larger decreases occur for binary log loss (0.705 to 0.489) and Brier score (0.774 to 0.498). Changes in the top-ranked model are also more frequent among imbalanced datasets under PR-AUC, binary log loss and Brier score (Table~\ref{tab:binary_metric_agreement}). ECE exhibits weak agreement with ROC-AUC in both groups (0.166 and 0.087), indicating that differences between discrimination and estimated calibration are not restricted to highly imbalanced tasks.

\section{In-Depth Metafeature and Regime Analysis}
\label{apx:metafeature_regime_analysis}

We analyze all 94 datasets and the same main 21 models. Score selection, normalization, and failure handling follow Appendix~\ref{apx:cd_analysis}. For each dataset, family performance is the best observed normalized test score within that family. The portfolios contain seven classical baselines, six tree-based models, three neural models, and five foundation models.

\subsection{Metafeature Extraction and Correlation Analysis}

\begin{figure}[!htbp]
    \centering
    \includegraphics[width=\linewidth]{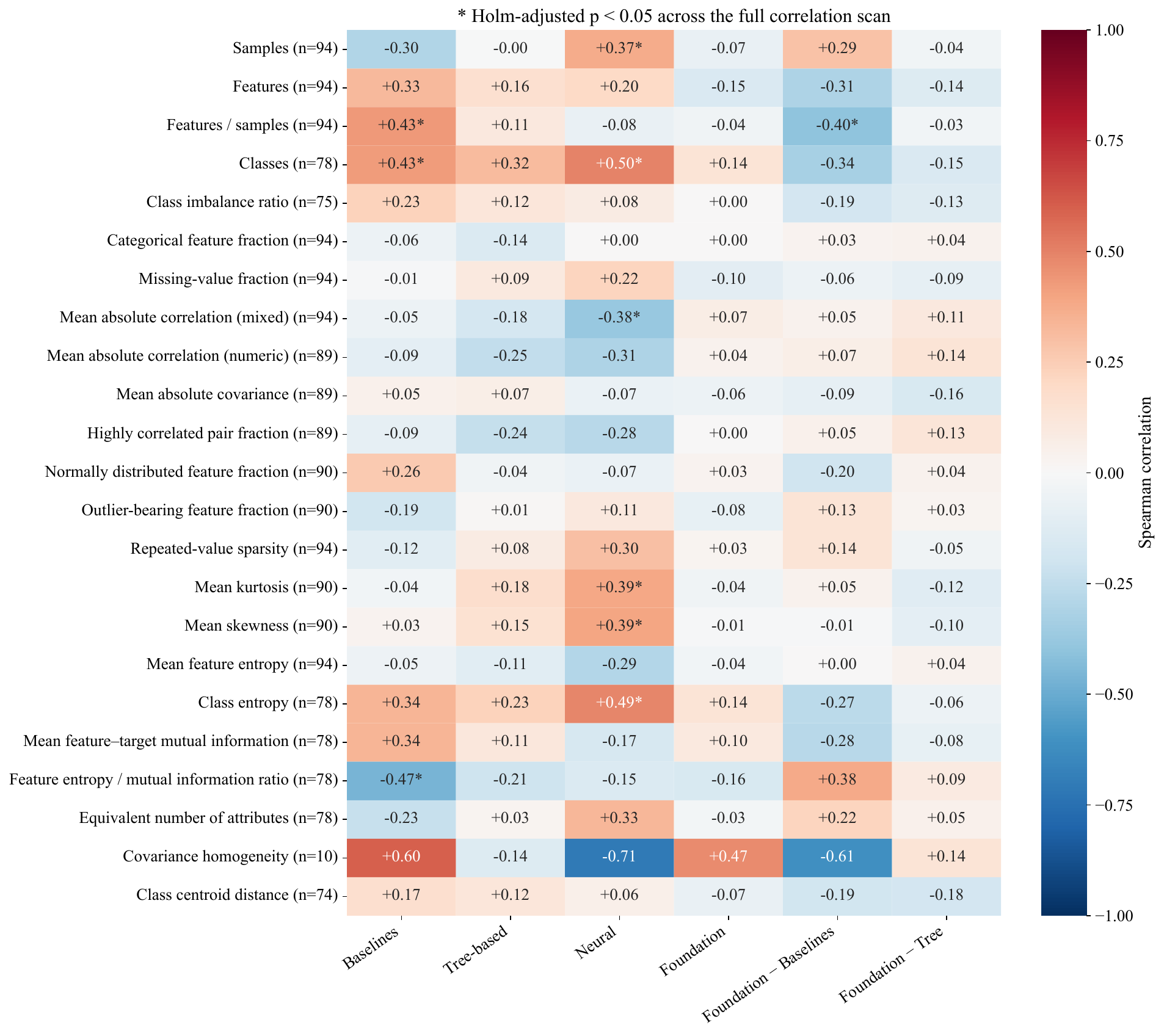}
    \caption{\textbf{Selected dataset metafeatures and relative model performance.} Spearman correlations with the best observed normalized score within each family and the foundation--baseline and foundation--tree score gaps. The valid dataset count is shown for each descriptor. Asterisks indicate Holm-adjusted $p < 0.05$ across all 180 comparisons.}
    \label{fig:metafeature_correlations}
\end{figure}

We extract metafeatures using PyMFE \cite{pymfe} from the stored training partitions, covering 210 partitions across the benchmark. For cross-validation datasets, descriptors are computed separately within each of the five outer training partitions and averaged only when all five values are available. Statistical metafeatures use deterministic samples of at most 2,048 training rows and 512 original features per partition, while dataset-size descriptors remain exact.
Numerical PyMFE statistics use numerical features. The mean absolute feature correlation used in the main text additionally includes one-hot encoded categorical features, with at most 512 encoded columns. Figure~\ref{fig:metafeature_correlations} reports correlations and valid dataset counts for selected descriptors.

We compute Spearman correlations between 30 metafeatures and six performance summaries: the best observed score within each of the four model families, the foundation--baseline score gap, and the foundation--tree score gap. Holm correction is applied jointly across all 180 comparisons.

Figure~\ref{fig:metafeature_correlations} highlights the increasing competitiveness of classical baselines at higher feature-to-sample ratios and the associations of neural performance with sample size, class count, and feature correlation.

\subsection{Dataset Regimes and Family Comparisons}
\label{sec:archetypes}

We cluster all 94 datasets using standardized log sample size, log feature-to-sample ratio, and mean absolute feature correlation. We compare K-Means and Ward clustering for $k=2,\ldots,8$ across six descriptor sets, assessing stability over 50 random subsamples containing 80\% of datasets, with standardization and clustering refitted in each subsample. Stability is measured by agreement with the full-data partition on the shared datasets, and for these choices we use dataset properties only.

We retain the three size/correlation descriptors and use K-Means with $k=3$. This partition has mean adjusted Rand index 0.974 under subsampling and silhouette score 0.388, compared with 0.676 and 0.353 for $k=4$. The two-cluster solution has slightly higher separation (0.399) but lower stability (0.952) and provides less detail.

\begin{figure}[t]
    \centering
    \includegraphics[width=\linewidth]{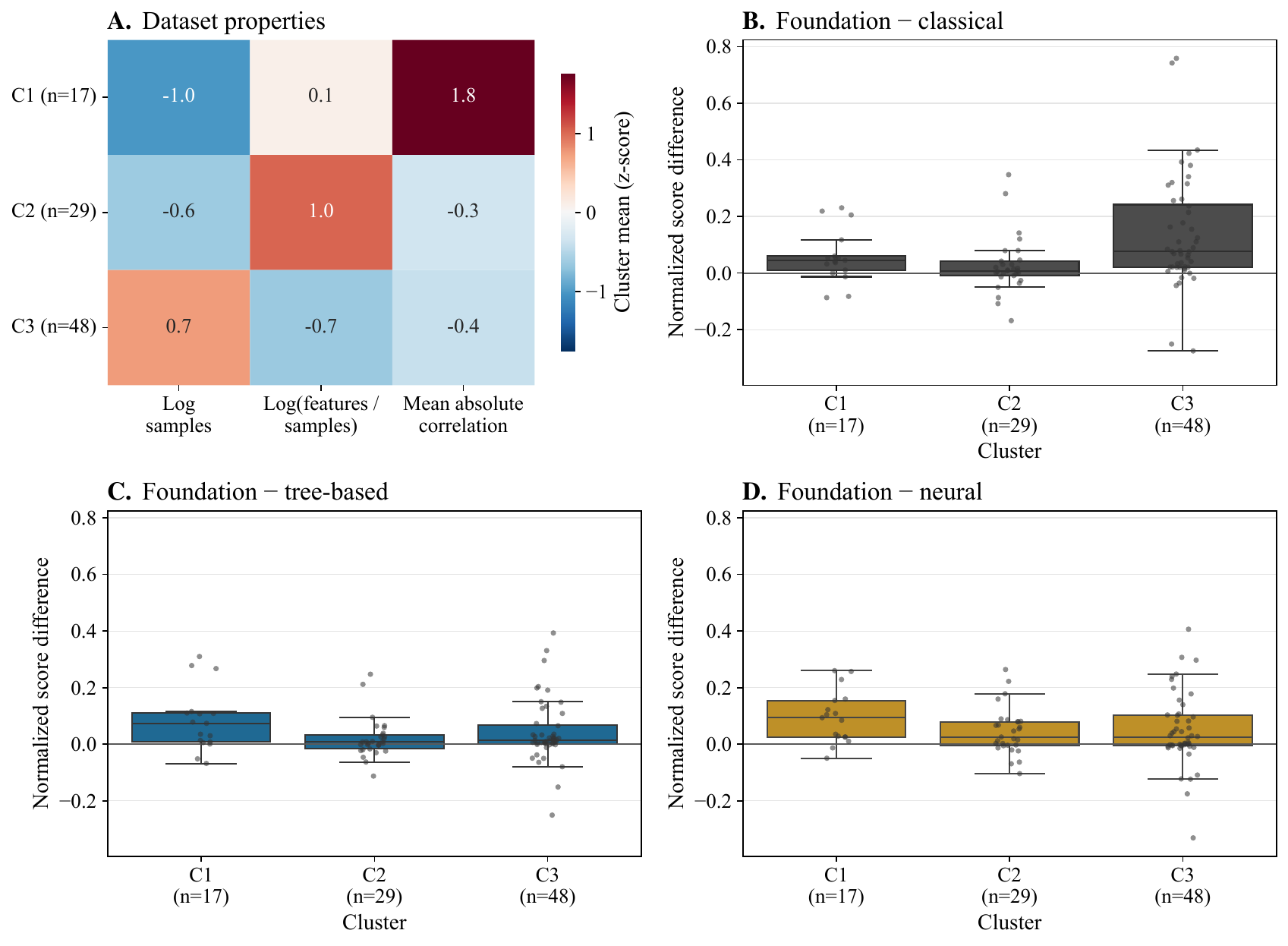}
    \caption{\textbf{Dataset regimes and family comparisons across all 94 datasets.} \textbf{A.} Mean standardized dataset properties within each cluster. \textbf{B--D.} Best foundation minus best classical, tree-based, and neural normalized scores, respectively, on a shared vertical scale. Positive differences favor foundation models. Boxes span the interquartile range, center lines mark medians, and dots represent datasets. Model performance is not used to form the clusters.}
    \label{fig:discovery_archetypes}
\end{figure}

Figure~\ref{fig:discovery_archetypes} shows the cluster profiles and three separate comparisons between the best observed foundation model and the best classical, tree-based, or neural model within each dataset. Positive differences favor foundation models.

The three profiles in panel~A of Figure~\ref{fig:discovery_archetypes} are:
\begin{itemize}
    \item \textbf{C1, smaller samples and strong feature correlations:} 17 datasets with median 267 samples and 52 features.
    \item \textbf{C2, smaller samples and many features per sample:} 29 datasets with median 366 samples and 13,242 features.
    \item \textbf{C3, larger samples and fewer features per sample:} 48 datasets with median 6,465 samples and 588 features.
\end{itemize}

Panels~B--D of Figure~\ref{fig:discovery_archetypes} show that the relative advantage depends on the competing family. The largest mean gap over classical baselines (0.136) occurs in C3, which combines larger sample sizes with lower feature-to-sample ratios. The gaps over trees and neural models are largest in the highly correlated, smaller-sample cluster C1 (0.084 and 0.096, respectively). The high-dimensional, smaller-sample cluster C2 has more modest mean gaps (0.027, 0.020, and 0.045 against classical, tree-based, and neural models, respectively).

\clearpage
\section*{NeurIPS Paper Checklist}
\addcontentsline{toc}{section}{NeurIPS Paper Checklist}

\begin{enumerate}

\item {\bf Claims}
    \item[] Question: Do the main claims made in the abstract and introduction accurately reflect the paper's contributions and scope?
    \item[] Answer: \answerYes{} %
    \item[] Justification: The abstract and introduction clearly describe the paper’s scope, contributions, and primary findings. Specifically, we claim that we: (1) introduce TabBioMed, a large-scale benchmark for biomedical tabular machine learning, (2) curate and standardize datasets across multiple biomedical domains, (3) evaluate a diverse set of classical, deep learning, and foundation-model-based approaches under unified evaluation protocols, (4) analyze default versus hyperparameter-optimized model performance, (5) provide ranking, efficiency, and dataset-regime-dependent performance analyses, and (6) release reproducible evaluation pipelines and benchmark resources.

    In addition, we claim that our empirical results demonstrate: (7) strong performance of tabular foundation models across eight biomedical domains, (8) substantial effects of hyperparameter optimization and refitting strategies and (9) performance vs.\ computational trade-offs across all the biomedical domains considered in the benchmark. These claims are supported throughout the paper as follows: dataset curation and benchmark construction are described in Subsection \ref{sec:datasets}; experimental protocols, preprocessing pipelines, and evaluation procedures are presented in Subsections~\ref{sec:models_hpo} and~\ref{sec:evaluation_metrics}. Large-scale benchmark results, efficiency, and scaling analyses, statistical ranking analyses, including Elo ratings and critical difference evaluations, are presented in Section~\ref{sec:results_analysis} and Appendix~\ref{apx:additional_results}. Reproducibility resources, implementation details, and extended benchmark analyses are provided in Appendices \ref{apx:dataset_curation_processing}, \ref{sec:model_implementation}, and \ref{apx:additional_results}.


\item {\bf Limitations}
    \item[] Question: Does the paper discuss the limitations of the work performed by the authors?
    \item[] Answer: \answerYes{} %
    \item[] Justification: Our limitations are discussed in Section \ref{sec:conclusion_limitations}.

\item {\bf Theory assumptions and proofs}
    \item[] Question: For each theoretical result, does the paper provide the full set of assumptions and a complete (and correct) proof?
    \item[] Answer: \answerNA{} %
    \item[] Justification: The paper does not include theoretical results, formal guarantees, or mathematical claims requiring proof. The contributions are empirical and methodological, focusing on dataset curation, preprocessing standardization, and benchmark evaluation.

    \item {\bf Experimental result reproducibility}
    \item[] Question: Does the paper fully disclose all the information needed to reproduce the main experimental results of the paper to the extent that it affects the main claims and/or conclusions of the paper (regardless of whether the code and data are provided or not)?
    \item[] Answer: \answerYes{} %
    \item[] Justification: The paper provides the information needed to reproduce the main experimental results. Dataset construction, preprocessing, and splitting are described in Subsection~\ref{sec:datasets} and Appendix~\ref{apx:dataset_curation_processing}. Model implementations and hyperparameter optimization are described in Subsection~\ref{sec:models_hpo} and Appendices~\ref{sec:model_implementation}, \ref{apx:model_impl_train_and_hpo}, and \ref{apx:hpo_search_space}. Evaluation metrics, score normalization, and benchmarking protocols are described in Subsection~\ref{sec:evaluation_metrics} and Appendix~\ref{apx:benchmark_evaluation}. The paper also reports the Optuna configuration, number of trials, computational budgets, random seed, and saved artifacts, providing a clear path to reproduce or verify the main empirical claims.

\item {\bf Open access to data and code}
    \item[] Question: Does the paper provide open access to the data and code, with sufficient instructions to faithfully reproduce the main experimental results, as described in supplemental material?
    \item[] Answer: \answerYes{} %
    \item[] Justification: We provide all the datasets with references, sources, and download links in Appendix \ref{apx:dataset_curation_processing} along with benchmark code publicly at \url{https://github.com/AI-sandbox/TabBioMed}, as well as instructions on how to reproduce the results of the benchmark code in the repository.

\item {\bf Experimental setting/details}
    \item[] Question: Does the paper specify all the training and test details (e.g., data splits, hyperparameters, how they were chosen, type of optimizer) necessary to understand the results?
    \item[] Answer: \answerYes{} %
    \item[] Justification: We provide our dataset curation and preprocessing along with the benchmark experiment setup in Subsections \ref{sec:datasets} and \ref{sec:models_hpo} and provide additional details in Appendices \ref{apx:dataset_curation_processing} and \ref{sec:model_implementation}.

\item {\bf Experiment statistical significance}
    \item[] Question: Does the paper report error bars suitably and correctly defined or other appropriate information about the statistical significance of the experiments?
    \item[] Answer: \answerYes{} %
    \item[] Justification: We report multiple forms of statistical uncertainty and significance analysis. Specifically, the paper includes statistical ranking analyses using Friedman, pairwise Wilcoxon-Holm significance analyses, and Elo-rating-based comparisons across models and benchmark settings. Confidence intervals and uncertainty estimates are reported where appropriate, including bootstrapped confidence intervals for Elo ratings and variability analyses across datasets and domains in Section \ref{sec:results_analysis}. In addition, the paper reports aggregated evaluations across multiple biomedical domains, task types, and experimental settings rather than relying on single-dataset conclusions in Appendix \ref{apx:additional_results}.

\item {\bf Experiments compute resources}
    \item[] Question: For each experiment, does the paper provide sufficient information on the computer resources (type of compute workers, memory, time of execution) needed to reproduce the experiments?
    \item[] Answer: \answerYes{} %
    \item[] Justification: We report cumulative main-benchmark runtime in Subsection~\ref{sec:models_hpo} and specify compute worker hardware (CPU cores, RAM, 24\,GB RTX A5500 GPUs, and 24-hour Slurm limits), supplementary 200-cap HPO continuations, external PCA runs, and 80\,GB A100 GPU retries in Appendices~\ref{apx:model_impl_train_and_hpo}, \ref{apx:hpo_budget_sensitivity}, \ref{apx:a100_sensitivity}, and~\ref{apx:pca_tfm}.

\item {\bf Code of ethics}
    \item[] Question: Does the research conducted in the paper conform, in every respect, with the NeurIPS Code of Ethics \url{https://neurips.cc/public/EthicsGuidelines}?
    \item[] Answer: \answerYes{} %
    \item[] Justification: The research was conducted following the NeurIPS code and ethical guidelines. Specifically, public datasets were used in accordance with their original licenses and usage conditions where applicable.

\item {\bf Broader impacts}
    \item[] Question: Does the paper discuss both potential positive societal impacts and negative societal impacts of the work performed?
    \item[] Answer: \answerYes{} %
    \item[] Justification: We discuss both potential positive and negative societal impacts in Section \ref{sec:conclusion_limitations}.

\item {\bf Safeguards}
    \item[] Question: Does the paper describe safeguards that have been put in place for responsible release of data or models that have a high risk for misuse (e.g., pre-trained language models, image generators, or scraped datasets)?
    \item[] Answer: \answerNA{} %
    \item[] Justification: Our work focuses on benchmarking and evaluating existing machine learning methods on publicly available biomedical tabular datasets obtained from established benchmark repositories and research sources. Additionally, we do not release any new models in this paper.

\item {\bf Licenses for existing assets}
    \item[] Question: Are the creators or original owners of assets (e.g., code, data, models), used in the paper, properly credited and are the license and terms of use explicitly mentioned and properly respected?
    \item[] Answer: \answerYes{} %
    \item[] Justification: We provide detailed information on our dataset curation and sources in Appendix \ref{apx:dataset_curation_processing}. We reference all dataset sources and respected the licenses of all code, models, and datasets, which are documented in \url{https://huggingface.co/TabBioMed}.

\item {\bf New assets}
    \item[] Question: Are new assets introduced in the paper well documented and is the documentation provided alongside the assets?
    \item[] Answer: \answerYes{} %
    \item[] Justification: We document the benchmark assets, including dataset curation procedures, preprocessing pipelines, evaluation protocols, model configurations, statistical analysis procedures, and benchmark usage instructions throughout the paper (Subsections \ref{sec:datasets}, \ref{sec:models_hpo} and \ref{sec:evaluation_metrics}), Appendix \ref{apx:dataset_curation_processing} and accompanying repository. Detailed instructions for reproducing experiments, running benchmark evaluations, and accessing benchmark resources are provided in the public GitHub repository at \url{https://github.com/AI-sandbox/TabBioMed}. The repository additionally documents dependencies, benchmark structure, configuration settings, dataset metadata, the Croissant file, and implementation details necessary for reproducing the reported results.

\item {\bf Crowdsourcing and research with human subjects}
    \item[] Question: For crowdsourcing experiments and research with human subjects, does the paper include the full text of instructions given to participants and screenshots, if applicable, as well as details about compensation (if any)?
    \item[] Answer: \answerNA{} %
    \item[] Justification: The paper does not involve crowdsourcing or research with human subjects.

\item {\bf Institutional review board (IRB) approvals or equivalent for research with human subjects}
    \item[] Question: Does the paper describe potential risks incurred by study participants, whether such risks were disclosed to the subjects, and whether Institutional Review Board (IRB) approvals (or an equivalent approval/review based on the requirements of your country or institution) were obtained?
    \item[] Answer: \answerNA{} %
    \item[] Justification: The paper does not involve crowdsourcing or research with human subjects.

\item {\bf Declaration of LLM usage}
    \item[] Question: Does the paper describe the usage of LLMs if it is an important, original, or non-standard component of the core methods in this research? Note that if the LLM is used only for writing, editing, or formatting purposes and does \emph{not} impact the core methodology, scientific rigor, or originality of the research, declaration is not required.
    \item[] Answer: \answerNA{} %
    \item[] Justification: LLMs have not been used as a core method to conduct this research.

\end{enumerate}

\end{document}